\documentclass[jair,twoside,11pt,theapa]{article}
\usepackage{jair, theapa, rawfonts}
\pdfoutput=1

\jairheading{60}{2017}{491-548}{06/16}{10/17}
\ShortHeadings{On Hash-Based Work Distribution Methods for  Parallel Best-First Search}
{Jinnai \& Fukunaga}
\firstpageno{491}

\usepackage{times}
\usepackage{helvet}
\usepackage{courier}

\usepackage{graphicx}

\usepackage[ruled,boxed,linesnumbered,noend]{algorithm2e}
\SetKwInOut{Input}{Input}
\usepackage{mathtools}
\usepackage{amsmath}

\usepackage{appendix}
\usepackage{enumitem}
\usepackage{subfig}
\usepackage{adjustbox}
\usepackage{rotating}
\usepackage{multirow}
\usepackage{relsize}
\usepackage{xspace}
\usepackage{pdflscape}
\usepackage{tabularx}
\usepackage{listings}
\usepackage{lstlang0}

\usepackage{comment}
\usepackage[list-entry=heading]{caption}
\usepackage{newfloat}

\DeclareFloatingEnvironment{experiment}

\newcommand{\sceff}{\mathit{eff_{esti}}}   
\newcommand{\teff}{\mathit{eff_{actual}}}    
\newcommand{\ceff}{\mathit{eff_{c}}}    
\newcommand{\seff}{\mathit{eff_{s}}}    
\newcommand{\wg}{G_W}    

\newcommand{\pddl}[1]{{\relsize{-1} \textsf{#1}}}

\newcommand{\simple}{GreedyAFG}

\newcommand{\supermicro}{Intel Xeon E5-2650 v2 2.60 GHz CPU with 128 GB RAM}
\newcommand{\lucy}{Intel Xeon E5410 2.33 GHz CPU with 16 GB RAM}

\newtheorem{observation}{Observation}

\newcommand{\term}[1]{$\mathit{#1}$} 
\newcommand{\hda}{H\!D\!A^*\!}

\newcommand{\GRAZHDAS}{\term{\hda[Z, A_{feature}/DTG_{sparsity}]}}
\newcommand{\GRAZHDA}{\term{\hda[Z, A_{feature}/DTG]}}
\newcommand{\FAZHDA}{\term{\hda[Z, A_{feature}/DTG_{fluency}]}} 
\newcommand{\GAZHDA}{\term{\hda[Z, A_{feature}/DTG_{greedy}]}}

\newcommand{\AZHDA}{\term{\hda[Z, A_{feature}]}} 

\newcommand{\DAHDA}{\term{\hda[Z, A_{state}/SDD_{dynamic}]}}
\newcommand{\OZHDA}{\term{\hda[Z_{operator}]}} 

\newcommand{\AHDA}{\term{\hda[P, A_{state}]}} 
\newcommand{\AHDAP}{\term{\hda[P, A_{state}]}}
\newcommand{\AHDAPSSD}{\term{\hda[P, A_{state}/SDD]}}

\newcommand{\AHDAZ}{\term{\hda[Z, A_{state}]}}
\newcommand{\AHDAZSSD}{\term{\hda[Z, A_{state}/SDD]}}

\newcommand{\HWD}{\term{\hda[Hyperplane]}}
\newcommand{\PHDA}{\term{\hda[P]}}
\newcommand{\ZHDA}{\term{\hda[Z]}}

\newcommand{\GRAZHDAST}{\term{[Z, A_{feauture}/DTG_{sparsity}]}}
\newcommand{\FAZHDAT}{\term{[Z, A_{feauture}/DTG_{fluency}]}}
\newcommand{\GAZHDAT}{\term{[Z, A_{feauture}/DTG_{greedy}]}}
\newcommand{\OZHDAT}{\term{[Z_{operator}]}}
\newcommand{\DAHDAT}{\term{[Z, A_{state}/SDD_{dynamic}]}}
\newcommand{\AHDAT}{\term{[Z, A_{state}/SDD]}}
\newcommand{\ZHDAT}{\term{[Z]}}

\newcommand{\apDAHDA}{\ref{appendix:dahda*}} 

\title{On Hash-Based Work Distribution Methods for \\ Parallel Best-First Search} 
\author{\name Yuu Jinnai	\email ddyuudd@gmail.com \\
	\name Alex Fukunaga	\email fukunaga@idea.c.u-tokyo.ac.jp \\
	\addr Graduate School of Arts and Sciences \\
	The University of Tokyo \\
	Tokyo, Japan}

\begin{document}

\maketitle

\begin{abstract}
Parallel best-first search algorithms such as Hash Distributed A* (HDA*) distribute work among the processes using a global hash function.
We analyze the search and communication overheads of state-of-the-art hash-based parallel best-first search algorithms, and show that 
although Zobrist hashing, the standard hash function used by HDA*, achieves good load balance for many domains, it incurs significant communication overhead since almost all generated nodes are transferred to a different processor than their parents.
We propose Abstract Zobrist hashing, a new work distribution method for parallel search which, instead of computing a hash value based on the raw features of a state, uses a feature projection function to generate a set of abstract features which results in a higher locality, resulting in reduced communications overhead.
We show that Abstract Zobrist hashing outperforms previous methods on search domains using hand-coded, domain specific feature projection functions.
We then propose GRAZHDA*, a graph-partitioning based approach to automatically generating feature
 projection functions.
GRAZHDA* seeks to approximate the partitioning of the actual search space graph by partitioning the domain transition graph, an abstraction of the state space graph.
We show that GRAZHDA* outperforms previous methods on domain-independent planning.



\end{abstract}

\section{Introduction}
The A* algorithm \cite{HartNR68} is used in many areas of AI, including planning, scheduling, path-finding, and sequence alignment.
Parallelization of A* can yield speedups as well as a way to overcome memory limitations -- 
the aggregate memory available in a cluster can allow problems that can't be  solved using a single machine to be solved. 
Thus, designing scalable, parallel search algorithms is an important goal.
The major issues which need to be addressed when designing parallel search algorithms are search overhead (states which are unnecessarily generated by parallel search but not by sequential search), communications overhead (overheads associated with moving work among threads), and coordination overhead (synchronization overhead). 

Hash Distributed A* (HDA*) is a parallel best-first search algorithm 
in which each processor executes A* using local open/closed lists, and
generated nodes are assigned (sent) to processors according to a global hash function \cite{kishimotofb13}.
HDA* can be used in distributed memory systems as well as multi-core, shared memory machines, and has been shown to scale up to hundreds of cores with little search overhead. 

The performance of HDA* depends on the hash function used for assigning nodes to processors.
\citeauthor{kishimotofb09} \citeyear{kishimotofb09,kishimotofb13} showed that using the Zobrist hash function \citeyear{Zobrist1970}, HDA* could achieve good load balance and low search overhead.
\citeauthor{burnslrz10} \citeyear{burnslrz10} noted that Zobrist hashing incurs a heavy communication overhead because many nodes are assigned to processes that are different from their parents, 
 and proposed AHDA*, which used an abstraction-based hash function originally designed for use with PSDD \cite{zhou2007parallel} and PBNF \cite{burnslrz10}.
Abstraction-based work distribution achieves low communication overhead, but at the cost of high search overhead.


In this paper, we investigate node distribution methods for HDA*.
We start by reviewing previous approaches to work distribution in parallel best-first search, including the HDA* framework (Section \ref{sec:background}). 
Then, in Section \ref{sec:analysis-of-parallel-overheads}, we present an in-depth investigation of  parallel overheads in state-of-the-art parallel best-first search methods.
We begin by investigating {\it why} search overhead occurs on parallel best-first search by analyzing how node expansion order in HDA* diverges from that of A*.
If the expansion order of a parallel search algorithm is identical to A*, there is no search overhead, so divergence in expansion order is a useful indicator for understanding search overhead.
We show that although HDA* incurs some search overhead due to load imbalance and startup overhead, HDA* using the Zobrist hash function incurs significantly less search overhead than other methods.
However, while HDA* with Zobrist hashing successfully achieves low search overhead, 
we show that communication overhead is actually as important as search overhead in determining the overall efficiency for parallel search, and Zobrist hashing results in very high communications overhead, resulting in poor performance on the grid pathfinding problem.

Next, in Section \ref{sec:azh}, we propose Abstract Zobrist hashing (AZH), which achieves both low search overhead and communication overhead by incorporating the strengths of both Zobrist hashing and abstraction. 
While the Zobrist hash value of a state is computed by applying an incremental hash function to the set of features of a state,
AZH first applies a feature projection function mapping features to  abstract features, and the Zobrist hash value of the abstract features (instead of the raw features) is computed.
We show that on the 24-puzzle, 15-puzzle, and multiple sequence problem, AZH with hand-crafted, domain-specific feature projection function 
significantly outperform previous methods on a multicore machine with up to 16 cores.

Then, we propose a domain-independent method to automatically generate an efficient feature projection function for AZH framework.
We first show that a work distribution can be modeled as graph partitioning (Section \ref{sec:implicit-graph-partitioning}). However, standard graph partitioning techniques for workload distribution in scientific computation are inapplicable to heuristic search because the state space is defined implicitly.
Then, in Section \ref{sec:graph-partitioning-based}, we propose GRAZHDA*, a new domain-independent method for automatically generating a work distribution function, which, instead of partitioning the actual state space graph (which is impractical), generates an approximation by partitioning a {\it domain transition graph}. We then propose a {\it sparsity}-based objective function for  GRAZHDA*, and 
experimentally show that GRAZHDA* using the  sparsity objective function outperforms all previous variants of HDA* on domain-independent planning, using experiments run on a 48-core cluster as well as a cloud-based cluster with 128 cores.
We conclude the paper with a summary of our results and directions for future work (Section \ref{sec:conclusions}).

Portions of this work have been previously presented in two conference papers \cite{jinnai2016structured,jinnai2016automated}, corresponding to Section \ref{sec:azh}, as well as parts of Section \ref{sec:background}. 
The two major, new contributions of this journal paper are: (1) GRAZHDA* which defines  an objective function that can be used to control the tradeoff between communications and search overhead (Section \ref{sec:implicit-graph-partitioning}, \ref{sec:graph-partitioning-based}), and (2) analysis of parallel overheads in HDA*, as well as a revisited comparison of HDA* with PBNF (Section \ref{sec:analysis-of-parallel-overheads} and \ref{sec:azh-order}).
All of the experimental data in Section \ref{sec:experiments-planning} is new -- the experimental data for OZHDA*, GAZHDA*, DAHDA*, and FAZHDA* use the newer CGL-B merge \& shrink heuristic function \cite{helmert2014merge}, in contrast to the  previous conference paper \cite{jinnai2016structured} which used the older LFPA merge\&shrink  heuristic \cite{helmert2007flexible}.
In addition, while the conference papers were limited to single multicore machines  \cite{jinnai2016structured} and clusters with up to 48 cores \cite{jinnai2016automated}, this paper includes an evaluation of the new HDA* variants on a 128 core cloud environment (Section \ref{sec:cloud-results}). 
Finally, Table \ref{dahda} in Appendix \apDAHDA{} shows new experimental results comparing DAHDA* vs. AHDA* \cite{burnslrz10} which were not included in the conference paper which introduced DAHDA* \cite{jinnai2016automated}.

\section{Preliminaries and Background}
\label{sec:background}

In this section, we first define the three major classes of overheads that pose a challenge for parallel search  (Section \ref{sec:parallel-overheads}). We then survey parallel best-first search algorithms (Section \ref{sec:parallel-best-first-search}) and review the HDA* framework (Section \ref{sec:hda*}). We then review the two previous approaches which have been proposed for the HDA* framework, Zobrist hashing (Section \ref{sec:zhda}) and abstraction (Section \ref{sec:ahda}).

\subsection{Parallel Overheads}
\label{sec:parallel-overheads}

Although an ideal parallel best-first search algorithm would achieve an $n$-fold speedup on $n$ threads, 
several overheads can prevent parallel search from achieving linear speedup.

\noindent {\bf Communication Overhead (CO)}: \footnote{In this paper, CO stands for communication overhead, not coordination overhead.} 
Communication overhead refers to the cost of exchanging information between threads.
In this paper we define communication overhead as the ratio of nodes transferred to other threads:
	$CO := \frac{\text{\# nodes sent to other threads}}{\text{\# nodes generated}}$.
CO is detrimental to performance because of delays due to message transfers (e.g., network communications), as well as access to data structures such as  message queues. 
In general, CO increases with the number of threads.
If nodes are assigned randomly to the threads, CO will be proportional to $1-\frac{1}{\#thread}$. 

\noindent {\bf Search Overhead  (SO):}
Parallel search usually expands more nodes than sequential A*. 
In this paper we define search overhead as 
	$
	SO := \frac{\text{\# nodes expanded in parallel}}{\text{\# nodes expanded in sequential search}} - 1
	$.
SO can arise due to inefficient load balance (LB), where we define load balance as $LB := \frac{\text{Maximum number of nodes assigned to a thread}}{\text{Average number of nodes assigned to a thread}}$.
If load balance is poor, a thread which is assigned more nodes than others will become a bottleneck -- other threads spend their time expanding less promising nodes, resulting in search overhead.
Search overhead is not only critical to the walltime performance, but also to the space efficiency.
Even in distributed memory environment, RAM per core is still an important issue to consider.



\noindent {\bf Coordination (Synchronization) Overhead:}
In parallel search, coordination overhead occurs when a thread has to wait in idle for an operation of other threads. 
Even when a parallel search itself does not require synchronization, coordination overhead can be incurred due to contention for the memory bus \cite{burnslrz10,kishimotofb13}.

There is a fundamental trade-off between CO and SO.
Increasing communication can reduce search overhead at the cost of communication overhead, and vice-versa.

\subsection{Parallel Best-First Search Algorithms}
\label{sec:parallel-best-first-search}
\begin{figure}[htb]
	\centering
	\includegraphics[width=1.0\linewidth]{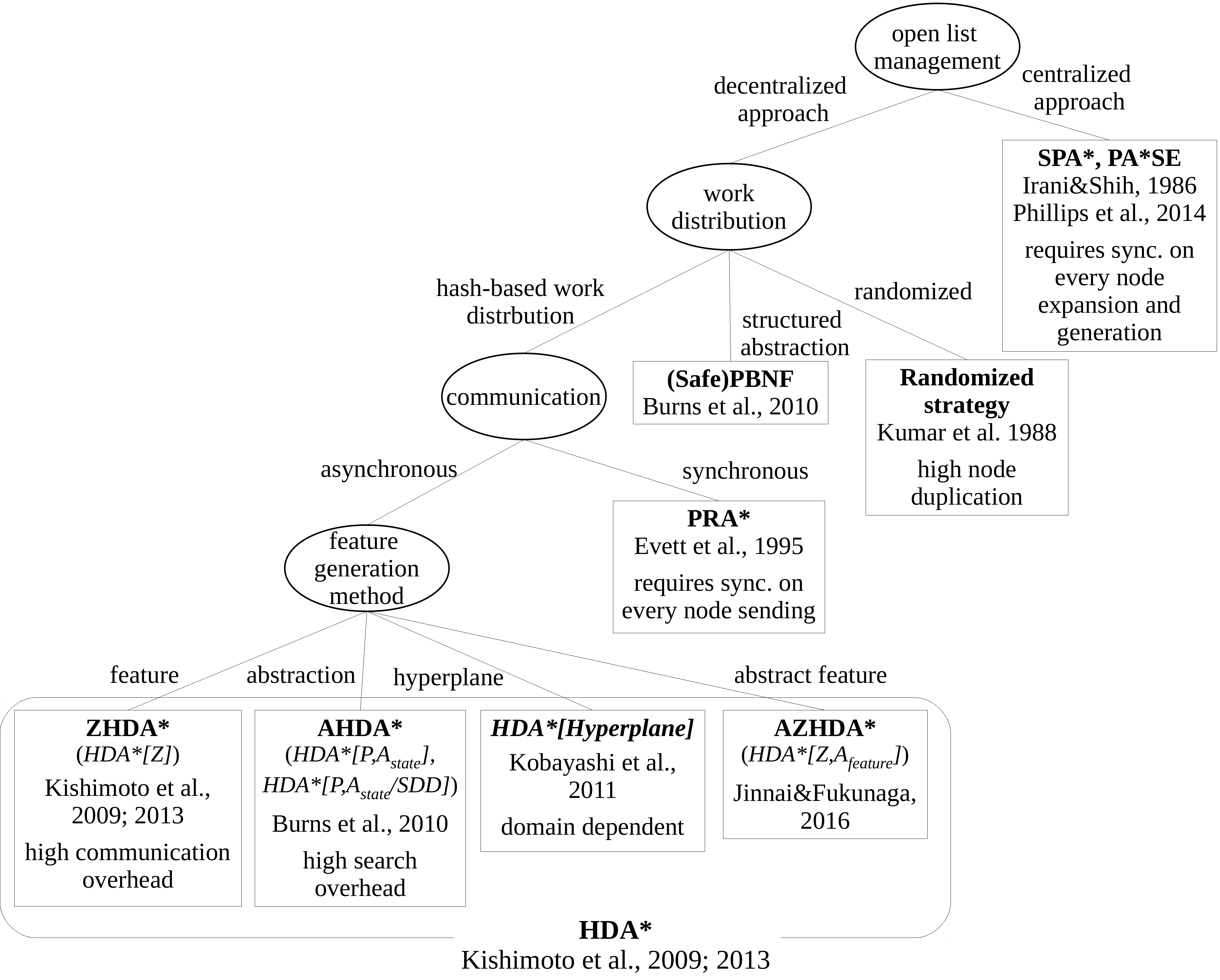}
	\caption{Classification of parallel best-first searches.}%
	\label{fig:parallel-best-first}
\end{figure}

The key to achieving a good speedup in parallel best-first search is to minimize communication, search, and coordination overheads.
In this section, we survey previous approaches. See the work of \citeauthor{fukunaga2017survey} \citeyear{fukunaga2017survey} for a recent, more in-depth survey.
Figure \ref{fig:parallel-best-first} presents a visual classification of the previous approaches which are discussed below.

Simple Parallel A* (SPA*) \cite{IraniS86} is a straightforward parallelization of A* which uses a single, shared open list. 
Since worker processes always expand the best node from the shared open list, this minimizes search overhead by eliminating the burst effect (Section \ref{sec:burst}). However, node reexpansions are possible in SPA* because (as with most other parallel A* variants including HDA*) SPA* does not guarantee that a state has an optimal $g$-value when expanded.
\citeauthor{PhillipsLK14} \citeyear{PhillipsLK14} have proposed PA*SE, a mechanism for reducing node reexpansions in SPA* which only expands nodes when their $g$-values are optimal, ensuring that nodes are not reexpanded.

\citeauthor{Kumar1988parallel} \citeyear{Kumar1988parallel} identified two classes of approaches to open list management in parallel A*.
SPA* and its variants are instances of a {\it centralized approach} which shares a single open list among all processes.
However, concurrent access to the shared open list becomes a bottleneck and inherently limits the scalability of this approach unless the cost of expanding each node is extremely expensive, even if lock-free data structures are used  \cite{burnslrz10}.
A {\it decentralized approach} addresses this bottleneck  by assigning each process to a separate open list. Each process executes a best-first search using its own local open list. While decentralized approaches eliminate coordination overhead incurred by a shared open list, load balancing becomes a problem.

There are several approaches to load balancing in decentralized best-first search.
The simplest approach is a randomized strategy which sends generated states to a randomly selected neighbor processes  \cite{Kumar1988parallel}. The problem with this  strategy is that duplicate nodes are not detected unless they are fortuitously sent to the same process, which can result in a tremendous amount of search overhead due to nodes which are  redundantly expanded by multiple processors.

Parallel Retracting A* (PRA*) \cite{evett1995massively} uses a {\it hash-based work distribution} to address simultaneously address both load balancing and duplicate detection. In PRA*, each process owns its local open and closed list. A global hash function maps each state to exactly one process which owns the state. 
Thus, hash-based work distribution solves the problem of duplicate detection and elimination, because each state has exactly one owner. 
When generating a state, PRA* distributes it to the corresponding owner synchronously. However, synchronous node sending was shown to degrade performance on domains with fast node expansion, such as grid pathfinding and sliding-tile puzzle \cite{burnslrz10}.


Transposition-Table Driven Work Scheduling (TDS) \cite{romein1999transposition} is a distributed memory, parallel IDA* with hash-based work distribution. 
In contrast to  PRA*, TDS sends a state to its owner process asynchronously.

An alternate approach for load balancing,
 which originated in a line work for using multiple processes in external memory search \cite{korf2005large,niewiadomski06,JabbarE06}, 
 is based on {\it structured abstraction}. 
Given a state space graph and a projection function, an abstract state graph is (implicitly) generated by projecting states from the original state space graph into abstract nodes.
For example, an abstract space for the sliding-tile puzzle domain can be created by projecting all nodes with the blank tile at position $b$ to the same abstract state.
While the use of abstractions as the basis for heuristic functions has a long history \cite{Pearl84}, 
the use of abstractions as a mechanism for partitioning search states originated in Structured Duplicate Detection (SDD), an external memory search which stores explored states on disk \cite{zhou2004structured}.
In SDD, an $n$-block is defined as the set of all nodes which map to the same abstract node.
SDD uses $n$-blocks to provide a solution to duplicate detection.
For any node $n$ which belongs to  $n$-block $B$, the {\it duplicate detection scope} of $n$ is defined as the set of $n$-blocks which can possibly contain duplicates of $n$, and duplicate checks can be restricted to the duplication detection scope, thereby avoiding the need to look for a duplicate of $n$ outside this scope.
SDD exploits this property for external memory search by expanding nodes within a single $n$-block $B$ at a time and keeping the duplicate detection scope of the nodes in $B$ in RAM, avoiding costly I/O.
Unlike stack-slicing, which requires a levelled search space, SDD is applicable to any state-space search problem. 
%
Parallel Structured Duplicate Detection (PSDD) is a parallel search algorithm which exploits $n$-blocks to address both  synchronization overhead and communication overhead \cite{zhou2007parallel}. Each processor is exclusively assigned to an $n$-block and its neighboring $n$-blocks (which are the duplication detection scopes). By exclusively assigning $n$-blocks with disjoint duplicate detection scopes to each processor, synchronization during duplicate detection is eliminated.
While PSDD used disjoint duplicate detection scopes to parallelize breadth-first heuristic search \cite{zhou2006breadth}, 
Parallel Best-NBlocks First (PBNF) \cite{burnslrz10} extends PSDD to best-first search on multicore machine by ensuring that $n$-blocks with the best current $f$-values are assigned to processors.
Since livelock is possible in PBNF  on domains with infinite state spaces, \citeauthor{burnslrz10} \citeyear{burnslrz10} proposed  SafePBNF,  a livelock-free version of PBNF. 
\citeauthor{burnslrz10} \citeyear{burnslrz10} also proposed AHDA*, a variant of HDA* which uses an abstraction-based node distribution function. AHDA* is described below in Section \ref{sec:ahda}.

Efficient abstractions can also be generated by exploiting prior knowledge of the structure of the state-space and/or machines on which search is performed.
Stack-slicing projects states to their path costs to achieve efficient communication in depth-first search \cite{holzmann2008stack}, and  is useful in domains with {\it levelled graphs}, where each state can be reached only by a unique path cost, such as model checking \cite{holzmannb07} (thus enabling dupicate detection).
LOcal HAshing of nodes (LOHA) applies path cost-based partitioning in A* search to reduce the number of inter-node communication in a hypercube multiprocessor \cite{mahapatra1997scalable}.


\subsection{Hash Distributed A* (HDA*)}
\label{sec:hda*}

Hash Distributed A* (HDA*) \cite{kishimotofb13} is a parallel A* algorithm 
which incorporates the idea of hash-based work distribution from  PRA* \cite{evett1995massively} and asynchronous communication from TDS \cite{romein1999transposition}.
In HDA*, 
each processor has its own open/closed lists.
A global hash function assigns a unique owner thread to every search node.
Each thread $T$ repeatedly executes the following: 
\begin{enumerate}
	\item 
           $T$ checks its message queue if any new nodes are in. For all new nodes $n$ in $T$'s message queue, if it is not in the open list (not a duplicate), put $n$ in the open list. 
	\item 
           Expand node $n$ with the highest priority in the open list. For every generated node $c$, compute hash value $H(c)$, and send $c$ to the thread that owns $H(c)$. 
\end{enumerate}

HDA* has two features which make it attractive as a parallel search algorithm. 
First, there is little coordination overhead because HDA* communicates asynchronously, and 
locks for an access to shared open/closed lists are not required because each thread has its own local open/closed list.
Second, the work distribution mechanism is simple, requiring only a hash function.
However, the effect of the hash function was not evaluated empirically, and the importance of the choice of hash function may not have been fully understood or appreciated --
at least one subsequent work which evaluated HDA* used an implementation of HDA* which failed to achieve uniform distribution of the nodes (see Section \ref{sec:hdavspbnf}).

\subsection{Zobrist Hashing (\ZHDA{}) and Operator-Based Zobrist Hashing (\OZHDA{})} 
\label{sec:zhda}
Since the work distribution in HDA* is completely determined by a global hash function, the choice of the hash function is crucial to its performance.
\citeauthor{kishimotofb09}  \citeyear{kishimotofb09,kishimotofb13} noted that it is  desirable to use a hash function that uniformly distributed nodes among processors, and used the Zobrist hash function \citeyear{Zobrist1970}, \index{hashing, Zobrist}
described below. 
The Zobrist hash value of a state $s$, $Z(s)$, is calculated as follows. For simplicity, assume that $s$ is represented as an array of $n$ propositions, $s = (x_0, x_1,..., x_n)$. Let $R$ be a table containing preinitialized random bit strings (Algorithm \ref{alg:init-zobrist-hashing}). 
\newcommand{\xor}{{\mbox{\textit{xor}}}}
\begin{equation}
\label{eq:zobrist}
	Z(s) := R[x_{0}]\; \xor \; R[x_{1}]\; \xor\; \cdots\; \xor\; R[x_{n}]%
\end{equation}

\begin{algorithm}
	\Input{$s = (x_0, x_1,...,x_n)$}
	$hash \leftarrow 0$\;
	\For {each $x_i \in s$} {
		$hash \leftarrow hash \; xor \; R[x_i]$\;
	}
	{\bf Return} $hash$\;
	\caption{\ZHDA{}}
	\label{alg:zobrist-hashing}
\end{algorithm}

\begin{algorithm}
	\Input{$F$: a set of features}
	\For {each $x \in F$} {
		$R[x] \leftarrow random()$\;
	}
	{\bf Return} $R$
	\caption{Initialize \ZHDA{}}
	\label{alg:init-zobrist-hashing}
\end{algorithm}

In the rest of the paper, we refer to the original version of HDA* in \citeauthor{kishimotofb09} \citeyear{kishimotofb09,kishimotofb13}, which  used Zobrist hashing, as ZHDA* or \ZHDA{}. 

Zobrist hashing seeks to distribute nodes uniformly among all processors, without any consideration of the neighborhood structure of the search space graph. 
As a consequence,
communication overhead is high.
Assume an ideal implementation that assigns nodes uniformly among threads.
Every generated node is sent to another thread with probability $1-\frac{1}{\#threads}$.
Therefore, with 16 threads, $>90\%$ of the nodes are sent to other threads,
so
communication costs are incurred for the vast majority of node generations.

Operator-based Zobrist hashing (OZHDA*) \cite{jinnai2016automated} 
partially addresses this problem by manipulating the random bit strings in $R$, the  table used to compute Zobrist hash values, such that for some selected states $S$, there are some operators $A(s)$ for  $s \in S$ such that the successors of $s$ that are generated when $a \in A(s)$
is applied to $s$ are guaranteed to have the same Zobrist hash value as $s$, which ensures that they are assigned to the same processor as $s$.
\citeauthor{jinnai2016automated} \citeyear{jinnai2016automated} showed that OZHDA* significantly reduces  communication overhead compared to Zobrist hashing. However, this may result in increased search overhead compared to \ZHDA, and it is not clear whether the extent of the increased search overhead in OZHDA* could be predicted \emph{a priori}.



\subsection{Abstraction (\AHDA{})} 
\label{sec:ahda}


In order to minimize communication overhead in HDA*, 
\citeauthor{burnslrz10}  \citeyear{burnslrz10} proposed AHDA*, which uses \emph{abstraction} based node assignment. 
The abstraction strategy in AHDA* applies the state space partitioning technique used in PBNF \cite{burnslrz10} and PSDD \cite{zhou2007parallel}, which projects nodes in the state space to \emph{abstract states}. 
After mapping states to abstract states, the AHDA* implementation in \citeauthor{burnslrz10} \citeyear{burnslrz10} 
assigns abstract states to processors using perfect hashing and a modulus operator.




Thus, nodes that are projected to the same abstract state are assigned to the same thread. 
If the abstraction function is defined so that children of node $n$ are usually in the same abstract state as $n$, then communication overhead is minimized.
The drawback of this method is that it focuses solely on minimizing communication overhead, and there is no mechanism for equalizing load balance, which can lead to high search overhead.

HDA* with abstraction can be characterized by two parameters which determine its behavior -- a hashing strategy and an abstraction strategy.
The AHDA* implementation in \citeauthor{burnslrz10} \citeyear{burnslrz10} implemented the hashing strategy using perfect hashing and a modulus operator, and an abstraction strategy following the construction for SDD \cite{Zhou2006} (for domain-independent planning), or a hand-crafted abstraction (for the sliding-tile puzzle and grid path-finding domains). 
Note that an abstraction strategy can itself be seen as a type of hashing strategy, but in this paper, we make the distinction between  the method used to project states onto some cluster of states (abstraction) and methods which are used to map states (or abstract states) to processors (hashing).



\citeauthor{jinnai2016automated} \citeyear{jinnai2016automated} showed that AHDA* with a static $N_{max}$ threshold performed poorly for a benchmark set with varying difficulty because a fixed size abstract graph results in very poor load balance, and proposed Dynamic AHDA* (DAHDA*), which dynamically sets the size of the abstract graph according to the number of features (the state space size is exponential in the number of features). We evaluate DAHDA* in detail in Appendix \apDAHDA{}.

\subsection{Classification of HDA* Variants and a Uniform Notation for HDA* Variants ($\mathit{H\!D\!A\!^*\![hash,abstraction]}$)}

At least 12 variants of HDA* have previously been proposed and evaluated in the literature.
Each variant of HDA* can be characterized according to two parameters: a  hashing strategy used (e.g., Zobrist hashing or perfect hashing), and an abstraction strategy (which corresponds to the strategy used to cluster states or features before the hashing, e.g., state projection based on SDD). 

Table \ref{tab:list-of-all-hda*-variants} shows all of the HDA* variants that are discussed in this paper.
In order to be able to clearly distinguish among these variants, we use the notation $\mathit{H\!D\!A\!^*\![hash,abstraction]}$ throughout this paper, where ``$\mathit{hash}$'' is the hashing strategy of HDA* and ``$\mathit{abstraction}$'' is the abstraction strategy. Variants that do not use any abstraction strategy are denoted by $\mathit{H\!D\!A\!^*\![hash]}$.
In cases where the unified notation is lengthy, we use the abbreviated name in the text (e.g., ``FAZHDA*'' for \FAZHDA).

For example, we denote AHDA* \cite{burnslrz10} using a perfect hashing and a hand-crafted abstraction as \AHDAP{}, and AHDA* using a perfect hashing and a SDD abstraction as \AHDAPSSD{}.
We denote HDA* with Zobrist hashing without any clustering (i.e., the original version of HDA* in \citeauthor{kishimotofb09}  \citeyearR{kishimotofb09,kishimotofb13}) as \ZHDA{}.
We denote  OZHDA* as \OZHDA{}, where $Z_{operator}$ stands for Zobrist hashing using operator-based initialization.

\begin{table}[htb]
	\caption{Overview of all HDA* variants mentioned in this paper}
	\label{tab:list-of-all-hda*-variants}
	\centering
\begin{small}

	\begin{tabularx}{\textwidth}{r|p{0.35\textwidth}|p{0.30\textwidth}} \hline
                \multicolumn{3}{l} {Algorithms Evaluated With Domain-Specific Solvers Using Domain-Specific, Feature Generation Techniques}\\
                \hline
		method &    & First proposed in  \\ \hline

		\ZHDA{} &    ZHDA* : Original version, using Zobrist hashing   [Sec \ref{sec:zhda}]  & \cite{kishimotofb09}\\
		\PHDA{} &       Perfect hashing. [Sec \ref{sec:experiments-order}] & \cite{burnslrz10}\\
		\AHDAP{} &      AHDA* with perfect hashing and state-based abstraction  [Sec \ref{sec:ahda}] & \cite{burnslrz10}\\
		\AHDAZ{} &      AHDA* with Zobrist hashing and state-based abstraction [Sec \ref{sec:ahda}] & {\footnotesize trivial variant of \AHDAP}\\
		\HWD{} &        Hyperplane work distribution  (Sec \ref{sec:msa}) & \cite{Kobayashi2011evaluations}\\
		\AZHDA{} &   Abstract Zobrist Hashing (feature abstraction)  [Sec \ref{sec:azh}] & \cite{jinnai2016structured}\\
		\multicolumn{3}{l} {}\\
                \hline
                \multicolumn{3}{l} {Automated, Domain-Independent Feature Generation Methods Implemented for Parallelized, Classical Planner}\\
                \hline
		method &    & First proposed in  \\ \hline
		\ZHDA{} &  Original version, using Zobrist hashing  [Sec \ref{sec:domain-independent}] & \cite{kishimotofb09}\\
		\AHDAZSSD{} &  AHDA* with Zobrist hashing and SDD-based abstraction [Sec \ref{sec:zhda}] & {\footnotesize trivial variant of \newline \AHDAPSSD, which was ussed for classical planning in \cite{burnslrz10};   uses Zobrist-based hashing instead of perfect hashing.} \\
		\DAHDA{}  &   DAHDA*: Dynamic AHDA*  [Sec \ref{sec:ahda} \& {\footnotesize Append. \apDAHDA{}}]  & \cite{jinnai2016automated}\\
		 \GAZHDA{} & GAZHDA*: Greedy Abstract Feature Generation   [Sec \ref{sec:greedyafg}] & \cite{jinnai2016structured}\\
		 \FAZHDA{} & FAZHDA*: Fluency-Dependent Abstract Feature Generation  [Sec \ref{sec:fluencyafg}] & \cite{jinnai2016automated}\\
		 \OZHDA{}  &  OZHDA*: Operator-based Zobrist  [Sec \ref{sec:zhda}] & \cite{jinnai2016automated}\\

		 \GRAZHDAS{} & GRAZHDA*/sparsity: Graph partitioning-based Abstract Feature Generation using the sparsity cut objective [Sec \ref{sec:graph-partitioning-based}] & This paper\\

	\end{tabularx}
\end{small}
\end{table}

\section{Analysis of Parallel Overheads in Multicore Best-First Search} 
\label{sec:analysis-of-parallel-overheads}

As discussed in Section \ref{sec:parallel-overheads}, there are three broad classes of parallel overheads in parallel search: search overhead (SO), communications overhead (CO), and coordination (synchronization) overhead. 
Since state-of-the-art parallel search algorithms such as HDA* and PBNF have successfully eliminated coordination overhead, the remaining overheads are SO and CO. 
Previous work has focused on evaluating SO quantitatively because SO is a fundamental algorithmic overhead, whereas CO is heavily influenced by the parallel machine environment and is therefore more difficult to evaluate.
Thus, in this section, we first evaluate the SO of \ZHDA{} and SafePBNF.

\citeauthor{kishimotofb13} \citeyear{kishimotofb13} previously analyzed search overhead for \ZHDA{}. They measured 
$R_<$, $R_=$, and $R_>$, the fraction of expanded nodes with $f < f^*$, $f = f^*$, and $f> f^*$ (where $f^*$ is optimal cost), respectively. They also measured $R_r$, the fraction of nodes which were reexpanded.
All admissible search algorithms must expand all nodes with $f < f^*$ in order to guarantee optimality. In addition, some of the nodes with  $f = f^*$ nodes are expanded.  
Thus, SO is the sum of $R_>$, $R_r$, and some fraction of $R_=$. 
These metrics enable estimating the SO on instances which are too hard to solve in sequential A*.
\citeauthor{burnslrz10} \citeyear{burnslrz10} analyzed the quality of nodes expanded by SafePBNF and \AHDA{} by comparing the number of nodes expanded according to their $f$-values, and showed that \AHDA{} expands nodes with larger $f$-value (lower quality nodes) compared to SafePBNF.

While these previous works measure the amount of search overhead, they 
do not provide a quantitative explanation for {\it why} such overheads occur.
In addition, previous work has not directly compared \ZHDA{} and SafePBNF, as \citeauthor{burnslrz10} \citeyear{burnslrz10} compared SafePBNF to |\AHDA{} and another variation of HDA* using a suboptimal hash function, which we refer to as \PHDA{} in this paper.

In this section, we propose a method to analyze SO and explain search overhead  in HDA* and SafePBNF. In light of the observation of this analysis, we revisit the comparison of HDA* vs. SafePBNF on sliding-tile puzzle and grid path-finding. We then analyze the impact communications overhead has on overall performance.


\subsection{Search Overhead and the Order of Node Expansion on Combinatorial Search} 
\label{sec:so}

\begin{figure}[htb]
	\centering
	\subfloat[\ZHDA{} on an easy instance with 2 threads. \ZHDA{} slightly diverges from A* expansion order with 2 threads (band effect). 
] {\label{fig:order_2_threads}{\includegraphics[width=0.60\linewidth]{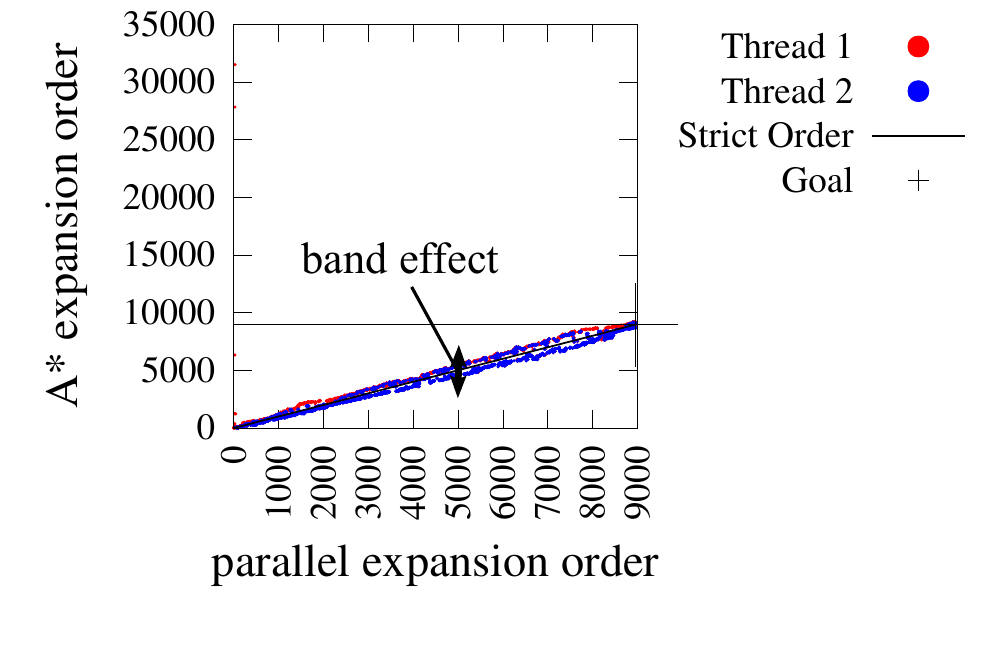} }}

	\subfloat[\ZHDA{} on an easy instance with 4 threads. At the beginning of the search, \ZHDA{} significantly diverges from A* expansion order, which mostly results in search overhead (burst effect). The band effect is larger with 4 threads than with 2 threads.] {\label{fig:order_4_threads}{\includegraphics[width=0.60\linewidth]{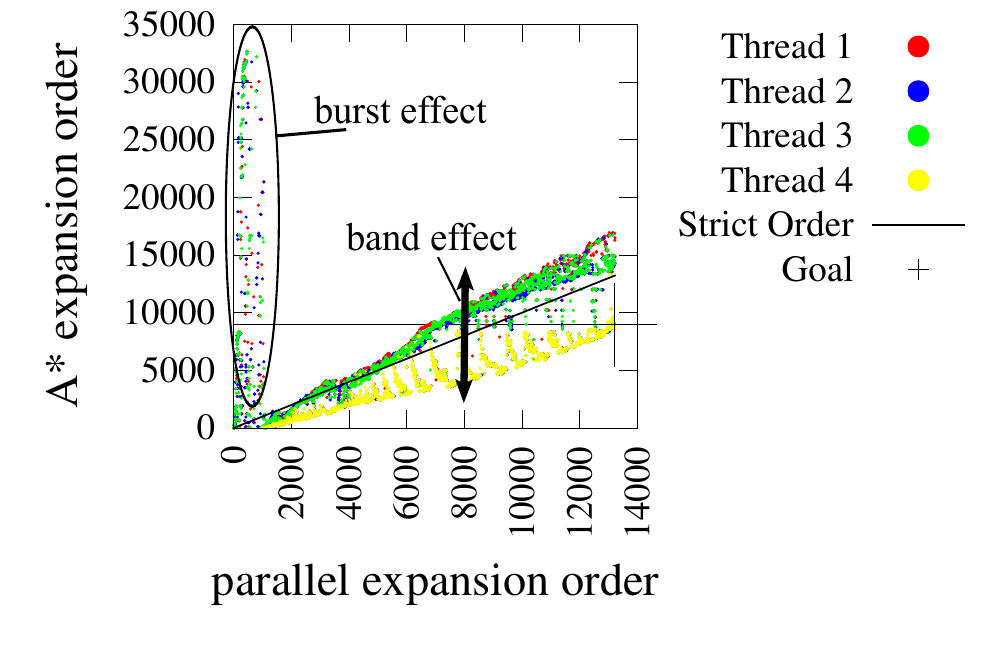} }}

	\caption{Illustration of Band Effect: Comparison of node expansion order on an {\bf easy instance of the 15-Puzzle}. The vertical axis represents the order in which state $s$ is expanded by parallel search, and the horizontal axis represents the order in which $s$ is expanded by A*. The line $y = x$ corresponds to an ideal, strict A* ordering in which the parallel expansion order is identical to the A* expansion order. The cross marks (``Goal'')  represents the (optimal) solution, and the vertical line from the goal shows the total number of node expansions by A*. Thus, all nodes above this line result in SO.}
	\label{fig:order_nthreads}
	\captionlistentry[experiment]{ExpansionOrder, Figure \ref{fig:order_nthreads}, supermicro, 15puzzle:random100:heap, pthread, newmaterial}
\end{figure}

Consider the global order in which states are expanded by a parallel search algorithm. If a parallel A* algorithm expands states in exactly the same order as A*, then by definition, there is no search overhead.
We ran A* and \ZHDA{} on 100 randomly generated instances of the 15-puzzle on \lucy, using a 15-puzzle solver with Manhattan distance heursitic based on the solver code used in the work of \citeauthor{burnslrz10} \citeyear{burnslrz10}. We recorded the order in which states were expanded.
We used a random generator by Burns to generate random instances\footnote{We obtained the instance generator from  \texttt{https://github.com/eaburns/pbnf/tree/master/tile-gen}}.
The results from runs on 2 representative instances (one ``easy'' instance which A* solves after 8966 expansions, and one ``difficult'' instance which A* solves after 4265772 expansions), are shown in Figures \ref{fig:order_nthreads}, \ref{fig:order_easy} and \ref{fig:order_difficult}. The results on the other difficult/easy problems were similar to these representative instances -- aggregate results are presented in Sections \ref{sec:node-reexpansions}-\ref{sec:experiments-order}.

\newcommand{\osize}{0.69}

\newcommand{\ttsize}{0.30}
\newcommand{\tspace}{6pt}

\begin{figure}[htbp]
	\centering
	\subfloat[\ZHDA{} on an easy instance with 8 threads. Both band and burst effects are more significant than with 4 threads.]
	{\label{fig:order_z_easy}{  \includegraphics[width=\ttsize\linewidth]{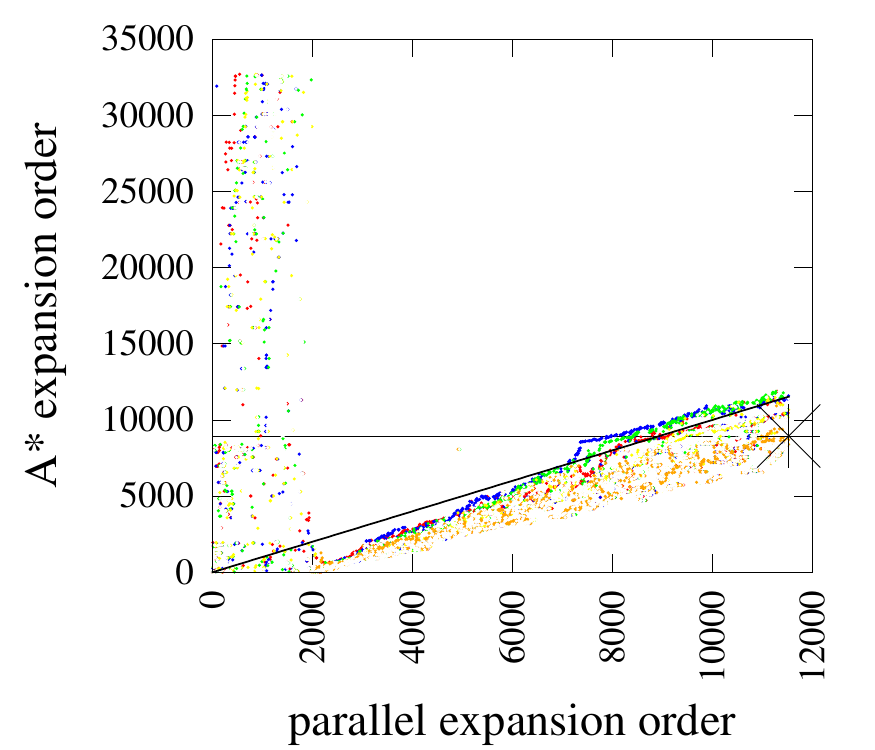} }} \hspace{\tspace}
	\subfloat[\AHDA{} on an easy instance with 8 threads. \AHDA{} has a significantly bigger band than \ZHDA{}. ] 
	{\label{fig:order_a_easy}{ \includegraphics[width=\ttsize\linewidth]{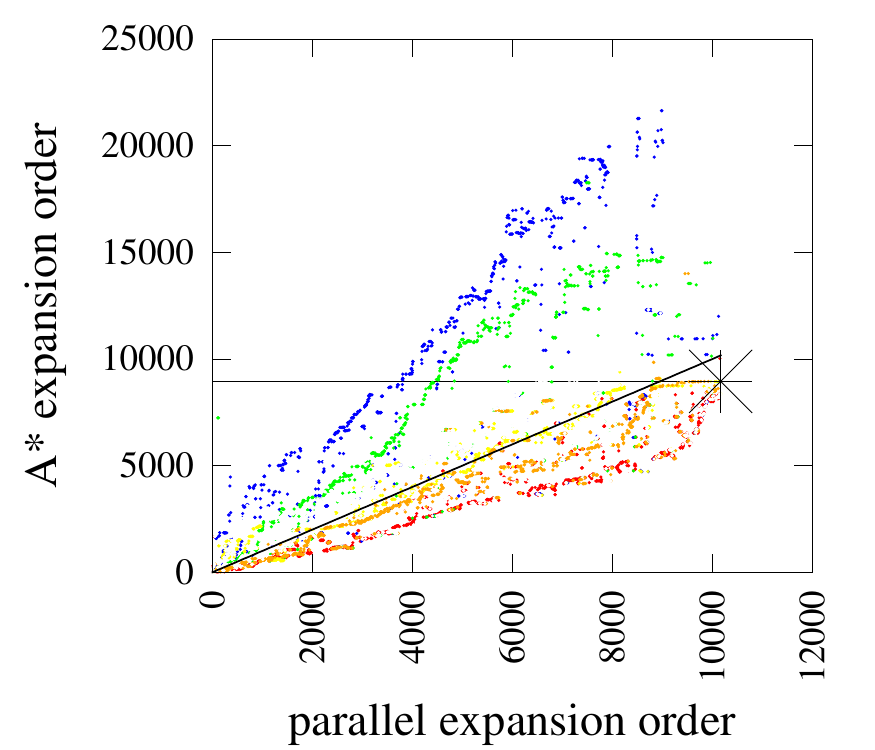} }} \hspace{\tspace}
	\subfloat[SafePBNF on an easy instance with 8 threads. As threads in SafePBNF requires exclusive access to nblocks, the expansion order differs significantly from A* (and HDA* variants).] 
	{\label{fig:order_safepbnf_easy}{ \includegraphics[width=\ttsize\linewidth]{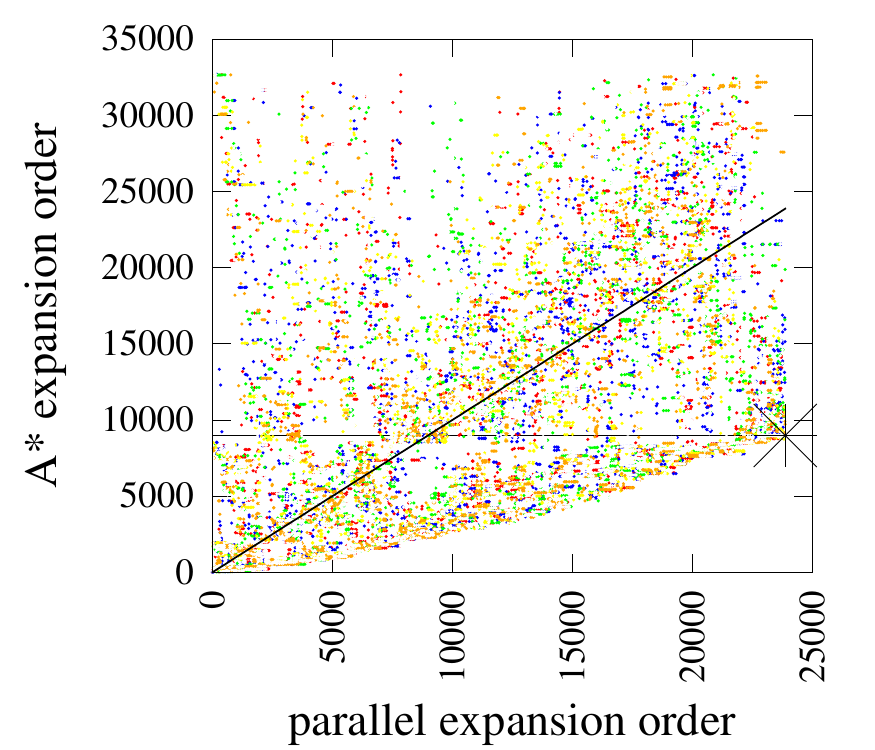} }}

	\subfloat[\ZHDA{} on an easy instance with 8 threads with artificially slowed expansion rate. The band effect remains clear, indicating that the band effect is not an accidental overhead cause by communications or lock contention.] 
	{ \label{fig:order_z_slow_easy}{\includegraphics[width=\ttsize\linewidth]{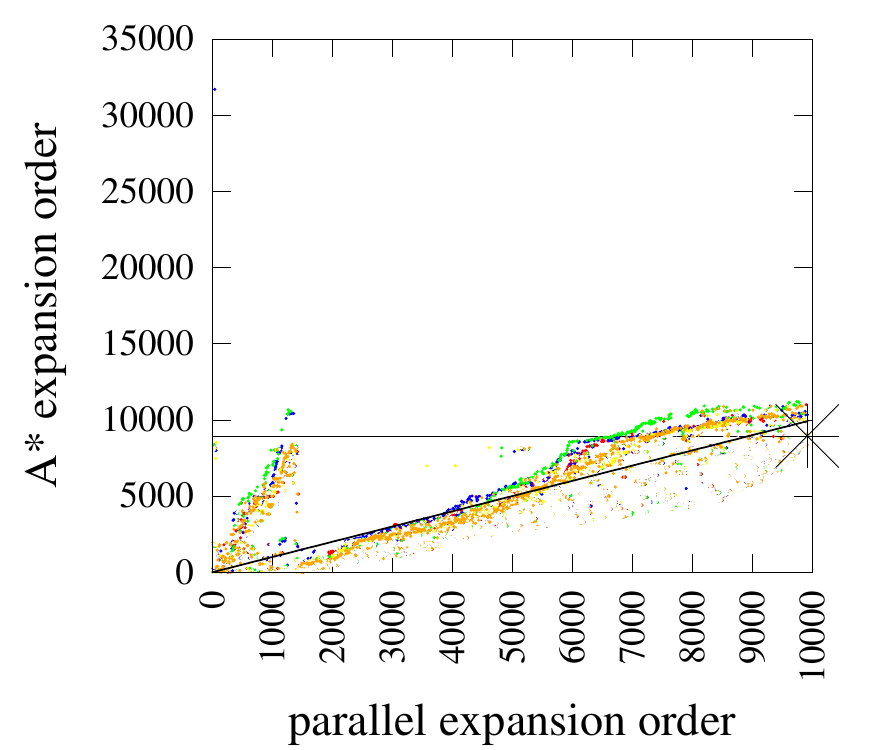} }}  \hspace{\tspace}
	\subfloat[\PHDA{} on an easy instance with 8 threads. \PHDA{} has a significantly bigger band than other methods, and many threads are expanding unpromising (high $f$-value) nodes. As a result, \PHDA{} expands $> 25000$ nodes to solve this instance, which A* solves with 8966 expansions.] 
	{\label{fig:order_p_easy}{\includegraphics[width=\ttsize\linewidth]{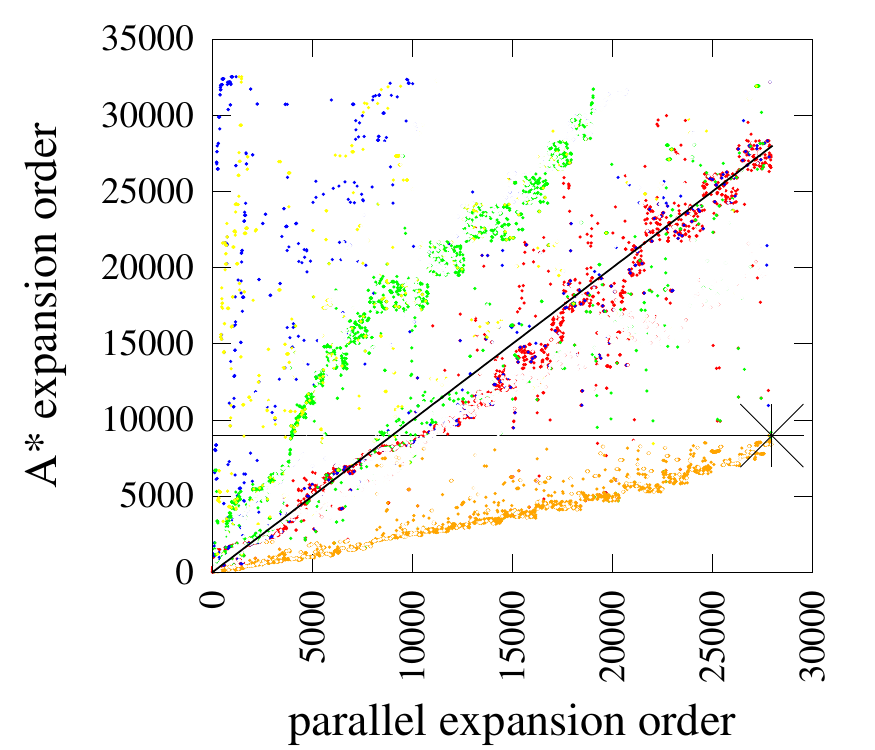} }} \hspace{\tspace}
	\subfloat[\ZHDA{} using FIFO tiebreaking on an easy instance with 8 threads (vs. A* using FIFO tiebreaking).] 
	{\label{fig:order_z_fifo_easy}{\includegraphics[width=\ttsize\linewidth]{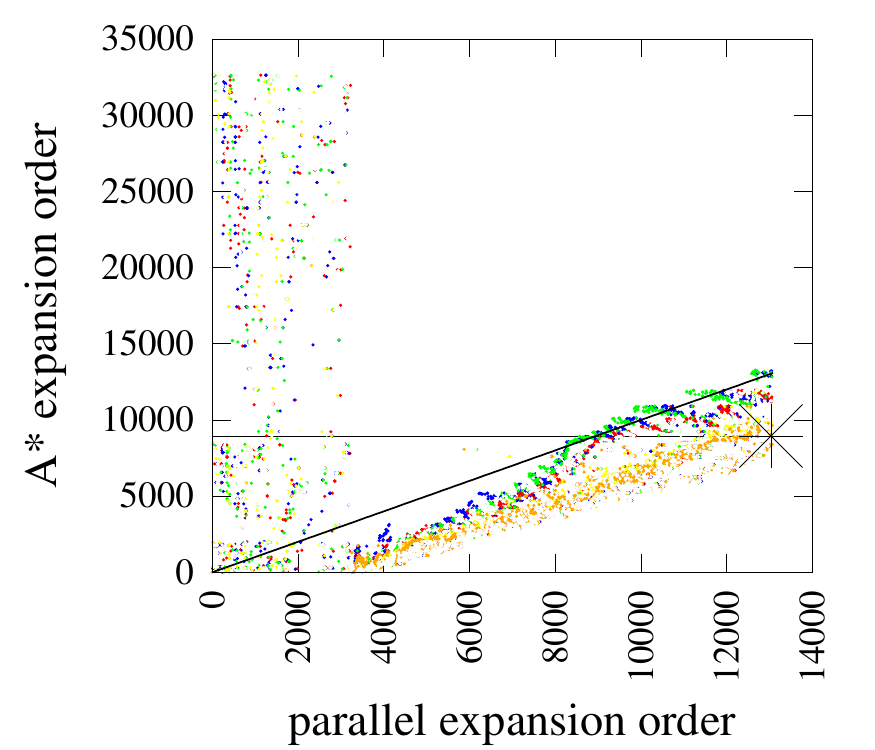} }}

	\subfloat {{\includegraphics[width=0.1\linewidth]{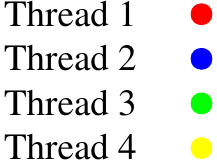} }}
	\subfloat{{\includegraphics[width=0.1\linewidth]{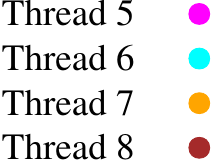} }}
	\subfloat{{\includegraphics[width=0.12\linewidth]{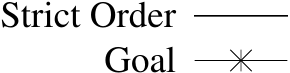} }} 

	\caption{Comparison of parallel vs. sequential node expansion order on an {\bf easy instance of the 15-Puzzle} with 8 threads. }
	\label{fig:order_easy}
\end{figure}

\begin{figure}[htbp]
	\centering
	\subfloat[\ZHDA{} on a difficult instance with 8 threads. As the instance is difficult enough, the relative significance of burst effect becomes negligible.]
	{\label{fig:order_z_difficult}{\includegraphics[width=\ttsize\linewidth]{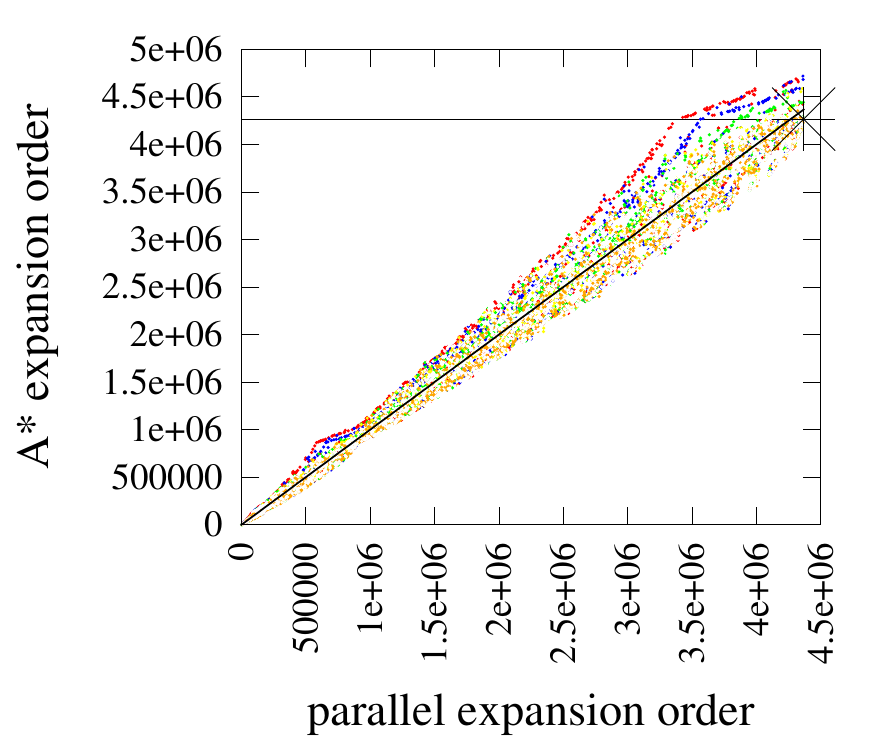} }} \hspace{\tspace}
	\subfloat[\AHDA{} on a difficult instance with 8 threads. As with the easy instance, \AHDA{} has a bigger band than \ZHDA{} on a difficult instance.]
	{\label{fig:order_a_difficult}{\includegraphics[width=\ttsize\linewidth]{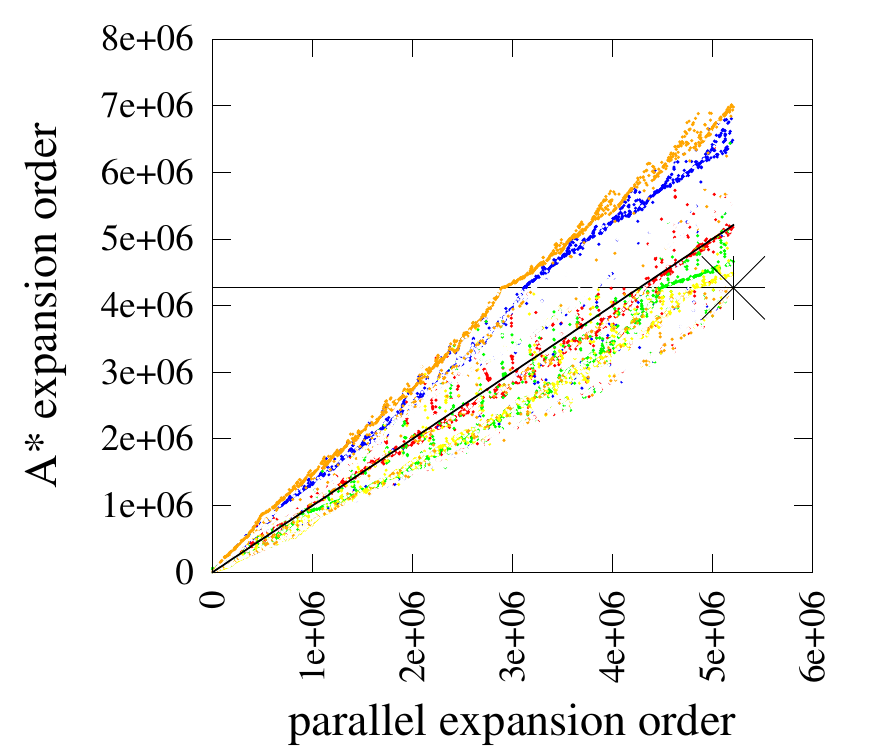} }} \hspace{\tspace}
	\subfloat[SafePBNF on a difficult instance with 8 threads. Because SafePBNF requires each thread to explore each nblock exclusively, the order of node expansion differs  significantly from A*. SafePBNF retains exploring promising nodes by switching nblocks at the cost of communication and coordination overhead.]
	{\label{fig:order_safepbnf_difficult}{\includegraphics[width=\ttsize\linewidth]{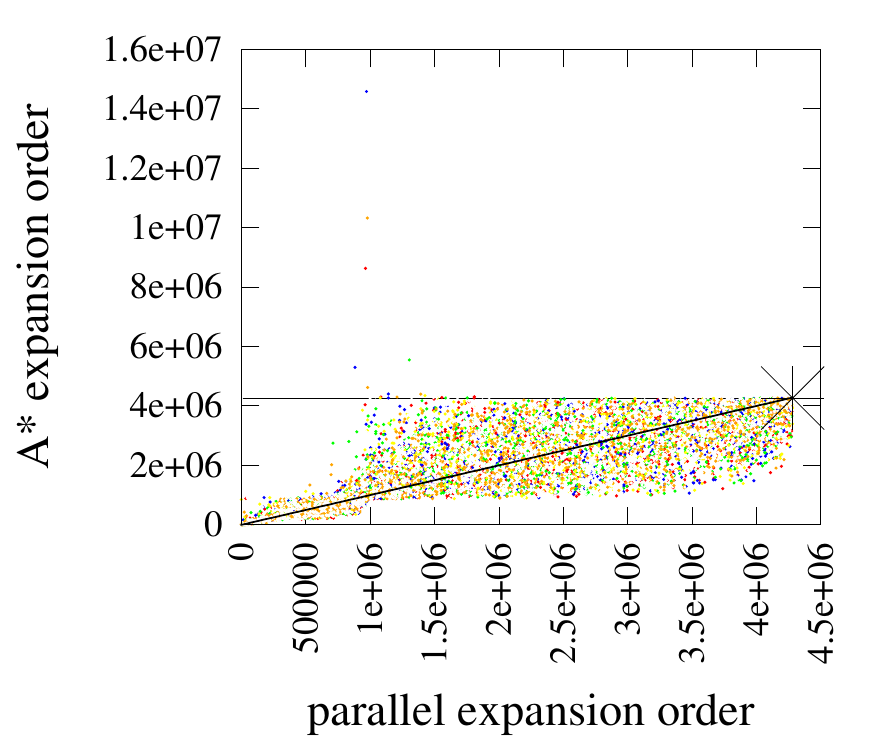} }} \hspace{\tspace}

	\subfloat[\ZHDA{} on a difficult instance with 8 threads with artificially slowed expansion rate. We did not observe a significant difference from \ZHDA without slow expansion.]
	{\label{fig:order_z_slow_difficult}{\includegraphics[width=\ttsize\linewidth]{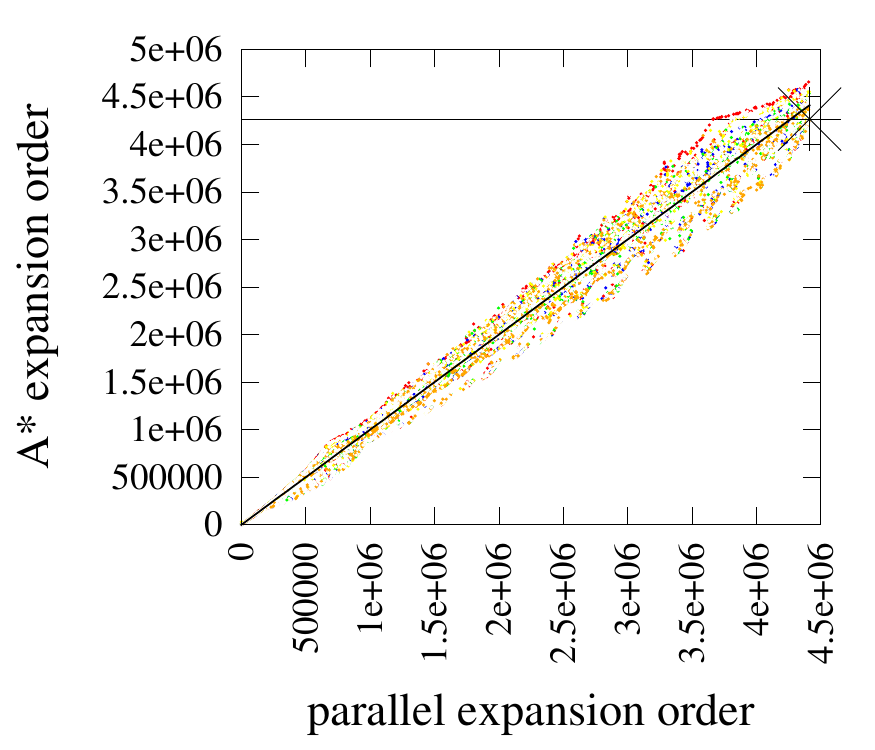} }} \hspace{\tspace}
	\subfloat[\PHDA{} on a difficult instance with 8 threads. \PHDA{} has the biggest band effect, significantly diverged from A*. \PHDA{} expands $> 7,000,000$ nodes to solve the instance which A* solves with $~4,000,000$ expansions.]
	{\label{fig:order_p_difficult}{\includegraphics[width=\ttsize\linewidth]{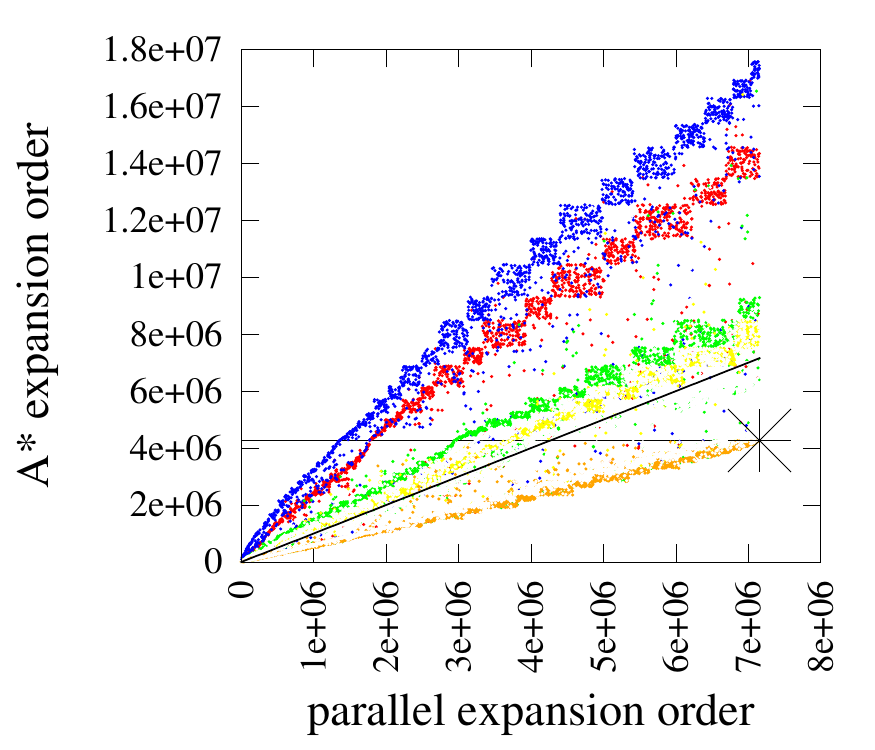} }} \hspace{\tspace}
   \subfloat[\ZHDA{} using FIFO tiebreaking  on a difficult instance with 8 threads (vs. A* using FIFO tiebreaking).]
	{\label{fig:order_z_fifo_difficult}{\includegraphics[width=\ttsize\linewidth]{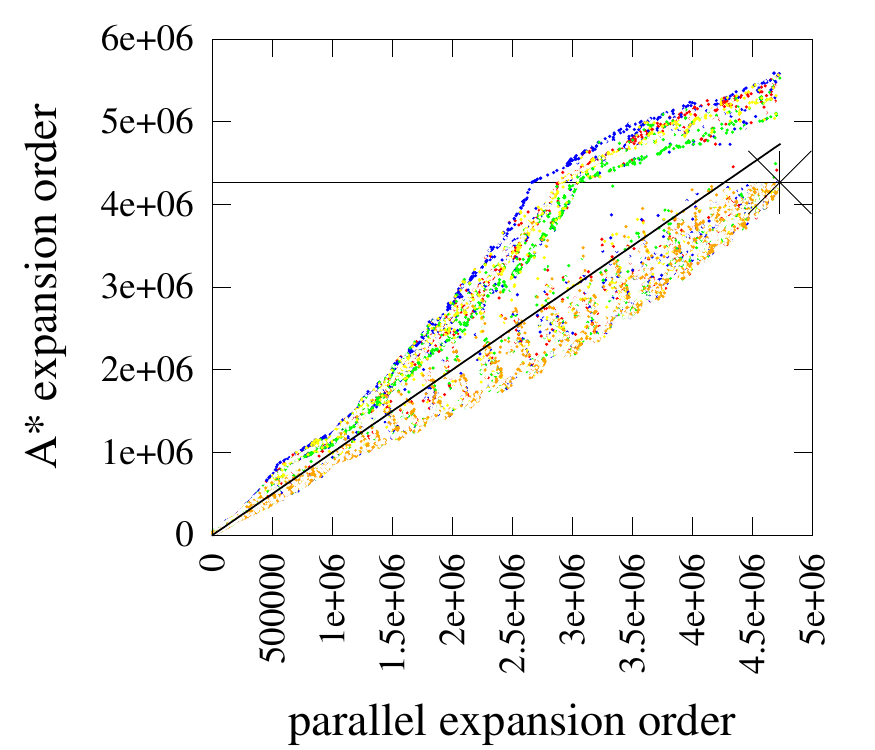} }}

	\subfloat {{\includegraphics[width=0.1\linewidth]{figures/order/legend1.pdf} }}
	\subfloat{{\includegraphics[width=0.1\linewidth]{figures/order/legend2.pdf} }}
	\subfloat{{\includegraphics[width=0.12\linewidth]{figures/order/legend3.pdf} }}    

	\caption{Comparison of node expansion order on a {\bf difficult instance of the 15-Puzzle} with 8 threads. The average node expansion order divergence of scores are \ZHDA{}: $\bar{d}=10,330.6$, \ZHDA{} (slowed): $\bar{d}=8,812.1$, \AHDA{}: $\bar{d}=245,818$, \PHDA{}: $\bar{d}=4,469,340$, SafePBNF: $\bar{d}=140,629.4$.}
	\label{fig:order_difficult}
\end{figure}

In Figures \ref{fig:order_nthreads}, \ref{fig:order_easy} and \ref{fig:order_difficult}, the horizontal axis represents the order in which state $s$ is expanded by parallel search (HDA* or SafePBNF).
The vertical axis represents the {\it A* expansion order} of state $s$, which is the order in which sequential A* expands node $s$. Note that although standard A* would terminate after finding an optimal solution, 
 we modified sequential A* for this set of experiments so that it continues to search even after the optimal solution has been found. 
This is because parallel search expands nodes that are not expanded by sequential A* (i.e., search overhead), and we want to know for all states expanded by parallel search which are not usually expanded by sequential A*, how much the parallel search has diverged from the behavior of sequential A*.

The line $y = x$ corresponds to an ideal, {\it strict} A* ordering in which the parallel expansion ordering is identical to the A* expansion order. The cross marks (``Goal'') in the figures represents the (optimal) solution found by A*, 
and the vertical line from the goal shows the total number of node expansions in A*. Thus, all nodes above this line results in SO. 
Note that unlike sequential A*, parallel A* can not terminate immediately after finding a solution, even if the heuristic is consistent, because when parallel A* finds an optimal solution it is possible that some nodes with $f < f^*$ have not been expanded (because they are assigned to a processor which is different from the processor where the solution was found). 

Although the traditional definition of A* \cite{HartNR68} specifies that nodes are expanded in order of nondecreasing $f$-value (i.e., best-first ordering), this is not sufficient to define a canonical node expansion ordering for sequential A* because many nodes can have the same $f$-value.
A tie-breaking policy can be used to impose a unique, canonical expansion ordering for sequential A*.
Our sequential A* uses a LIFO tie-breaking policy, which has been shown to result in good performance on the 15-puzzle \cite{burns2012implementing}, as well as domain-independent planning \cite{asai2016tiebreaking}.
In addition, all of our HDA* variants, as well as SafePBNF  uses LIFO tie-breaking for each local open list. Thus, by ``strict A*'' order, we mean ``the order in which A* with LIFO tie-breaking expands nodes'', and in Figures  \ref{fig:order_nthreads}, \ref{fig:order_easy} and \ref{fig:order_difficult} compare the expansion ordering of this ordering vs. HDA*/SafePBNF with local LIFO tiebreaking.

To verify that the results are not dependent on the particular tie-breaking policy, Figures \ref{fig:order_z_fifo_easy} and \ref{fig:order_z_fifo_difficult} show results where both sequential A* and the parallel algorithms use FIFO tie-breaking. These show that the results are not qualitatively affected by the choice of tie-breaking policy.

By analyzing the results, we observed three causes of search overhead on HDA*, (1) {\it Band Effect}, the divergence from the A* order due to load imbalance, (2) {\it Burst Effect}, an initialization overhead, and (3) node reexpansions. Below, we explain and discuss each of these overheads.

\subsubsection{Band Effect}
\label{sec:band}

The order in which states are expanded by \ZHDA{} is fairly consistent with sequential A*. However, 
there is some divergence from the strict A* ordering, within a ``band'' that is symmetrical around the strict A* ordering line.
For example, in Figure \ref{fig:order_2_threads}, we have highlighted a band showing that 
the (approximately) 5000'th state expanded by HDA* corresponds a strict A* order between 4500-5500 (i.e., a band width of approximately 1000 at this point in the search).
The width of the band tends to increase as the number of threads increases
 (see the bands in Figure \ref{fig:order_2_threads}, \ref{fig:order_4_threads},  \ref{fig:order_z_easy}).
Although the width of the band tends to increase as the search progresses, the rate of growth is relatively small.
Also, the harder the instance (i.e., the larger the number of nodes expanded by A*), the narrower the band tends to be (Figure \ref{fig:order_z_difficult}). 

A simple explanation for  this band effect is load imbalance. 
Suppose we use 2 threads, and assume that threads $t_1$ and $t_2$ share $p$ and $1-p$ of the nodes with $f$-value $=f_i$ for each $f_i$. 
Consider the $n$'th node expanded by  $t_1$. This should roughly correspond to the  $\frac{n}{p}$'th node expanded by sequential A*; at the same time,  $t_2$ should expand the node which roughly corresponds to the  $\frac{n}{1-p}$'th node expanded by sequential A*. In this case, the band size is $|\frac{n}{p} - \frac{n}{1-p}|$. Therefore, if $p=0.5$ (perfect load balance), the band is small, and as $p$ diverges from $0.5$, the band size becomes larger.

One possible, alternative interpretation of the band effect is that it is somehow related to or caused by other factors such as communications overhead or lock contention. 
To test this, we ran \ZHDA{} on 8 cores where the state expansion code was intentionally slowed down by adding a meaningless but time-consuming computation to each state expansion.\footnote{At the beginning of the search on each thread, we initialize a thread-local, global integer $i$ to 7.  On each thread, after each node expansion, we perform the following computation 100,000 times: $j = 11 i  \bmod  9999943$, and then set $i \leftarrow j$. This is a heavy computation with a small memory footprint and is intended to occupy the thread without causing additional memory accesses.} 
If the band effect was caused by communications or lock contention related issues, it should not manifest itself if the node expansion rate is so slow that the relative cost of communications and synchronization is very small. 
However, as shown in Figure \ref{fig:order_z_slow_easy} and \ref{fig:order_z_slow_difficult}, the band effect remains clearly visible even when the node expansion rate is very slow, indicating that the band effect is not an accidental overhead caused by communications or lock contention (similar results were obtained for other instances).

\begin{observation}
The band effect on \ZHDA{} represents load imbalance between threads.
The width of the band  determines the extent to which superlinear speedup or search overhead (compared to sequential A*) can occur.
Furthermore, the band effect is independent of node evaluation rate.
\label{observation:}
\end{observation}

The expansion order of SafePBNF is shown in Figure \ref{fig:order_safepbnf_easy} and \ref{fig:order_safepbnf_difficult}.
Because SafePBNF requires each thread to explore each nblock (and duplicated detection scope) exclusively, the order of node expansion is significantly different from A*. However, SafePBNF tries to explore promising nodes by switching among nblocks to focus on nblocks which contain the most promising nodes.
This requires communication and coordination overhead, which increases the walltime by about  $<$10\% of the time on the 15-puzzle \cite{burnslrz10}. 

\subsubsection{Burst Effect}
\label{sec:burst}
At the beginning of the search, it is possible for the node expansion order of HDA* to deviate significantly from strict A* order due to a temporary ``burst effect''.
Since there is some variation in the amount of time it takes to initialize each individual thread and populate all of the thread open lists with ``good'' nodes, it is possible that some threads may initially expand nodes in poor regions of the search space because good nodes have not yet been sent to their open lists from threads that have not yet completed their initialization.
For example, suppose that $n_1$ is a child of the root node $n_0$, and $n_1$ has a significantly worse $f$-value than other descendants of $n_0$.
Sequential A* will not expand $n_1$ until all nodes with lower $f$-values have been expanded.
However, at the beginning of search, $n_1$ may be assigned to a thread $t_1$ whose queue $q_1$ is empty, in which case $t_1$ will immediately expand $n_1$. The children of $n_1$  may also have $f$-values which are significantly worse than other descendants of $n_0$, but if those children of $n_1$ are in turn assigned to threads with queues that are (near) empty or otherwise populated by other ``bad'' nodes with poor $f$-values, then those children will get expanded, and so on.
Thus, at the beginning of the search, many such bad nodes will be expanded because all queues are initially empty, bad nodes will continue to be expanded until the queues are filled with ``good'' nodes.
As the search progresses, all queues will be filled with good nodes, and the search order will more closely approximate that of sequential A*.

Furthermore, these burst-overhead nodes tend to be reached through suboptimal paths (because states necessary for better paths are unavailable during the burst phase), and therefore tend to be revisited later via shorter paths, contributing to revisited node overhead.

The burst phenomenon is clearly illustrated in Figures \ref{fig:order_4_threads} and \ref{fig:order_z_easy}, which shows the behavior of \ZHDA{} with 8 threads on a small 15-puzzle problem (solved by A* in 8966 expansions). 
The large vertically oriented cluster at the left of the figure shows that states with a strict A* order of over 30,000 are being expanded within the first 2,000 expansions by HDA*.
The A* implementation we used expands over 85,248 nodes per second (the node expansion includes overhead for storing node information in the local data structure, so it is slower than base implementation in \citeauthor{burnslrz10} \citeyearR{burnslrz10}), this burst phenomenon is occurring within the first 0.023 seconds of search. 

Figure  \ref{fig:order_z_difficult} shows that on a harder problem instance which requires $>$ 4,000,000 state expansions by A*, the overall effect of this initial burst overhead is negligible.

Figure \ref{fig:order_z_slow_easy} shows that when the node expansion rate is artificially slowed down, the burst effect is not noticeable even if the number of states expansions necessary to solve the problem with A* is small ($<$ 10,000). This is consistent with our explanation above that the burst effect is caused by brief, staggered initialization of the threads -- when state expansions are slow, the staggered start becomes irrelevant.

From the above, we can conclude that the burst effect is only significant when the problem can be solved very quickly ($<$ 0.88 seconds) by A* and the node expansion rate is fast enough that the staggered initialization can cause a measurable effect. 

The practical significance of the burst effect depends on the characteristics of the application domain. In puzzle-solving domains, the time scales are usually such that the burst effect is inconsequential. However, in domains such as real-time path planning, the total time available for planning can be just as a fraction of a second, so the burst effect can have a significant impact.

\begin{observation}
The burst effect in \ZHDA{} can dominate search behavior on easy problems, resulting in large search overhead. 
However,  the burst effect is insignificant on harder problems, as well as when node expansion rate is slow.
\label{observation:}
\end{observation}

The burst effect is less pronounced in SafePBNF compared to \ZHDA{}, because
a thread in SafePBNF prohibits other threads from exploring its duplicate detection scope.   Thus, the nodes shown in Figure \ref{fig:order_safepbnf_easy} are actually band effect (see above), which means that this overhead is persistent in SafePBNF through the search (Figure \ref{fig:order_safepbnf_difficult}).

\subsubsection{Node Reexpansions}
\label{sec:node-reexpansions}
With a consistent heuristic, A* never reexpands a node once it is saved in the closed list, because the first time a node is expanded, we are guaranteed to have reached through a lowest-cost path to that node.  
However, in parallel best-first search, nodes may need to be reexpanded even if they are in the closed list. 
For example, in HDA*, each processor selects the best (lowest $f$-cost) node in its local open list, but the selected node may not have the current globally lowest $f$-value.
As a result, although HDA* tends to find shortest paths to a node first, the paths may not be lowest-cost paths, and 
some node $n$ which is expanded by some thread in HDA* may have been reached through a suboptimal path, and must later be reexpanded after it is reached through a lower-cost path.

This is not a significant overhead for unit-cost domains because shorter paths always have smaller cost.
In fact, we observed that \ZHDA{}, \AHDA{} and SafePBNF had low reexpansion rates for on the 15-puzzle. For \ZHDA{} with 8 threads, the average reexpansion rate $R_r$ was $2.61 \times 10^{-5}$ for 100 instances.  

Node reexpansions are more problematic in non-unit cost domains, because a shorter path does not always mean a smaller cost.
\cite{Kobayashi2011evaluations} analyzed node reexpansion on multiple sequence alignment which \ZHDA{} suffers from high node duplication rate.
We discuss node reexpansions by HDA* on the multiple sequence alignment problem in Section \ref{sec:msa}.

\subsubsection{The Impact of Work Distribution Method on the Order of Node Expansion} 
\label{sec:experiments-order}
In addition to \ZHDA{}, we investigated the order of node expansion on \AHDA{}, \PHDA{}, and SafePBNF.
The abstraction used for \AHDA{} ignores the positions of all tiles except tiles 1, 2, and 3.\footnote{We tried (1) ignoring all tiles except tiles 1,2, and 3, (2) ignoring all tiles except tiles 1, 2, 3, and 4, (3) mapping cells to rows, and (5) mapping cells to the blocks 
, and chose (1) because it performed the best.}
\PHDA{} is an instance of HDA* which is called ``HDA*'' in the work of \citeauthor{burnslrz10} \citeyear{burnslrz10}.
Unlike the original HDA* in \citeauthor{kishimotofb09} \citeyear{kishimotofb09}, which uses Zobrist hashing,
\PHDA{} uses a perfect hashing scheme  which maps permutations (tile positions) to lexicographic indices (thread IDs) in \citeauthor{korf2005large} \citeyear{korf2005large}.
A perfect hashing scheme computes a unique mapping from permutations (abstract state encoding) to lexicographic indices (thread ID)\footnote{The permutation encoding used by \PHDA{} is defined as:
$H(s) = c_1 k! + c_2 (k-1)! +...+ c_k 1!$ where the position of tile $p(i)$ is the $c_i$-th smallest number in the set $\{1,2,3,...,16\} \setminus \{c_1,c_2,...c_{i-1}\}$. State $s$ is sent to a process with process id $H(s) \; mod \; n$, where $n$ is the number of processes. Therefore, if $n=8$ then $H(s) \; mod \; n = \{c_{k-2} 3! + c_{k-1} 2! + c_k 1!\}$, thus it only depends on the relative positions of tiles 12, 13, and 14. In addition, processes with odd/even id only send nodes to processes with odd/even id unless the position of 14 changes.}.
While this encoding is effective for its original purpose of efficient representation of states for external-memory search, 
it was \emph{not} designed for the purpose of work distribution.
For SafePBNF, we used the configuration used in \citeauthor{burnslrz10} \citeyear{burnslrz10}.

Figures \ref{fig:order_easy} and \ref{fig:order_difficult} compare the expansion orders of \ZHDA{}, \AHDA{}, \PHDA{}, and SafePBNF. 
Although some trends are obvious by visual inspection, e.g., the band effect is larger for \AHDA{} than on \ZHDA{}, a quantitative comparison is useful to gain more insight.

Thus, we calculated the average {\it divergence} of each algorithm, where divergence of a parallel search algorithm $B$ 
on a problem instance $I$
is defined as follows:  Let $N_{A*}(s)$ be the order in which state $s$ is expanded by A*, and let $N_{B}(s)$ be order in which $s$ is expanded by $B$, and let $V(A^*,B)$ be the set of all states expanded  by both A* and $P$. In case $s$ is reexpanded by an algorithm, we use the first expansion order. 
Then the divergence of $B$ from A* on instance $I$ is $d(I) = \sum_{s \in V(A^*,B)} | N_{A^*}(s) - N_{B}(s) | \; / \; |V(A^*, B)|$. 
We computed the average divergence $\bar{d}$ for the 50 most difficult instances in the instance set. 

In addition to the divergence $d$, we calculated the average number of {\it premature expansions} $p$, which is the number of nodes expanded before all nodes with lower $f$-value than that node are expanded. 
Unlike the divergence, the number of premature expansions is not significantly influenced by the expansion order within the same $f$-value. 

\begin{table}
	\centering
	\caption{Comparison of the average divergence ($\bar{d}$) and premature expansions ($\bar{p}$) for the 50 most difficult 15-puzzle instances.}
	\label{divergence}
	\begin{tabular}{l|rr} \hline
		                & $\bar{d}$   & $\bar{p}$  \\ \hline
		\ZHDA{}         &    10,330.6 &   563,605 \\
		SafePBNF        &   140,629.4 &   598,759 \\
		\AHDA{}         &   245,818.0 & 2,595,540 \\
		\PHDA{}         & 4,469,340.0 & 3,725,942
	\end{tabular}
\end{table}

The average divergence and premature expansions for these difficult instances are shown in Table \ref{divergence}.
These results indicate that the order of node expansion of \ZHDA{} is the most similar to that of A*. 
Therefore, \ZHDA{} is expected to have the least SO. 
The abstraction-based methods, \AHDA{} and SafePBNF, have significantly higher divergence than \ZHDA{}, which is not surprising, since by design, these methods do not seek to simulate A* expansion order. 
Finally, \PHDA{} has a huge divergence, and is expected to have very high SO -- it is somewhat surprising that a work distribution function can have divergence (and search overhead) which is so much higher than methods that focus entirely on reducing communications overhead such as \AHDA{}.
We evaluate the SO and speedup of each method below in Section \ref{sec:hdavspbnf}.




\subsection{Revisiting HDA* (\ZHDA{}, \AHDAP{}, \AHDAZ{}, \PHDA{}) vs. SafePBNF for Admissible Search}
\label{sec:hdavspbnf}

Previous work  compared \PHDA{}, \AHDA{}, and SafePBNF on the 15-puzzle and grid pathfinding problems \cite{burnslrz10}. They also compared SafePBNF with \AHDA{} on domain-independent planning. The overall conclusion of this previous study was that among the algorithms evaluated, SafePBNF performed best for optimal search.
We now revisit this evaluation, in light of the results in the previous section, as well as recent improvements to implementation techniques.
There are three issues to note regarding the experimental settings used in \citeauthor{burnslrz10} \citeyear{burnslrz10}:

Firstly, {\it the previous comparison did not include \ZHDA{}, the original HDA* which uses Zobrist hashing} \cite{kishimotofb09,kishimotofb13}.
\citeauthor{burnslrz10} \citeyear{burnslrz10} evaluated two variants of HDA*:  \PHDA{} (which was called ``HDA*'' in their paper) and \AHDA{} (called ``AHDA*'' in their paper).
As shown above, the node expansion order of \ZHDA{} has a much smaller divergence from A* compared to SafePBNF and \AHDA{}.
While \ZHDA{} seeks to minimizes search overhead, and both \AHDA{} as well as SafePBNF seek to reduce communications overhead, \PHDA{} minimizes neither communications nor search overheads (as shown above, it has much higher expansion order divergence than all other methods), so \PHDA{} is not a good representative of the HDA* framework.
Therefore, a direct comparison of SafePBNF and \AHDA{} (which minimize communications overhead) to \ZHDA{} (which minimizes search overhead) is necessary in order to understand how these opposing objectives affect performance.

Secondly, the 15-puzzle and grid search instances used in the previous study only required a small amount of search, so the behavior of these algorithms on difficult problems has not been compared.
In the previous study, the grid domains consisted of 5000x5000 grids, and the 15-puzzle instances were all solvable within 3 million expansions by A*. 
Since grid pathfinding solvers can generate $~10^6$  
nodes per second, and 15-puzzle solvers can generate 
$~0.5 \times 10^6$ 
nodes per second, these instances are solvable in under a second by a 8-core parallel search algorithm. 
  As shown in section \ref{sec:so}, when the search only takes a fraction of a second, HDA* incurs significant search overhead due to the burst effect, but the burst effect is a startup overhead whose impact is negligible on problem instances that require more search.

Thirdly, in the previous study, for all algorithms, a  binary heap implementation for the open list priority queue was used, which incurs $O(log N)$ costs for insertion. This introduces a bias favoring PBNF over all of the HDA* variants, because
PBNF uses a separate binary heap for each $n$-block. Splitting the open list into many binary heaps significantly decreases the $N$ in the $O(log N)$ cost node insertions compared to algorithms such as HDA* which use a single open list per thread. 
However, 
it has been shown that a bucket implementation ($O(1)$ for all operations) results in significantly faster performance on state-of-the-art A* implementations \cite{burns2012implementing}.


Therefore, we revisit the comparison of HDA* and SafePBNF by 
(1) using Zobrist hashing for HDA* (i.e., \ZHDA{}) in order to minimize search overhead (2) using both easy instances (solvable in $<$ 1 second) and hard instances (requiring up to 1000 seconds to solve with sequential A*) of the sliding-tile puzzle and grid path-finding domains in order to isolate the startup costs associated with the burst effect, and (3) using both bucket and heap implementations of the open list in order to isolate the effect of data structure efficiency (as opposed to search efficiency).

\newcommand{\cmpsize}{0.5}

\begin{figure}[htb]
	\centering
	\subfloat[15-puzzle (bucket open list)] {\label{fig:15puzzle_vector_walltime}{\includegraphics[width=\cmpsize\linewidth]{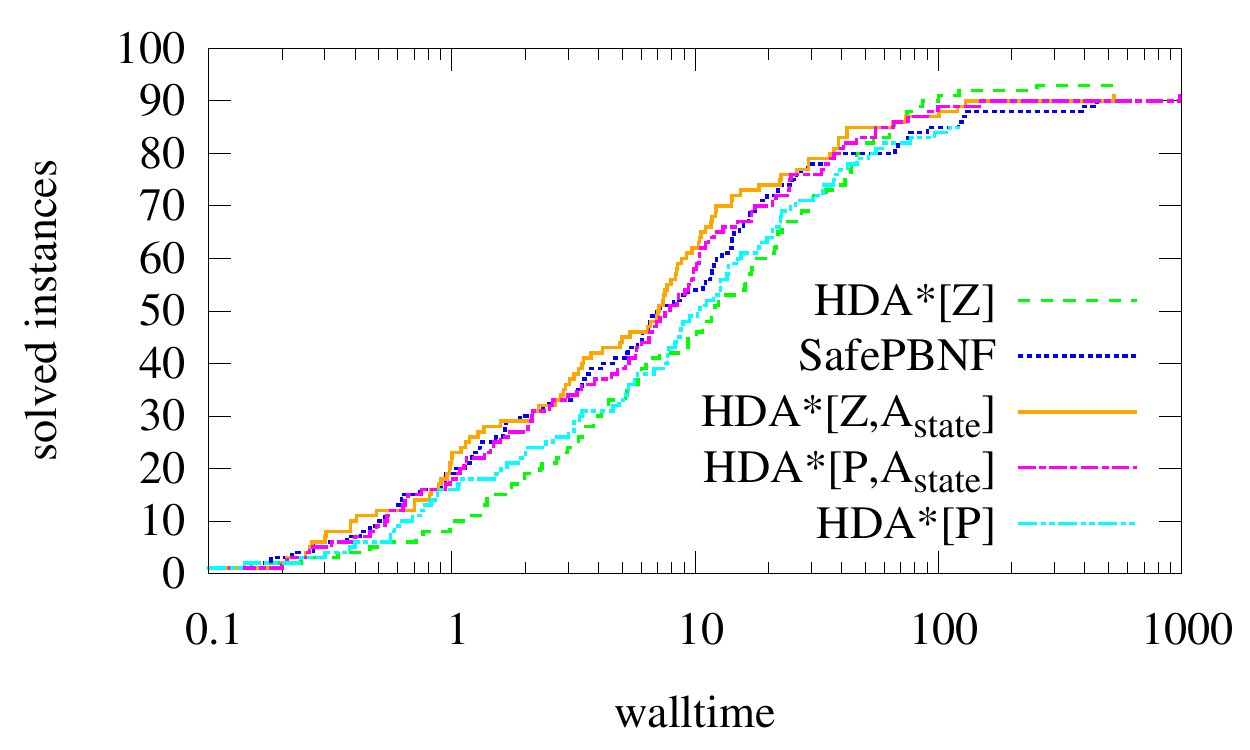}}}
	\subfloat[15-puzzle (heap open list)]{\label{fig:15puzzle_heap_walltime}{\includegraphics[width=\cmpsize\linewidth]{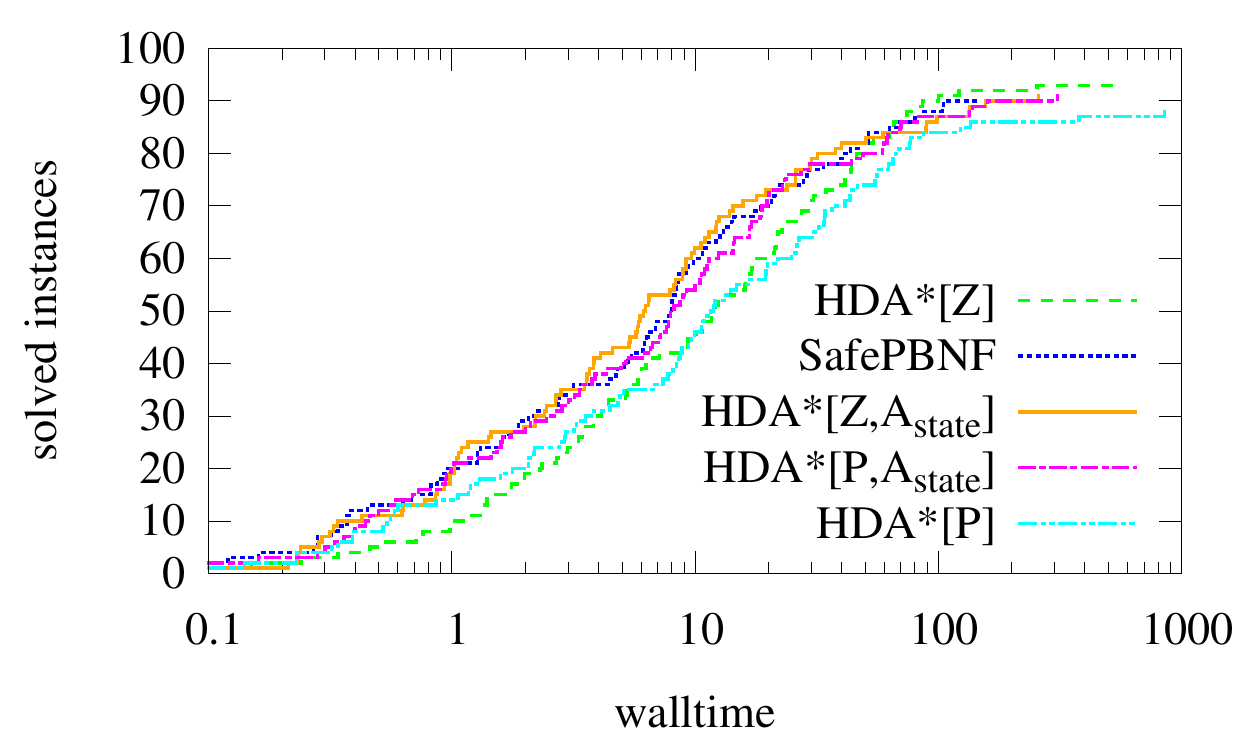}}}

	\subfloat[24-puzzle (bucket open list)]{\label{fig:24puzzle_vector_walltime}{\includegraphics[width=\cmpsize\linewidth]{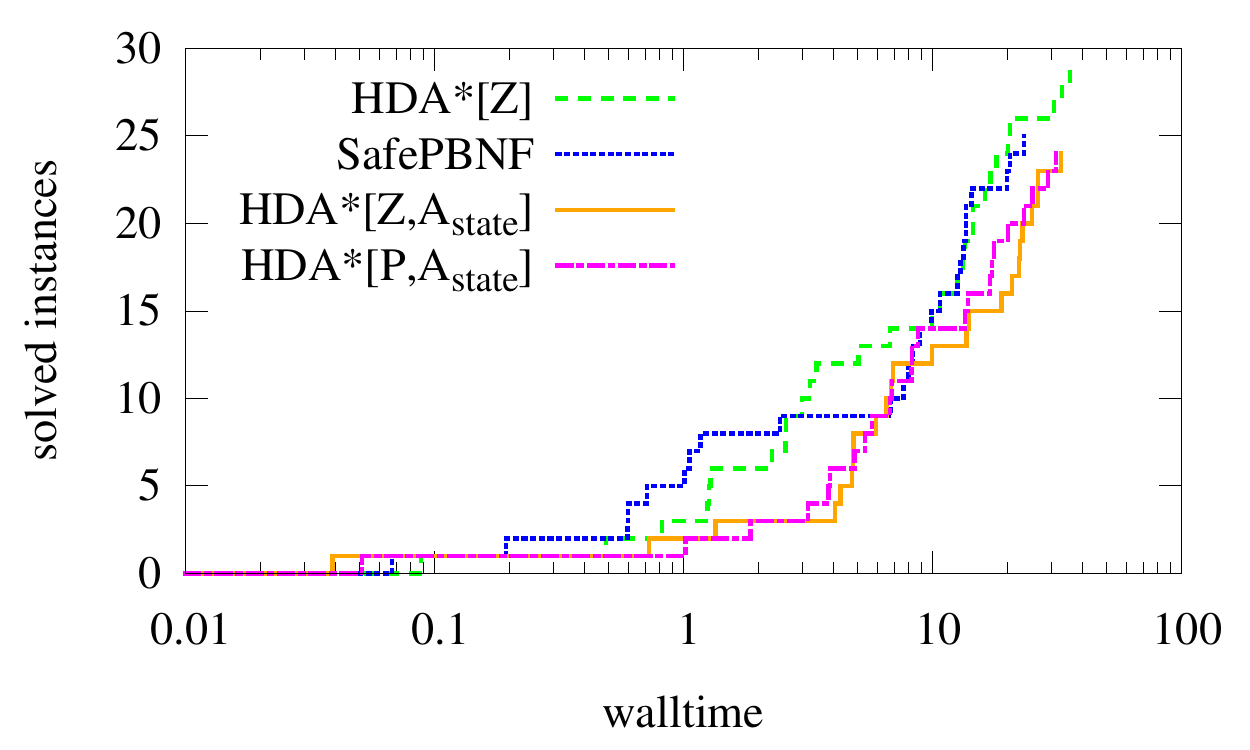}}}
	\subfloat[Grid Pathfinding (bucket open list)]{\label{fig:grid_walltime}{\includegraphics[width=\cmpsize\linewidth]{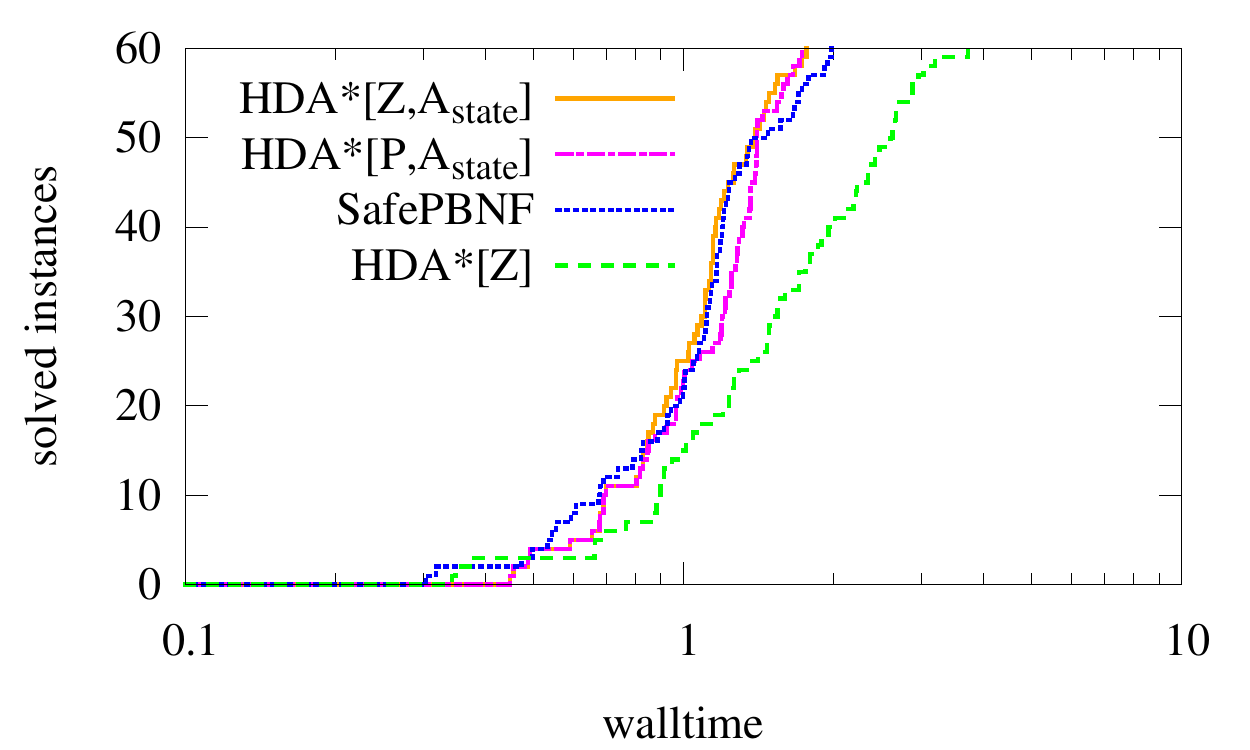}}}
	\caption{Comparison of the number of instances solved within given walltime. The x axis shows the walltime and y axis shows the number of instances solved by the given walltime. In general, \ZHDA{} outperforms SafePBNF on difficult instances ($>10$ seconds) and SafePBNF outperforms \ZHDA{} on easy instances ($<10$ seconds).}
	\label{fig:comparison}
	\captionlistentry[experiment]{HDA*vsPBNF:walltime, Figure \ref{fig:comparison}, supermicro, 15puzzle:random100:bucket 15puzzle:random100:heap 24puzzle:TODO:bucket grid:random60-0.45:bucket, pthread, newmaterial}
\end{figure}

\begin{figure}[htb]
	\centering
	\subfloat[15-puzzle (bucket open list)] {\label{fig:15puzzle_vector_expd}{\includegraphics[width=\cmpsize\linewidth]{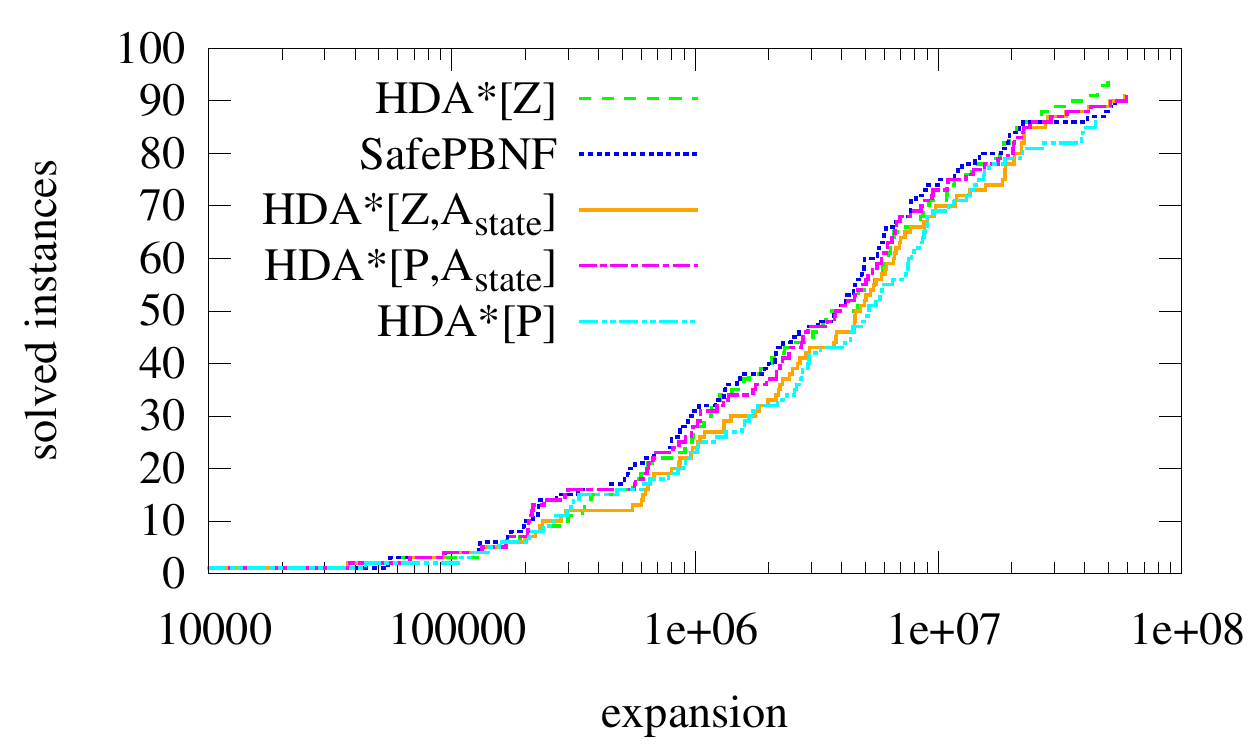}}}
	\subfloat[15-puzzle (heap open list)]{\label{fig:15puzzle_heap_expd}{\includegraphics[width=\cmpsize\linewidth]{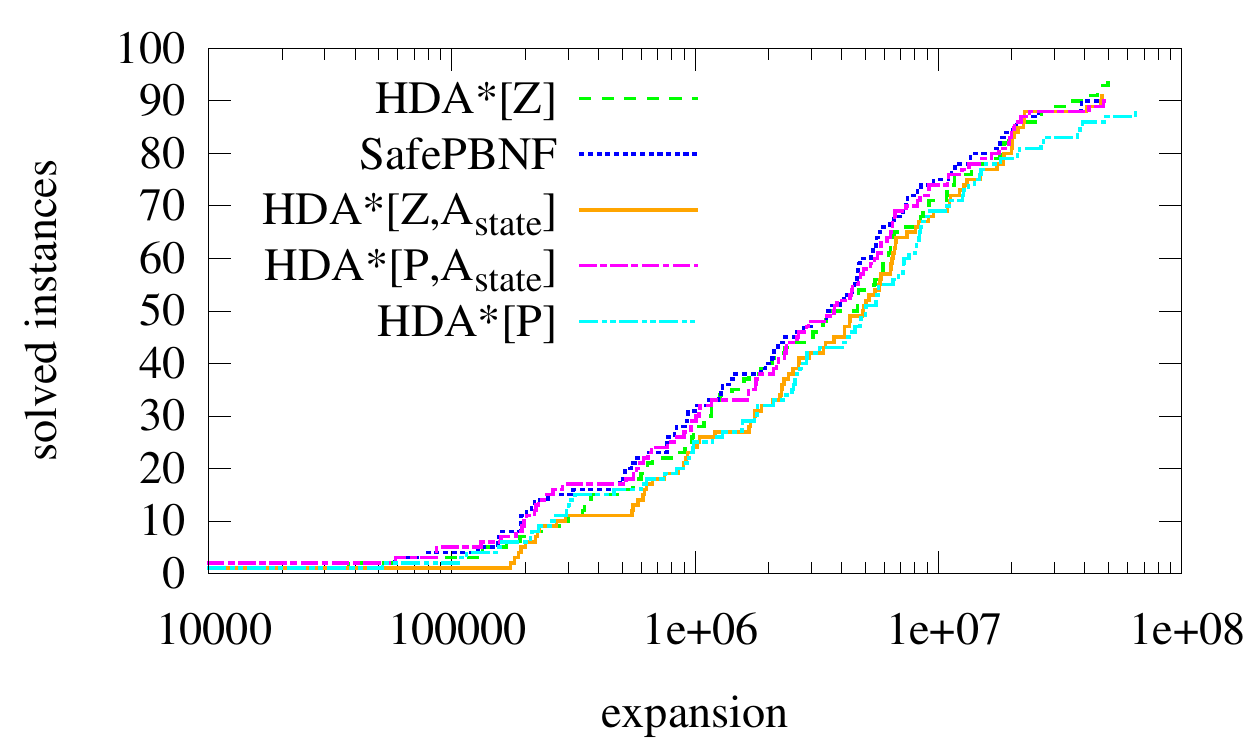}}}

	\subfloat[24-puzzle (bucket open list)]{\label{fig:24puzzle_vector_expd}{\includegraphics[width=\cmpsize\linewidth]{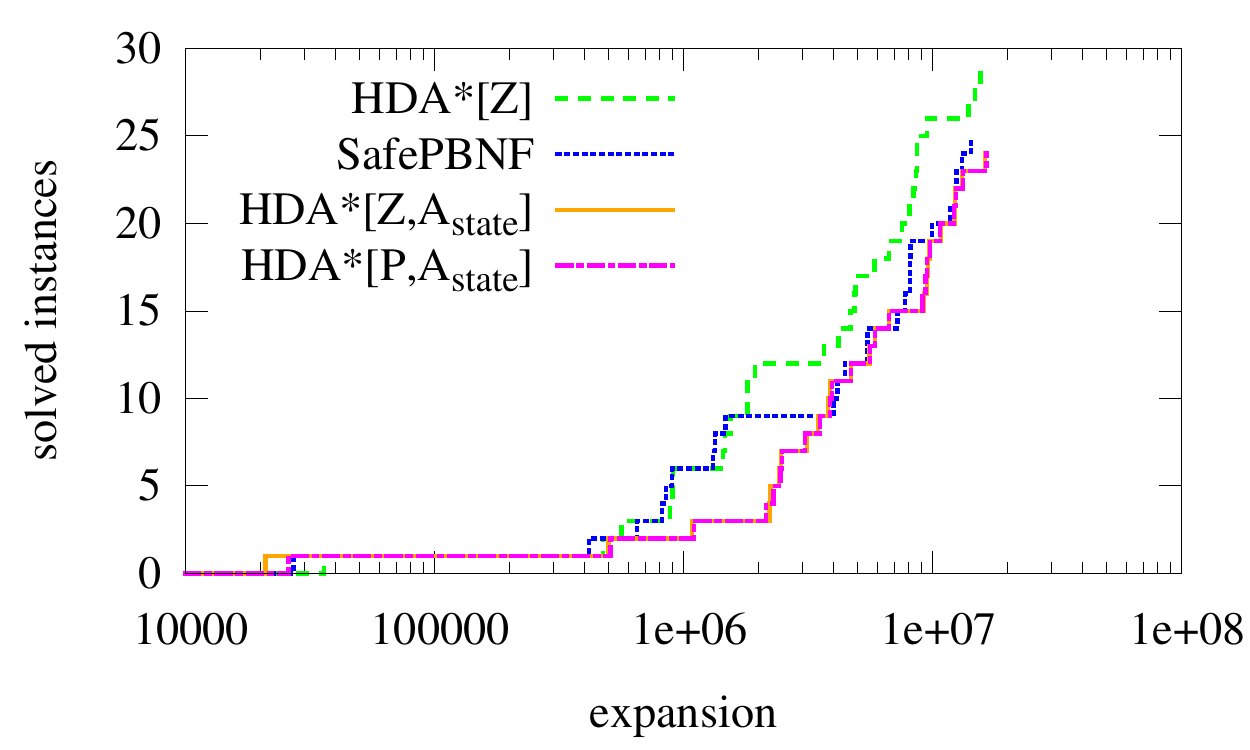}}}
	\subfloat[Grid Pathfinding (bucket open list)]{\label{fig:grid_expd}{\includegraphics[width=\cmpsize\linewidth]{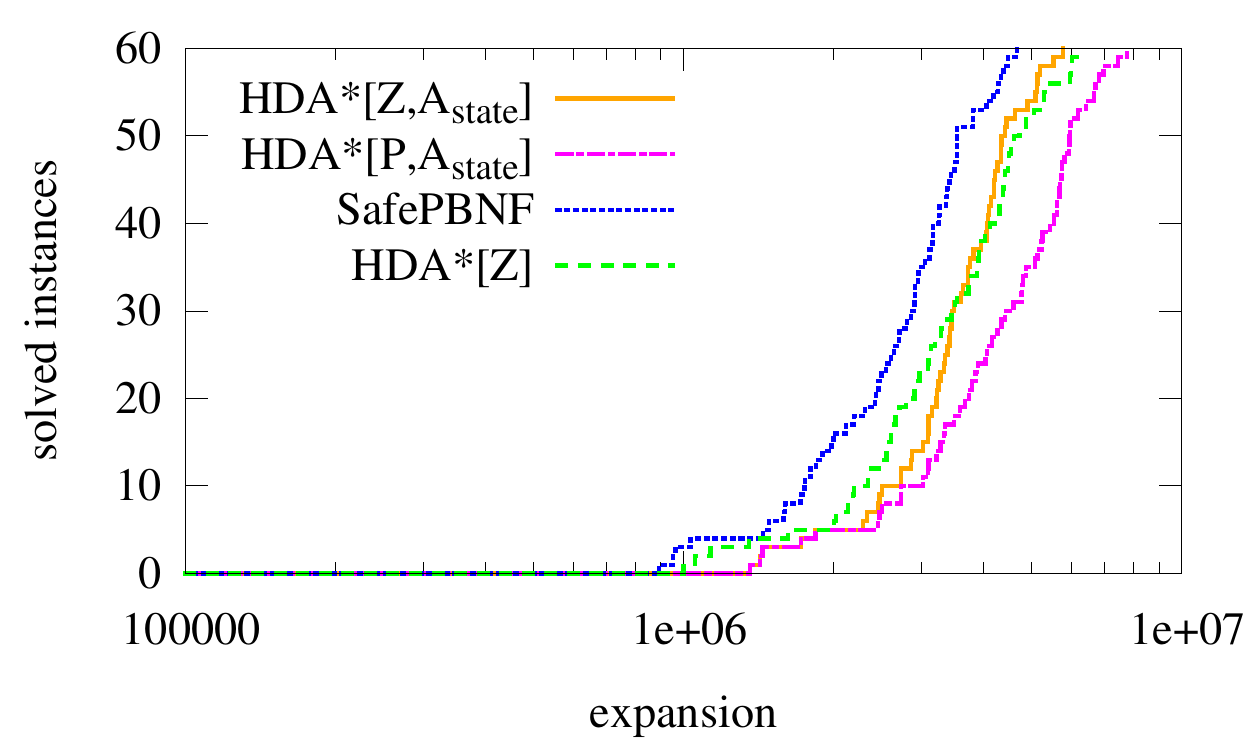}}}
	\caption{Comparison of the number of instances solved within given number of node expansions. The x axis shows the walltime and y axis shows the number of instances solved by the given node expansion. Overall, \ZHDA{} has the lowest SO except in grid pathfinding, where \ZHDA{} suffers from high node duplication because the node expansion is extremely fast in grid pathfinding. \AHDAZ{} and \AHDAP{} expanded almost identical number of nodes in 24-puzzle.}
	\label{fig:comparison_expd}
	\captionlistentry[experiment]{HDA*vsPBNF:expd, Figure \ref{fig:comparison_expd}, supermicro, 15puzzle:random100:bucket 15puzzle:random100:heap 24puzzle:TODO:bucket grid:random60-0.45:bucket, pthread, newmaterial}
\end{figure}%

For the 15-puzzle, we used the standard set of 100 instances in \citeauthor{korf:85a} \citeyear{korf:85a}, and used the Manhattan Distance heuristic. We used the same configuration used in Section \ref{sec:experiments-order} for all algorithms (except without the instrumentation to storing the expansion order information for each state).
For the 24-puzzle, we used 30 instances randomly generated which could be solved within 1000 seconds by sequential A*, and used the pattern database heuristic \cite{Korf2002}. 
The abstraction used by \AHDAP{}, \AHDAZ{}, and SafePBNF ignores the numbers on all of the tiles except tiles 1,2,3,4, and 5.\footnote{We tried (1) ignoring all tiles except tiles 1-5, (2) ignoring all tiles except tiles 1-6, (3) ignoring all tiles except tiles 1-4, (4) mapping cells to rows, and (5) mapping cells to the blocks, 
and  chose (1), the best performer.}
For (4-way unit-cost) grid path-finding, we used 60 instances based obtained by  randomly generating 5000x5000 grids where  0.45 of the cells are obstacles. We used Manhattan distance as a heuristic. 
The abstraction used for \AHDAP{} and \AHDAZ{} maps 100x100 nodes to an abstract node, which performed the best among 5x5, 10x10, 50x50, 100x100, and 500x500 (Section \ref{sec:co}). 
For SafePBNF we used the same configuration used in previous work \cite{burnslrz10}.
The queue of free nblocks is implemented using a binary tree as there was no significant difference in performance using vector implementation.

Figure \ref{fig:comparison} compares the number of instances solved as a function of wall-clock time by \ZHDA{}, \AHDA{}, \PHDA{}, and SafePBNF. 
The results show that on the 15-puzzle and  24-puzzle, grid pathfinding, PBNF initially outperforms \ZHDA{}, but as more time is consumed, \ZHDA{} solves more instances than PBNF, i.e., PBNF outperforms  \ZHDA{} on easier problems due to the burst effect (Section \ref{sec:burst}), while \ZHDA{} outperforms SafePBNF on more difficult instances because after the initial burst effect subsides, \ZHDA{} diverges less from A* node expansion order and therefore incurs less search overhead. 

\begin{observation}
\ZHDA{} significantly outperforms SafePBNF on 15-puzzle and 24-puzzle instances that require a significant amount of search. On instances that can be solved quickly, SafePBNF outperforms \ZHDA{} due to the burst effect.
\end{observation}

Comparing the results for the 15-puzzle for the bucket open list implementation (Figure \ref{fig:15puzzle_vector_walltime})
and the heap open list implementation (Figure \ref{fig:15puzzle_heap_walltime}), we observe that all of the HDA* variants benefit from using a bucket open list implementation. Not surprisingly, for the more difficult problems, the benefit of the more efficient data structure ($O(1)$ vs. $O(logN)$ insertion for $N$ states) becomes more significant. PBNF does not benefit as much from the bucket open list because in PBNF, there is a separate queue associated with each $n$-block, so the difference between bucket and heap implementations is $O(1)$ vs. $O(logN/B)$, where $B$ is the number of $n$-blocks.

Figure \ref{fig:comparison_expd} compares the number of solved instances within the number of node expanded. Due to the burst effect, with a small number of expansions,  \ZHDA{} solves fewer instances compared to SafePBNF, especially in the grid pathfinding domain.



\subsubsection{On the Effect of Hashing Strategy in AHDA* (\AHDAZ{} vs. \AHDAP{})}

In addition to the original implementation of AHDA* \cite{burnslrz10}, which distributes abstract states using a perfect hashing (\AHDAP{}), we implemented \AHDAZ{} which uses Zobrist hashing to distribute.
Interestingly, Figure \ref{fig:comparison_expd} shows that both \AHDAZ{} and \AHDAP{} achieved lower search overhead than \PHDA{} in 15-puzzle. A possible explanation is that the abstraction is hand-crafted so that the abstract nodes are sized equally and distributed evenly in the search space.
On the other hand, as an abstract state is already a large set of nodes, distributing abstract states using Zobrist hashing (\AHDAZ{}) does not yield significantly better search overhead compared to \AHDAP{}.

\subsection{The Effect of Communication Overhead on Speedup}
\label{sec:co}

Although \ZHDA{} is competitive with the abstraction-based  methods (\AHDA{} and SafePBNF) on the sliding-tile puzzle domains,  Figure \ref{fig:grid_walltime} shows that \AHDA{} and SafePBNF significantly outperformed \ZHDA{} in the grid path-finding domain. 
Interestingly, Figure \ref{fig:grid_expd} shows that \AHDA{} and \ZHDA{} solve roughly the same number problems, given the same number of node expansions. This indicates that the performance difference between \AHDA{} and \ZHDA{} on the grid domain is {\it not} due to search overhead, but rather due to the fact that \AHDA{} is able to expand nodes faster than \ZHDA{}.
In previous work, \citeauthor{burnslrz10} \citeyear{burnslrz10} showed that \PHDA{} suffers from high communications overhead on the grid domain.\footnote{\citeauthor{burnslrz10} \citeyear{burnslrz10} evaluated HDA* (\PHDA{}) on the grid problem using a perfect hash function {\it processor(s)} $= (x \cdot y_{max} + y) \; \bmod \; p$ ($p$ is the number of processes) of the state location for work distribution. 
This hash function results in different behavior depending on the number of processes. If $(y_{max} \; \bmod \; p) = 0$, then all cells in each row have the same hash value, but all pairs of adjacent rows are guaranteed to have different hash values. If $(y_{max} \; \bmod \; p) \neq 0$, all pairs adjacent cells are guaranteed to have different hash values. Both cases result in high communication overhead, thus \AHDA{} (100x100) significantly outperformed \PHDA{} in both cases.} 


Although HDA* uses asynchronous communication, sending/receiving messages require access to data structure such as message queues.
Communication costs are crucial in grid path-finding because the node expansion rate is extremely high in grid path-finding.
Fast node expansion means that the relative time to send a node is higher.
Our grid solver expands 955,789 nodes/second, much faster than our 15-puzzle (bucket) solver (565,721 node/second). 
Thus, the relative cost of communication in the grid domain is twice as high as in the 15-puzzle.

To understand the impact of communications overhead, we evaluated the 
 speedup, communications overhead (CO), and search overhead (SO)  of \AHDA{} with different abstraction sizes. 
The abstraction used for \AHDA{} 
maps $k \times k$ blocks in the grid to a single abstract state.
Note that in this domain, an abstraction size of 1 corresponds to \PHDA{}.
Table \ref{grid-ahda} shows the results.
As the size of the $k \times k$ block increases, communications is reduced, and
as a result, 100x100 \AHDA{} is faster than \ZHDA{} and \PHDA{} although it has the same amount of SO.
However, there is a point of diminishing returns due to load imbalance -- in the extreme case when the entire $N \times N$ grid is mapped to a single abstract state, there would be no communications but only 1 processor would have work.
Thus, a 500x500 abstraction results in worse performance  than a 100x100 abstraction.

\begin{table}[htb]
	\caption{Comparison of speedup, communication overhead, and search overhead of \AHDA{} on grid path-finding using different abstraction sizes. CO: communication overhead $(= \frac{\text{\# nodes sent to other threads}}{\text{\# nodes generated}})$, SO: search overhead $(= \frac{\text{\# nodes expanded in parallel}}{\text{\# nodes expanded in sequential search}} - 1)$.}
	\label{grid-ahda}
	\centering
	\begin{tabular}{l|rrr} \hline
		abstraction size & speedup & CO & SO \\ \hline
		\ZHDA{} & 2.61 & 0.87 & 0.05 \\
		1x1 (= \PHDA{}) & 2.57 & 0.87 & 0.05 \\
		5x5 & 3.50 & 0.19 & 0.05  \\
		10x10 & 3.82 & 0.10 & 0.06  \\
		50x50 & 4.16 & 0.02 & 0.06  \\
		100x100 & 4.22 & 0.01 & 0.05  \\
		500x500 & 3.24 & 0.01 & 0.42  \\
	\end{tabular}
	\captionlistentry[experiment]{AbstractionSizeToCO, Table \ref{grid-ahda}, supermicro, grid:random60-0.45:bucket, pthread, newmaterial}

\end{table}

Note that while this experiment was run on a  a single multicore machine using Pthreads 
and low-level instructions (try\_lock) for moving states  among processors, 
communications overhead becomes an even more serious issue using interprocess communication (e.g. MPI) in distributed environments because the communication cost for each message is higher on such environments. 


\begin{observation}
SafePBNF and \AHDA{} outperform \ZHDA{} on the grid pathfinding problem, even though SafePBNF and \AHDA{} require more node expansions than \ZHDA{}. Communications overhead accounts for the poor performance of \ZHDA{} on grid pathfinding.
\end{observation}

\subsection{Summary of the Parallel Overheads for \ZHDA{} and \AHDA{}}


Table \ref{analysis-summary} summarizes the comparison of the Zobrist hashing based \ZHDA{} and structured abstraction based \AHDAP{} work distribution strategies on the sliding-tile puzzle and grid pathfinding domains.
As we showed in Section \ref{sec:experiments-order} and \ref{sec:hdavspbnf}, \ZHDA{} outperforms \AHDAP{} on sliding-tile puzzle domain because \AHDAP{} suffers from high SO.
On the other hand, \AHDAP{} outperforms \ZHDA{} on grid pathfinding because \ZHDA{} has high CO (Section \ref{sec:co}).
In summary, both \ZHDA{} and \AHDAP{} have clear weakness -- \ZHDA{} has no mechanism which explicitly seeks to reduce the amount of communication, whereas \AHDAP{} has no mechanism which explicitly minimizes load balancing.

\begin{table}[htb]
	\caption{Comparison of speedup, communication overhead, and search overhead of \ZHDA{} and \AHDAP{} on 15-puzzle, 24-puzzle, and grid pathfinding with 8 threads. CO: communication overhead, SO: search overhead. \ZHDA{} outperformed \AHDAP{} on the 15-puzzle and 24-puzzle, while \AHDAP{} outperformed \ZHDA{} on grid pathfinding.}
	\label{analysis-summary}
	\centering
	\begin{tabular}{l|rrr} \hline
		15-puzzle & speedup & CO & SO \\ \hline
		{\bf \ZHDA{}} & {\bf 5.10} & 0.86 & 0.03 \\
		\AHDAP{} & 3.90 & 0.22 & 0.13 \\ \hline \hline
		24-puzzle & speedup & CO & SO \\ \hline
		{\bf \ZHDA{}} & {\bf 6.28} & 0.85 & 0.04 \\
		\AHDAP{} & 4.20 & 0.38 & 0.14 \\ \hline \hline
		grid & speedup & CO & SO \\ \hline
		\ZHDA{} & 2.57 & 0.87 & 0.05 \\
		{\bf \AHDAP{}} & {\bf 4.22} & 0.01 & 0.05  \\ \hline
	\end{tabular}
	\captionlistentry[experiment]{SummaryOfAnalysis, Table \ref{analysis-summary}, supermicro, pthread, newmaterial}

\end{table}

\section{Abstract Zobrist Hashing(AZH)}
\label{sec:azh}

As we discussed in Section \ref{sec:analysis-of-parallel-overheads}, both search and communication overheads have a significant impact on the performance of HDA*, and methods that only address one of these overheads are insufficient.
\ZHDA{}, which uses  Zobrist hashing, assigns nodes uniformly to processors, achieving near-perfect load balance, but at the cost of incurring communication costs on almost all state generations.
On the other hand, abstraction-based methods such as PBNF and \AHDAP{} significantly reduce communication overhead by trying to keep generated states at the same processor as where they were generated, but this results in significant search overhead because all of the productive search may be performed at 1 node, while all other nodes are searching unproductive nodes which would not be expanded by A*.
Thus, we need a more balanced approach that simultaneously addresses both search and communication overheads.

{\it Abstract Zobrist hashing} (AZH) is a hybrid hashing strategy which 
augments the Zobrist hashing framework with the idea of projection from abstraction, incorporating the strengths of both methods.
The AZH value of a state, $AZ(s)$ is:
\begin{equation}
\label{eq:sz}
	AZ(s) := R[A(x_{0})] \; \xor \; R[A(x_{1})]\; \xor \; \cdots \; \xor \; R[A(x_{n})]
\end{equation}
where $A$ is a {\it feature projection function}, a many-to-one mapping from each raw feature to an {\it abstract feature}, and $R$ is a precomputed table 
for each abstract feature.

\begin{figure}[htb]
	\centering
	\subfloat[Zobrist hashing]{{\includegraphics[width=0.55\linewidth]{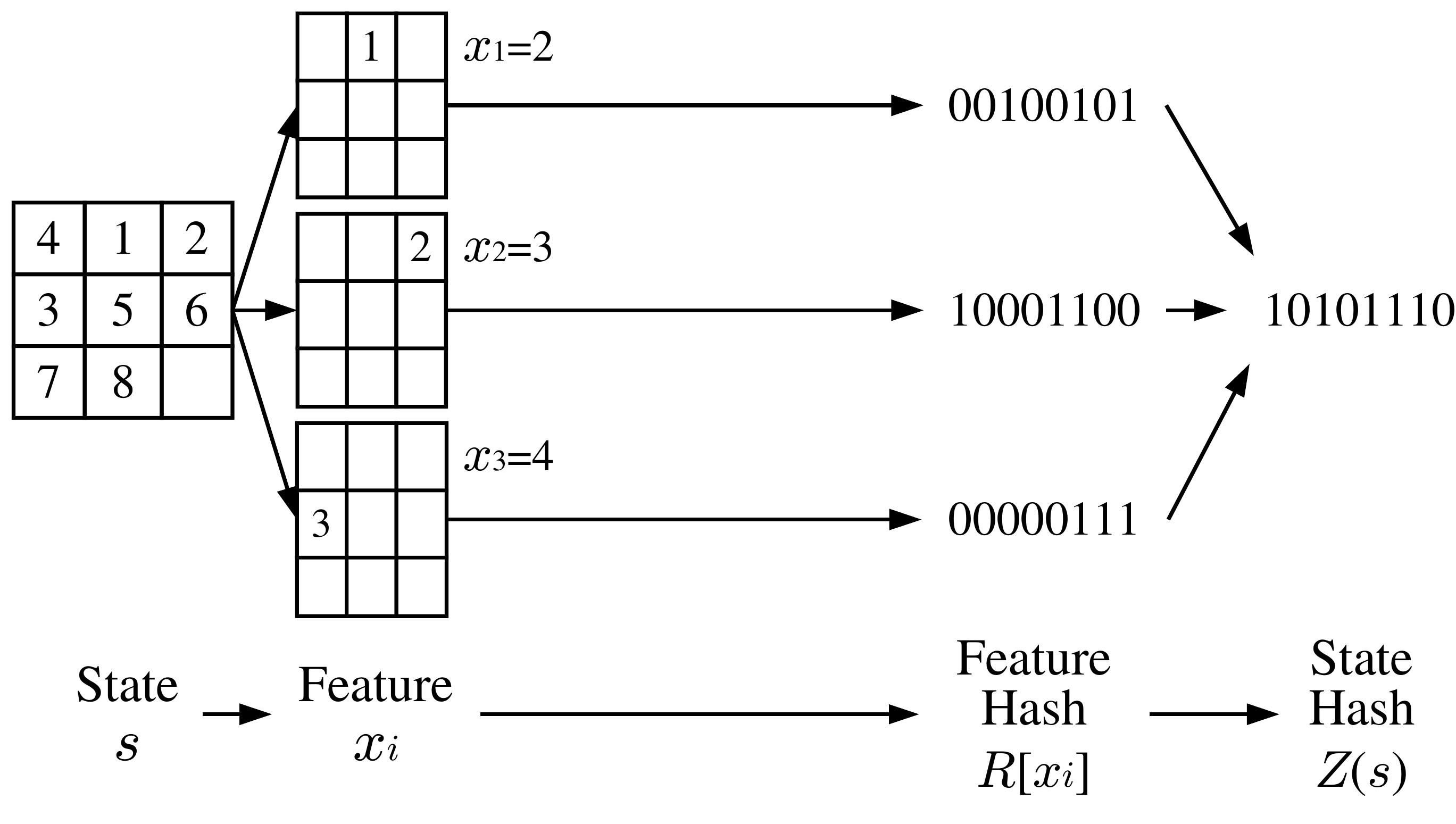}}}

	\subfloat[Abstract Zobrist hashing]{{\includegraphics[width=0.55\linewidth]{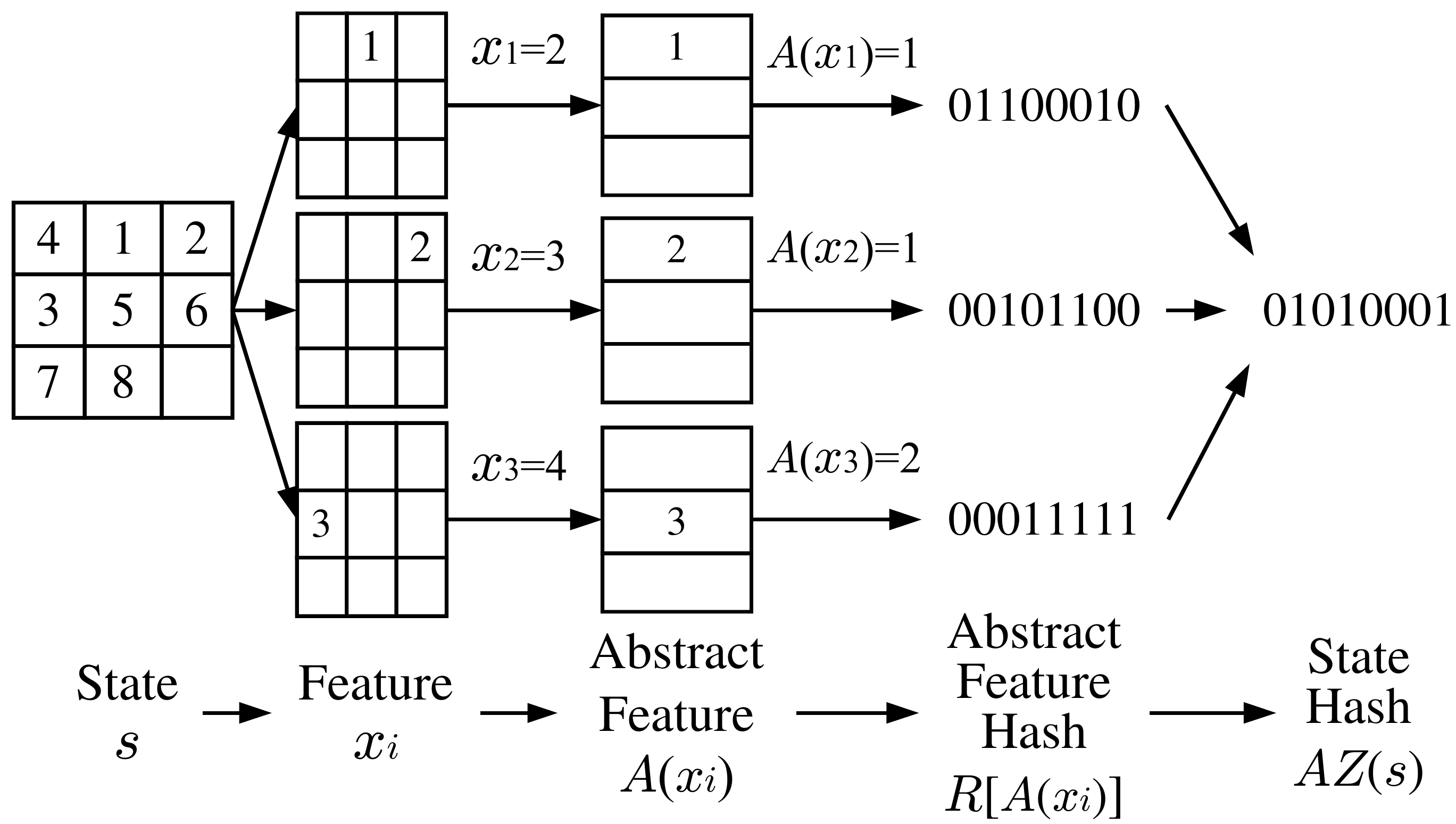}}}
	\caption{{\small Calculation of Abstract Zobrist Hash (AZH) value $AZ(s)$ for the 8-puzzle: 
State 
$s = (x_1, x_2,..., x_8)$, where $x_i = 1, 2,..., 9$ ($x_i = j$ means tile $i$ is placed at position $j$).
The Zobrist hash value of $s$ is the result of xor'ing a preinitialized random bit vector $R[x_{i}]$ for each feature (tile) $x_i$. AZH incorporates an additional step which projects features to abstract features (for each feature $x_i$, look up $R[A(x_{i})]$ instead of $R[x_{i}]$).}} 
	\label{fig:hash-calculation}
\end{figure}

Thus, AZH is a 2-level, hierarchical hash, where 
raw features are first projected to abstract features, and Zobrist hashing is applied to the abstract features.
In other words, we project state $s$ to an abstract state $s' = (${\small $A(x_{0}), A(x_{1}),...,A(x_{n})$}$)$, and $AZ(s) = Z(s')$.
Figure \ref{fig:hash-calculation} illustrates the computation of the AZH value for an 8-puzzle state.

AZH seeks to combine the advantages of both abstraction and Zobrist hashing.
Communication overhead is minimized by building abstract features that share the same hash value (abstract features are analogous to how abstraction projects state to abstract states), and load balance is achieved by applying Zobrist hashing to the abstract features of each state. 

Compared to Zobrist hashing, AZH incurs less CO due to abstract feature-based hashing.
While Zobrist hashing assigns a hash value to each node independently, AZH assigns the same hash value to all nodes that share the same abstract features for all features, reducing the number of node transfers. 
Also, in contrast to abstraction-based node assignment, which minimizes communications but does not optimize load balance and search overhead,
AZH seeks good load balance, because the node assignment considers
all features in the state, rather than just a subset.

\begin{algorithm}
	\Input{$F$: a set of features, $A$: a mapping from features to abstract features (abstraction strategy)}
	\For {each $a \in \{A(x) | x \in F\}$} {
		$R'[a] \leftarrow random()$\;
	}
	\For {each $x \in F$} {
		$R[x] \leftarrow R'[A(x)]$\;
	}
	{\bf Return} $R$
	\caption{Initialize \AZHDA{}}
	\label{alg:init-abstract-zobrist-hashing}
\end{algorithm}

AZH is simple to implement, requiring only an additional projection per feature compared to Zobrist hashing, and we can pre-compute this projection at initialization (Algorithm \ref{alg:init-abstract-zobrist-hashing}). Thus, there is no additional runtime overhead per node during the search. In fact, except for initialization, the same code to Zobrist hashing can be used (Algorithm \ref{alg:zobrist-hashing}). The projection function $A(x)$ can be generated either hand-crafted or automatically generated.
Following the notation of AHDA* in Section \ref{sec:ahda}, we denote AZHDA* with hand crafted feature abstraction as \AZHDA{}, where $A_{feature}$ stands for feature abstraction. 
The key difference of \AZHDA{} from \AHDAZ{} is that \AZHDA{} applies abstraction to each feature and applies Zobrist hashing to abstract features, whereas \AHDAZ{} applies abstraction to a state and applies Zobrist hashing to the abstract state. 

\subsection{Evaluation of Work Distribution Methods on Domain-Specific Solvers} 
\label{sec:experiments-comb}

We evaluated the performance of the following HDA* variants 
on several standard benchmark domains with different characteristics. 
\begin{itemize}
\item \AZHDA{}: HDA* using AZH
\item \ZHDA{}:  HDA* using Zobrist hashing \cite{kishimotofb09}
\item \AHDAP{}:  HDA* using Abstraction based work distribution \cite{burnslrz10}
\item \PHDA{}:  HDA* using a perfect hash function  \cite{burnslrz10}
\end{itemize}
The experiments were run on an \supermicro, using up to 16 cores.


The 15-puzzle experiments in Section \ref{sec:15-puzzle} incorporated enhancements from the more recent work by \citeauthor{burns2012implementing} \citeyear{burns2012implementing} to the code used in Section \ref{sec:so}, which is based on the code in \citeauthor{burnslrz10} \citeyear{burnslrz10}, 
which includes \PHDA{}, 
\AHDAP{}, and SafePBNF (we implemented 15-puzzle \ZHDA{} and \AZHDA{} as an extension of their code).

For the 24-puzzle and multiple sequence alignment (MSA), we used our own implementation of HDA* for overall performance (different from the code used in Section \ref{sec:hdavspbnf}), using the Pthreads library, 
try{\_}lock for asynchronous communication, and
the Jemalloc memory allocator \cite{evans2006scalable}. 
We implemented the open list as a 2-level bucket \cite{burns2012implementing} for the 15-puzzle and 24-puzzle, and a binary heap for MSA (binary heap was faster for MSA).

Note that although we evaluated \ZHDA{}, \AHDAP{}, and SafePBNF on the grid pathfinding problem in Section \ref{sec:analysis-of-parallel-overheads}, we do not evaluate \AZHDA{} on the grid pathfinding problem because in the case of grid pathfinding, the obvious feature projection function for \AZHDA{} corresponds to the abstraction used by \AHDAP{}.


\subsubsection{15-Puzzle}
\label{sec:15-puzzle}

\begin{figure}[htb]
	\centering
	\subfloat[15-puzzle \ZHDA{}]{\hspace{16pt}{\includegraphics[width=0.12\linewidth]{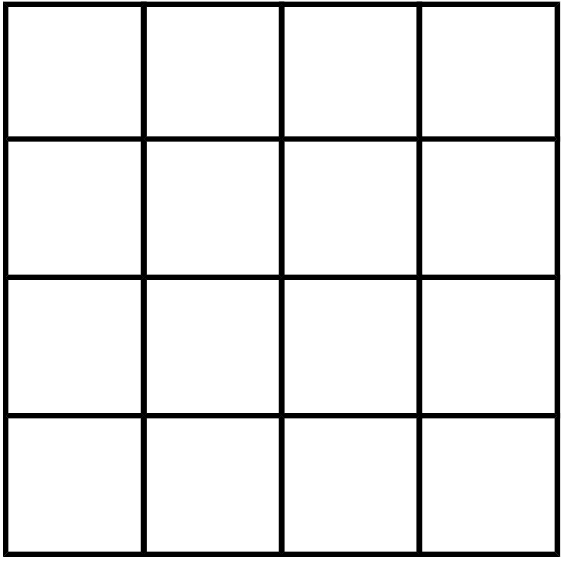}}\hspace{16pt}} \hspace{4pt}
	\subfloat[15-puzzle \AZHDA{}]{\hspace{32pt} \label{fig:15azh} {\includegraphics[width=0.12\linewidth]{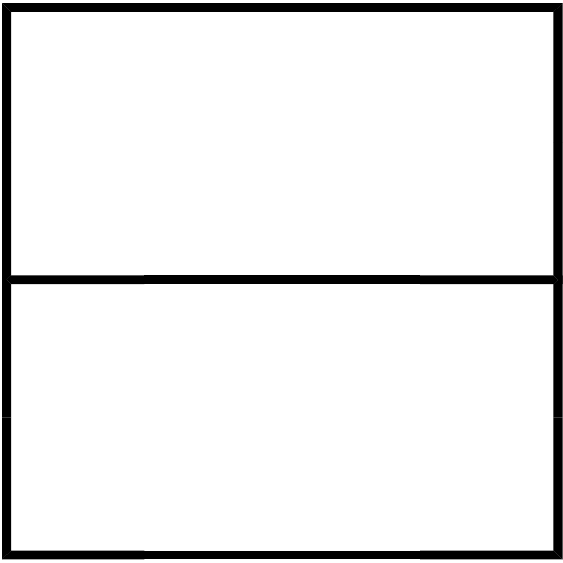} } \hspace{32pt}} \hspace{4pt}
	\subfloat[24-puzzle \ZHDA{}]{\hspace{16pt} {\includegraphics[width=0.12\linewidth]{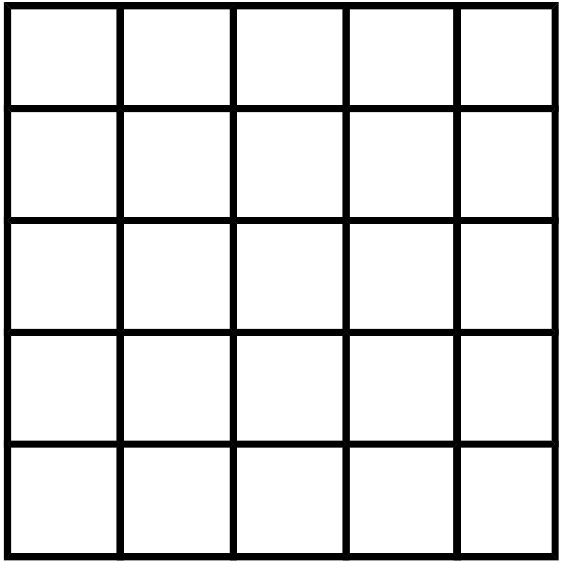}}\hspace{16pt}} \hspace{4pt}
	\subfloat[24-puzzle \AZHDA{}]{\hspace{32pt} \label{fig:24azh}{\includegraphics[width=0.12\linewidth]{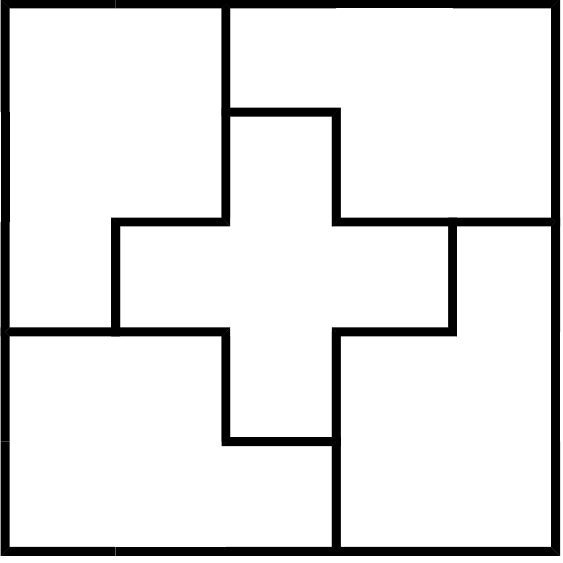}}\hspace{32pt}}

	\caption{The hand-crafted abstract features used by AZH for the 15 and 24-puzzle.} 
	\label{fig:structures}
\end{figure}

We solved 100 randomly generated instances with solvers using the Manhattan distance heuristic. These are not the same instances as the 100 instances used in  Section \ref{sec:so} because the solver used for this experiment was faster than the solver used in Section \ref{sec:so}\footnote{In Section \ref{sec:so}, the code is based on the code used in the work of \citeauthor{burnslrz10} \citeyear{burnslrz10}, while the code used in this section incorporated all of the enhancements from their more recent work on efficient sliding-tile solver code \cite{burns2012implementing}}, 
and some of the instances used in Section \ref{sec:so} were too easy for an evaluation of parallel efficiency.\footnote{This was intentional -- in Section \ref{sec:so}, we needed a distribution of instances which included easy instances to highlight the burst effect (Section \ref{sec:burst}) as well as for comparison with other methods (Section \ref{sec:hdavspbnf}).}   We selected instances which were sufficiently difficult enough to avoid the results being dominated by the initial startup overhead of the burst effect  (Section \ref{sec:burst}) --  
 sequential A* required an average of 52.3 seconds to solve these instances.
In addition to \AZHDA{}, \ZHDA{}, and \AHDAP{}, we also evaluated
SafePBNF \cite{burnslrz10} and \PHDA{}.
The projections $A(x_i)$ (abstract features) we used for AZH in \AZHDA{} are shown in Figure \ref{fig:15azh}.
The configurations for the other work distribution methods (\ZHDA{}, \AHDAP{}, SafePBNF, and \PHDA{}) were the same as in Section \ref{sec:so}.

\begin{figure}[htb]
	\centering
	\includegraphics[width=0.5\linewidth]{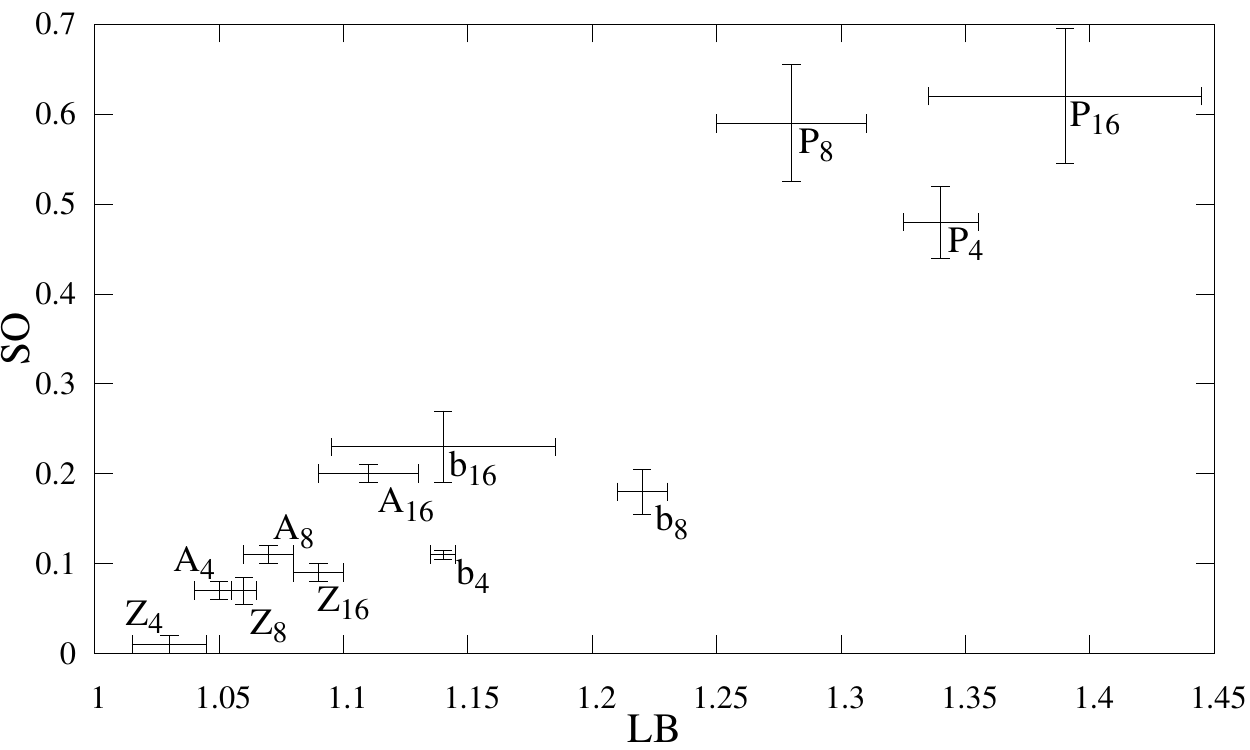}
	\captionof{figure}{
Load balance (LB) and search overhead (SO) on 100 instances of the 15-Puzzle for 4/8/16 threads. ``A'' = \AZHDA{}, ``Z'' = \ZHDA{}, ``b'' = \AHDAP{}, ``P'' = \PHDA{}, e.g., ``$\text{Z}_8$'' is the LB and SO for Zobrist hashing on 8 threads. 2-D error bars show standard error of the mean for both SO and LB.
}
	\label{fig:15puzzle-lb-so}
	\captionlistentry[experiment]{Combinatorial:LB-SO, Figure \ref{fig:15puzzle-lb-so}, supermicro, 15puzzle:random100difficult:bucket, pthread, AAAImaterial}

\end{figure}

\begin{figure}
	\centering
	\subfloat[15-puzzle: Efficiency
]{\label{fig:15puzzle-bucket-eff} {\includegraphics[width=0.45\linewidth]{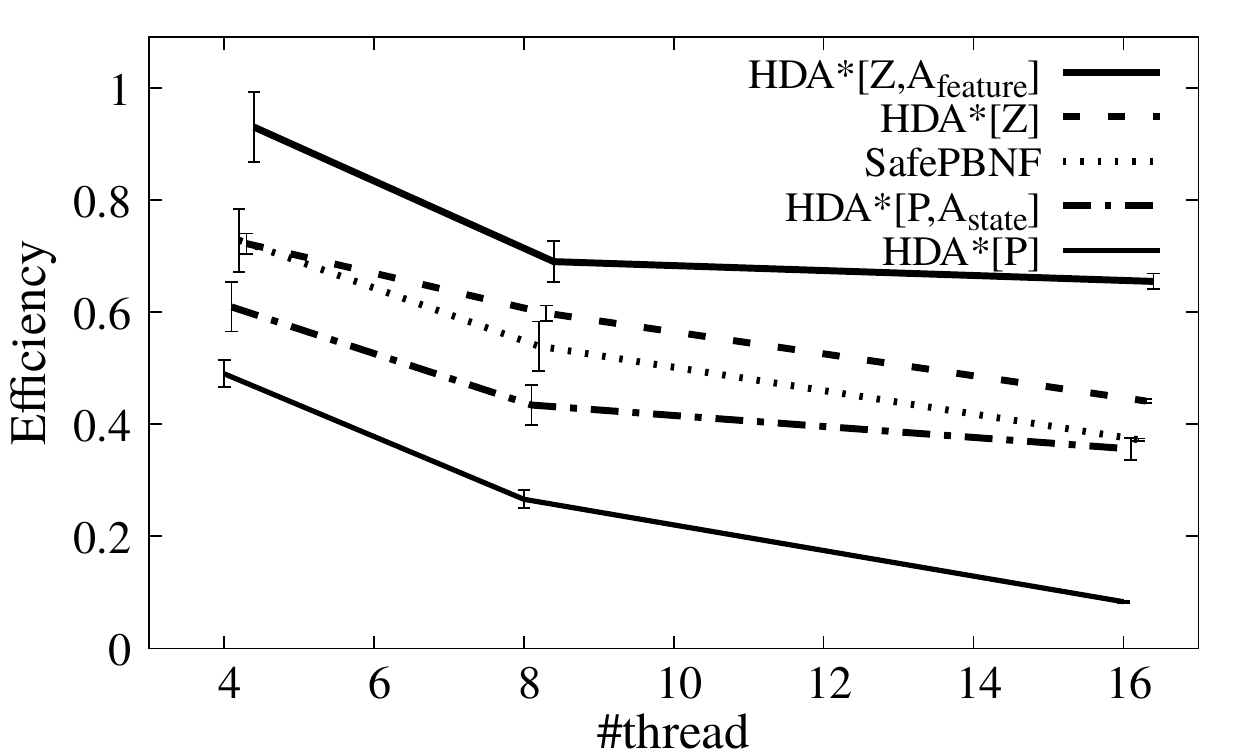}}} 
	\subfloat[15-puzzle: CO vs. SO
]{\label{fig:15puzzle-bucket-tradeoff}{\includegraphics[width=0.45\linewidth]{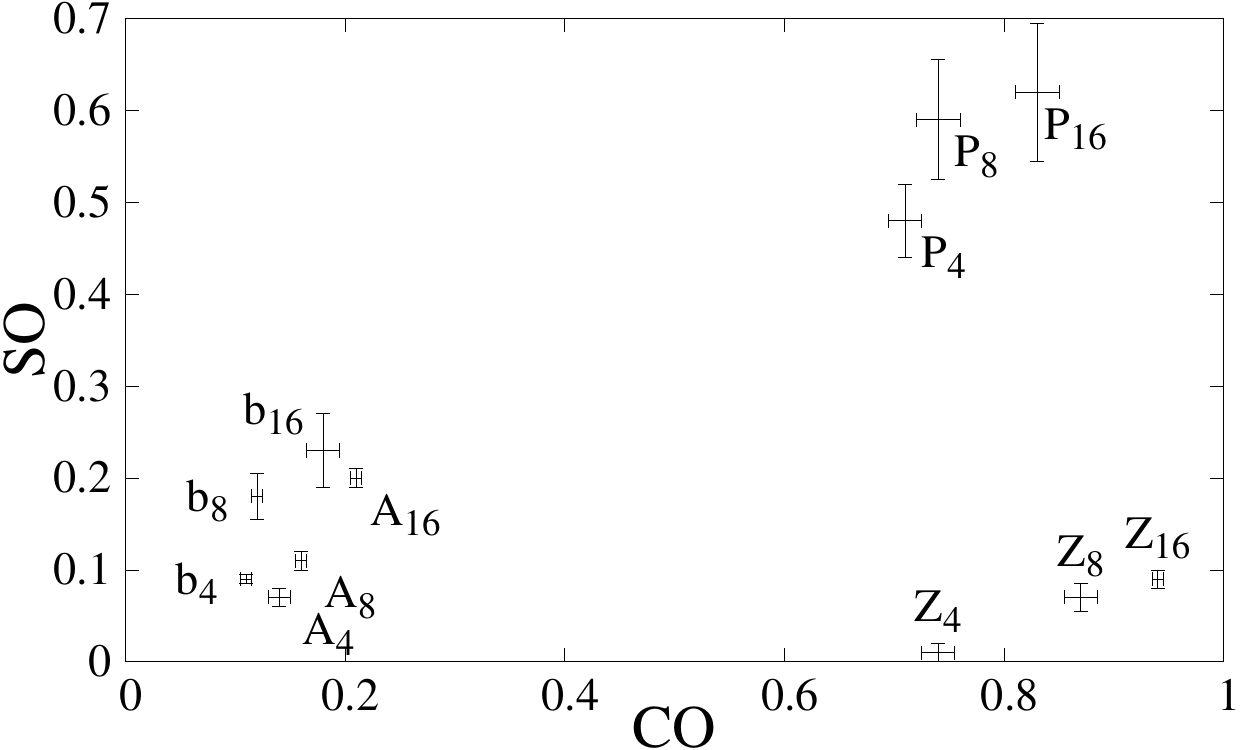}}}

        \subfloat[24-puzzle: Efficiency]{\label{fig:24puzzle-pdb-eff}{\includegraphics[width=0.45\linewidth]{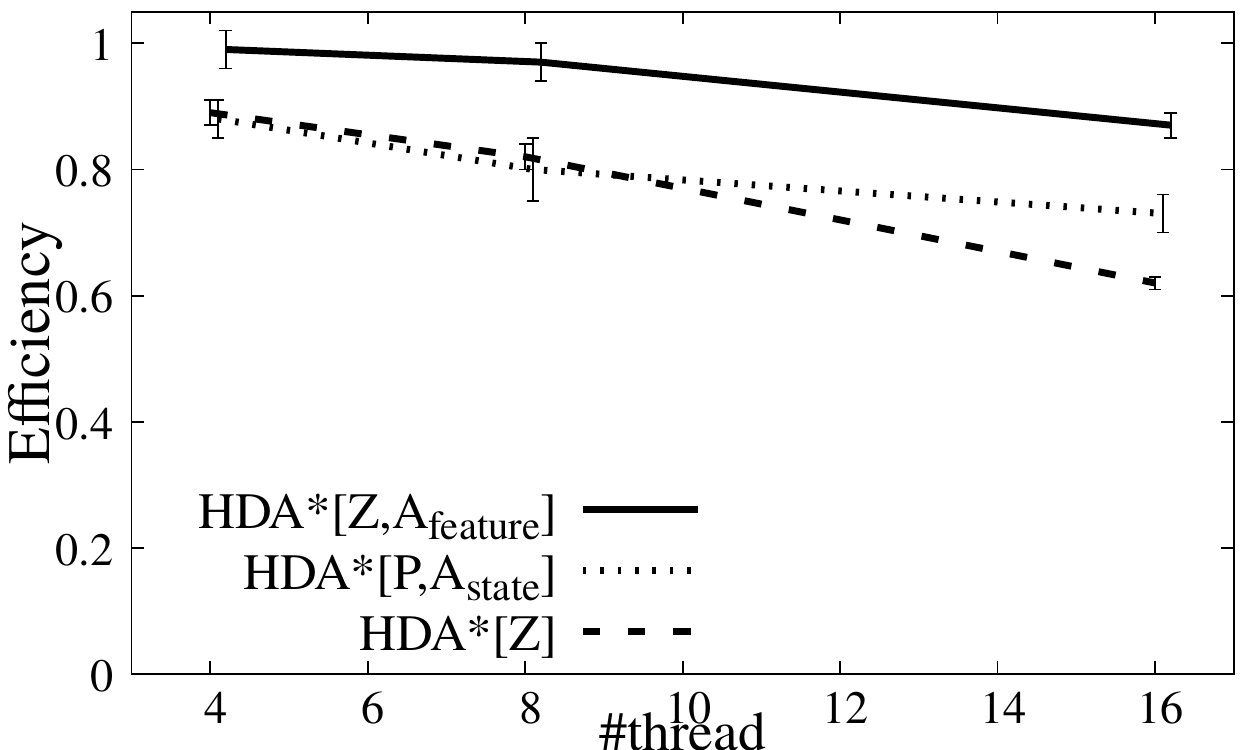}}}
	\subfloat[24-puzzle: CO vs. SO]{\label{fig:24puzzle-pdb-tradeoff}{\includegraphics[width=0.45\linewidth]{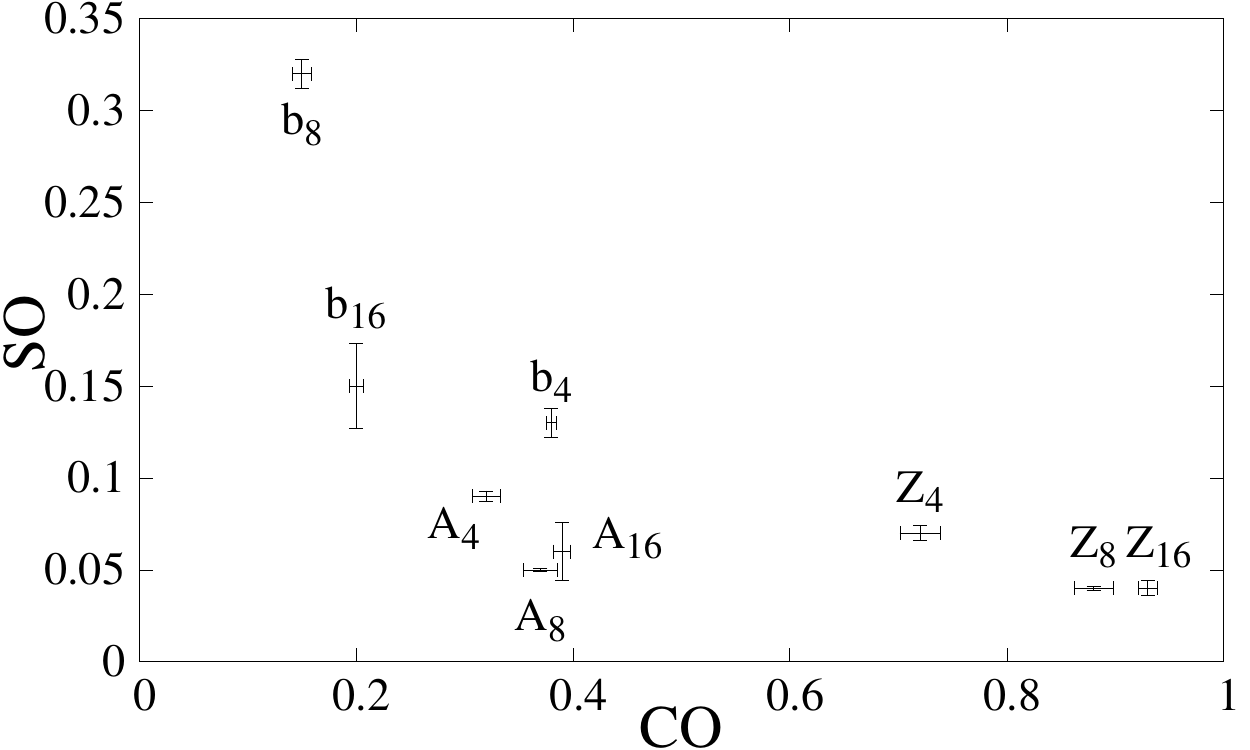}}}

	\subfloat[MSA: Efficiency]{\label{fig:msa-eff}{\includegraphics[width=0.45\linewidth]{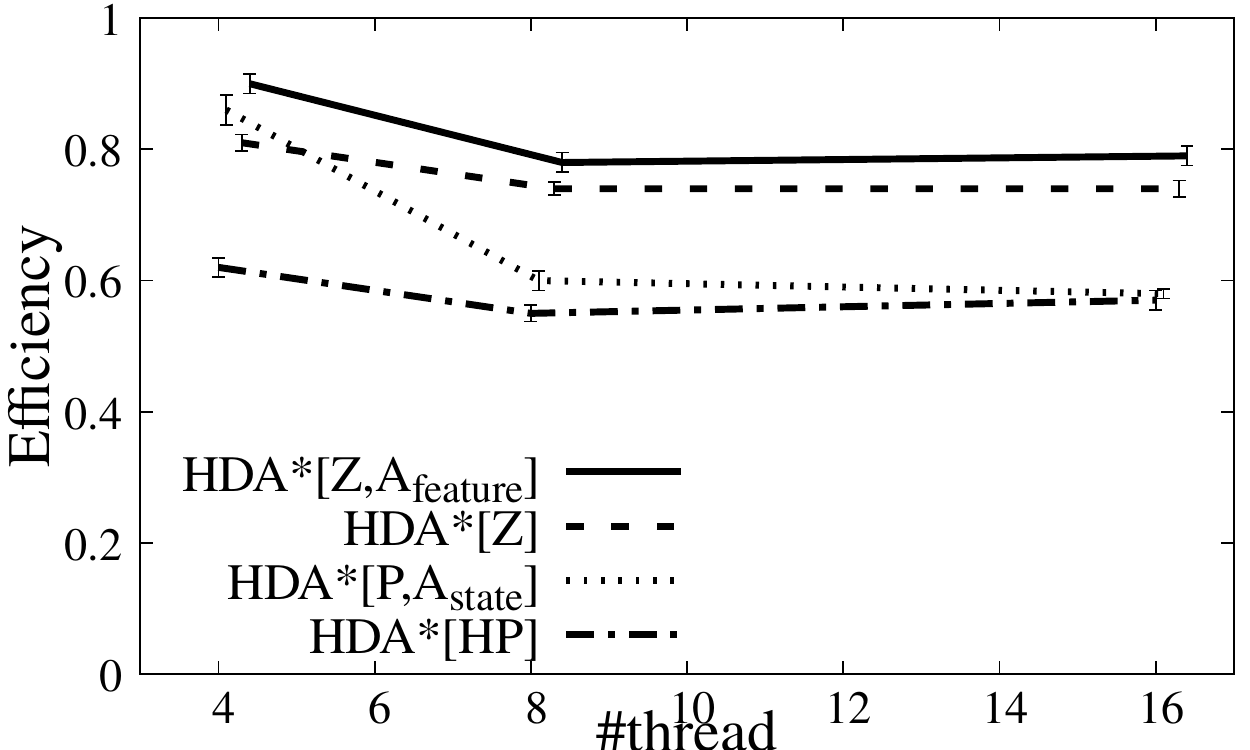}}}
	\subfloat[MSA: CO vs. SO]{\label{fig:msa-tradeoff}{\includegraphics[width=0.45\linewidth]{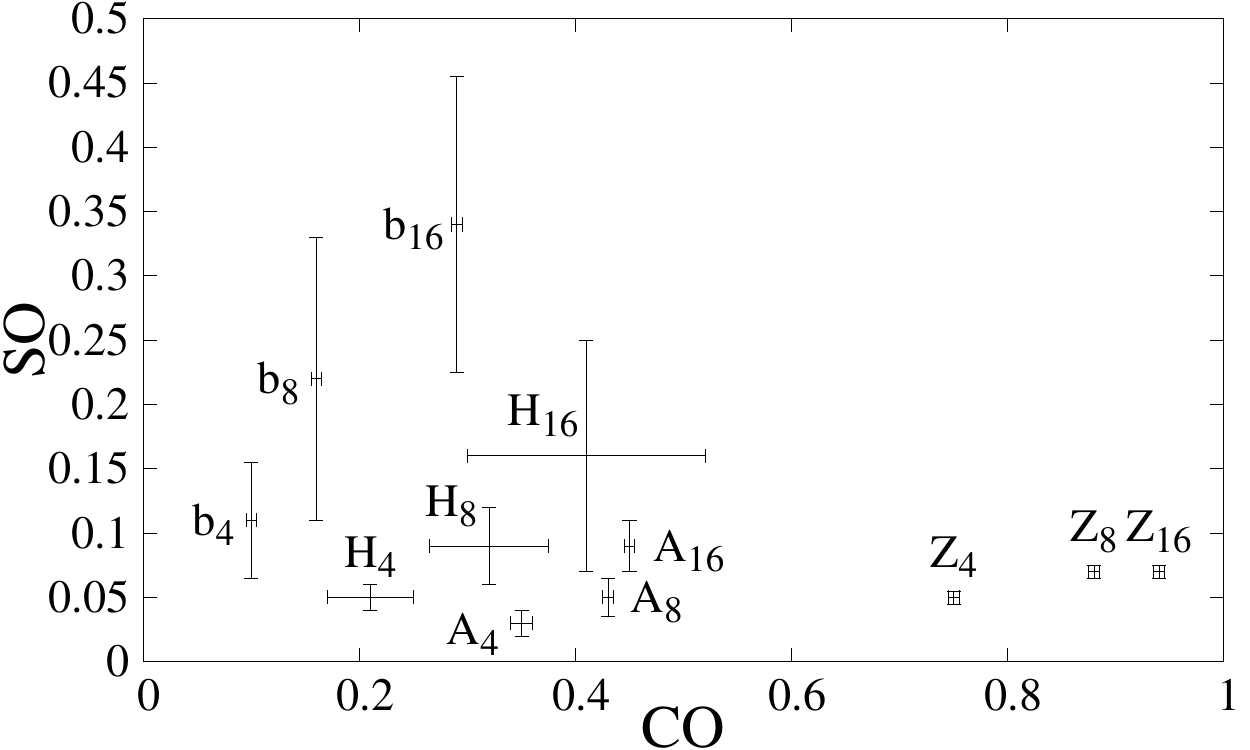}}}
	\caption{Efficiency ($=\frac{\mbox{speedup}}{\mbox{\# cores}}$), Communication Overhead (CO), and Search Overhead (SO) for 15-puzzle (100 instances), 24-puzzle (100 instances), and MSA (60 instances) on 4/8/16 threads. The open list  is implemented using a 2-level bucket for sliding-tile puzzle, and as a binary heap for MSA.
In the CO vs. SO plot, ``A'' = \AZHDA{} (AZHDA*), ``Z'' = \ZHDA{} (ZHDA*), ``b'' = \AHDAP{} (AHDA*), ``P'' = \PHDA{}, ``H'' = \HWD{}, e.g., ``$\text{Z}_8$'' is the CO and SO for Zobrist hashing on 8 threads.
Error bars show standard error of the mean.}
	\label{fig:results}
	\captionlistentry[experiment]{Combinatorial:eff-co-so, Figure \ref{fig:results}, supermicro, 15puzzle:random100difficult:bucket 24puzzle:random100:bucket msa:random+bali:heap, pthread, AAAImaterial}

\end{figure}


First, as discussed in Section \ref{sec:background}, high search overhead is correlated with load balance.
Figure \ref{fig:15puzzle-lb-so}, which shows the relationship between load balance and search overhead, indicates a very strong correlation between high load imbalance and search overhead. We discuss the relationship of load balance and search overhead in detail in Section \ref{sec:analysis}.

Figure \ref{fig:15puzzle-bucket-eff} shows the efficiency ($=\frac{\mbox{speedup}}{\mbox{\# cores}}$) of each method. 
\PHDA{} performed extremely poorly compared to all other HDA* variants and SafePBNF.
The reason is clear from Figure \ref{fig:15puzzle-bucket-tradeoff}, which shows the communication and search overheads.
\PHDA{} has both extremely high search overhead and communication overhead compared to all other methods.
This shows that the hash function used by \PHDA{} is not well-suited as a work distribution function.

\AHDAP{} had the lowest CO among HDA* variants (Figure \ref{fig:15puzzle-bucket-tradeoff}), and significantly outperformed \PHDA{}.
However, \AHDAP{} has worse LB than \ZHDA{}  (Figure \ref{fig:15puzzle-lb-so}), resulting in higher SO.
For the 15-puzzle, this tradeoff is not favorable for \AHDAP{}, and Figures \ref{fig:15puzzle-bucket-eff}-\ref{fig:15puzzle-lb-so} show that \ZHDA{}, which has significantly better LB and SO, outperforms \AHDAP{}. %

According to Figure \ref{fig:15puzzle-bucket-eff}, SafePBNF outperforms \AHDAP{}, and is comparable to \ZHDA{} on the 15-puzzle.
Although our definition of communication overhead does not apply to SafePBNF,
SO for SafePBNF was comparable to \AHDAP{}, 0.11/0.17/0.24 on 4/8/16 threads.  


\AZHDA{} significantly outperformed \ZHDA{}, \AHDAP{}, and SafePBNF.
As shown in Figure \ref{fig:15puzzle-bucket-tradeoff}, although \AZHDA{} had higher SO than \ZHDA{} and higher CO than \AHDAP{}, 
it achieved a balance between these overheads, resulting in high overall efficiency.
The tradeoff between CO and SO depends on each domain and instance.  
By tuning the size of the abstract feature, we can choose a suitable tradeoff.

\subsubsection{24-Puzzle}
\label{sec:24puzzle}
We generated a set of 100 random instances that could be solved by A* within 1000 seconds.
For the same reason as with the 15-puzzle experiments above in Section \ref{sec:15-puzzle}, these are different from the 24-puzzle instances used in \ref{sec:hdavspbnf}.
We chose the hardest instances solvable given the memory limitation (128GB).
The average runtime of sequential A* on these instances was 219.0 seconds.
The average solution length of our 24-puzzle instances was 92.9 (the average solution length in \citeR{Korf2002} was 100.8).
We used a disjoint pattern database heuristic \cite{Korf2002}.
For the  sliding-tile puzzle, the disjoint pattern database heuristic is much more efficient than Manhattan distance, thus the average walltime of 24-puzzle with disjoint pattern database heuristic was much faster than that of 15-puzzle with Manhattan distance heuristic, even though the 24-puzzle search space is much larger than the 15-puzzle search space. 
Figure \ref{fig:24azh} shows the feature projections we used for 24-puzzle.
For \ZHDA{} and \AHDAP{}, we used same configurations as  in Section \ref{sec:hdavspbnf}.
The abstraction used by SafePBNF ignores the numbers on all of the tiles except tiles 1,2,3,4, and 5.\footnote{We tried (1) ignoring all tiles except blank and tiles 1-2, (2) ignoring all tiles except blank and tiles 1-3, (3) ignoring all tiles except blank and tiles 1-4, (4) ignoring all tiles except tiles 1-3, (5) ignoring all tiles except tiles 1-4, (6) ignoring all tiles except tiles 1-5, and  chose (6), the best performer.}

Figure \ref{fig:24puzzle-pdb-eff} shows the efficiency of each method.
As with the 15-puzzle, \AZHDA{} significantly outperformed \ZHDA{} and \AHDAP{}, and
Figure \ref{fig:24puzzle-pdb-tradeoff} shows that 
as with the 15-puzzle, \ZHDA{} and \AHDAP{} succeed in mitigating only one of the overheads (SO or CO). In contrast, \AZHDA{} outperformed both \ZHDA{} and \AHDAP{}, as its SO was comparable to that of \ZHDA{} while its CO was roughly equal to that of \AHDAP{}.

\subsubsection{Multiple Sequence Alignment}
\label{sec:msa}

Multiple Sequence Alignment (MSA) is the problem of finding a minimum-cost alignment of a set of DNA or amino acid sequences by inserting gaps in each sequence.
MSA can be solved by finding the min-cost path between corners in a $n$-dimensional grid, where each dimension corresponds to the position of each sequence.
We used 60 benchmark instances, consisting of 10 actual amino acid sequences from BAliBASE 3.0 \cite{Thompson2005}, and  50 randomly generated instances. 
The BAliBASE instances we used are: BB12021, BB12022, BB12036, BBS11010, BBS11026, BBS11035, BBS11037, BBS12016, BBS12023, BBS12032.
We generated random instances by (1) select number of sequences $n$ from 4 to 9 uniformly randomly, (2) For each sequence select a number of acids $l$ from $5000/n  \cdot 0.9 < l < 5000/n \cdot 1.1$, (3) choose each acid uniformly random from 20 acids.
Edge costs are based on the PAM250 matrix score 
with gap penalty 8 \cite{pearson1990}.
Since there was no significant difference between the behavior of HDA* among actual and random instances, we report the average of all 60 instances. 
We used the pairwise sequence alignment heuristic \cite{korf2005frontier}.

The features for Zobrist hashing and AZH were the positions of each sequence. 
For AZH, we grouped  4 positions per row into an abstract feature.
Thus, with $n$ sequences, nodes in the $n$-dimensional hypercube with edge length $l$ share the same hash value.
The abstraction used by \AHDAP{} only considers the position of the longest sequence and ignores the others.
We chose this abstraction for \AHDAP{} as it performed the best among (1) only considering the position of the longest sequence, 
(2) only considering the two longest sequences, and (3) only considering the three longest sequences.
We also evaluated the performance of Hyperplane Work Distribution \cite{Kobayashi2011evaluations}. 
\ZHDA{} suffers from node reexpansion in non-unit cost domains such as MSA. 
Hyperplane work distribution seeks to reduce node reexpansions by mapping the $n$-dimension grid to hyperplanes (denoted as \HWD{}). 
For \HWD{}, we determined the plane thickness $d$ using the tuning method in \citeauthor{Kobayashi2011evaluations}  \citeyear{Kobayashi2011evaluations} where $\lambda=0.003$, which yielded the best performance among 0.0003, 0.003, 0.03, and 0.3.

Figure \ref{fig:msa-eff} compares the efficiency of each method, and 
Figure \ref{fig:msa-tradeoff} shows the CO and SO.
\AZHDA{} outperformed the other methods.
With 4 or 8 threads, \AZHDA{}  had smaller SO than \ZHDA{}.
This is because like \HWD{}, \AZHDA{} reduced the amount of duplicated nodes in some domains compared to \ZHDA{}.
%
%
Our MSA solver expands 300,000 nodes/second, which is relatively slow compared to, e.g., our 24-puzzle solver, which expands 1,400,000 node/sec. 
When node expansions are slow, 
the relative importance of CO decreases, and SO has a more significant impact on performance in MSA than in the 15/24-Puzzles.
Thus, \AHDAP{}, which incurs higher SO, did not perform well compared to \ZHDA{}.
\HWD{} did not perform well, but it was designed for large-scale, distributed search,
and we observed \HWD{} to be more efficient on difficult instances than on easier instances -- it is included 
in this evaluation 
only to provide another point of reference for evaluating \ZHDA{} and \AZHDA{}.

\subsubsection{Node Expansion Order of \AZHDA{}}
\label{sec:azh-order}

\begin{figure}[htb]
	\centering
%
%
	\centering
	\subfloat[\AZHDA{} on a difficult instance with 8 threads.] {\label{fig:order_azh_difficult}{\includegraphics[width=0.3\linewidth]{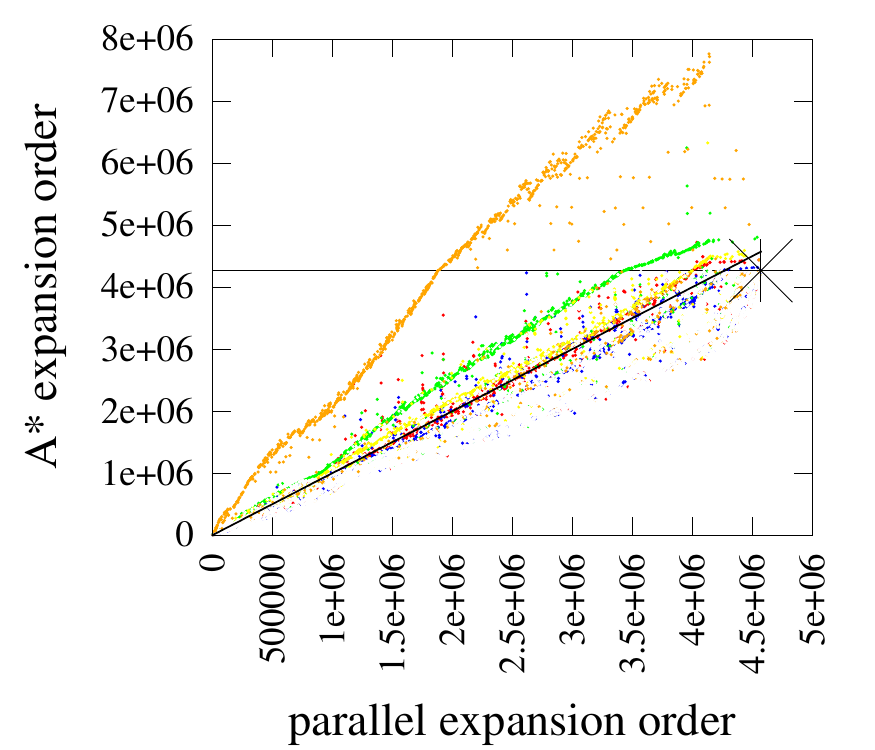} }}\hspace{10pt}%
	\subfloat[\ZHDA{} on a difficult instance with 8 threads (copy of Figure \ref{fig:order_z_difficult}).] {\label{fig:order_z_difficult_2}{\includegraphics[width=0.3\linewidth]{figures/order/z_difficult.pdf} }}\hspace{10pt}%
	\subfloat[\AHDAP{} on a difficult instance with 8 threads (copy of Figure \ref{fig:order_a_difficult}).] {\label{fig:order_a_difficult_2}{\includegraphics[width=0.3\linewidth]{figures/order/a_difficult.pdf} }}

	\subfloat {{\includegraphics[width=0.1\linewidth]{figures/order/legend1.pdf} }}
	\subfloat{{\includegraphics[width=0.1\linewidth]{figures/order/legend2.pdf} }}
	\subfloat{{\includegraphics[width=0.12\linewidth]{figures/order/legend3.pdf} }} 
	\caption{Comparison of \AZHDA{} node expansion order vs. sequential A* node expansion order on a {\bf difficult instance of the 15-puzzle} with 8 threads. The average node expansion order divergence scores for difficult instances are \ZHDA{}: $\bar{d}=10330.6$, \AHDAP{}: $\bar{d}=245818$, \AZHDA{}: $\bar{d}=76932.2$. AZHDA{} has a  bigger band effect than \ZHDA{}, but smaller than \AHDAP{}. Although the band of \AZHDA{} appears to be as large as \AHDAP{}, the actual divergence score $\bar{d}$ is higher on \AHDAP{} as \AHDAP{} expands more nodes.}
	\label{fig:order_easy_2}
\end{figure}

In Section \ref{sec:experiments-order}, 
in order to  see why search overhead occurs in HDA* and PBNF, 
we analyzed how the node expansion order of parallel search diverges from that of sequential A*.
Figure \ref{fig:order_easy_2} shows the expansion order of \AZHDA{} on a difficult instance (\ZHDA{} and \AHDAP{} are included for comparison).
\AZHDA{} has a bigger band effect than \ZHDA{}, but smaller than that of \AHDAP{}. 
The average divergence of nodes for difficult instances are \ZHDA{}: $\bar{d}=10330.6$, \AHDAP{}: $\bar{d}=245818$, \AZHDA{}: $\bar{d}=76932.2$.
Note that although the band effect of \AZHDA{} in Figure \ref{fig:order_azh_difficult} appears to be as large as the band effect of \AHDAP{} in Figure \ref{fig:order_a_difficult}, \AHDAP{} actually has a significantly higher divergence score $\bar{d}$ ($\bar{d}=245818$) than  \AZHDA{} ($\bar{d}=76932.2$), 
because \AHDAP{} expanded more nodes ($>$5,000,000 nodes) than \AZHDA{} ($>$4,500,000 nodes).

\subsection{Automated, Domain Independent Abstract Feature Generation} 
\label{sec:domain-independent}

In Section \ref{sec:experiments-comb}, we evaluated hand-crafted, domain-specific feature projection functions for instances of the HDA* framework (\ZHDA{}, \PHDA{}, \AHDA{}, \AZHDA{}), and  showed that AZH outperformed previous methods. Next, we turn our focus to fully automated, domain-independent methods for generating feature projection functions which can be used when a formal model of a domain (such as PDDL/SAS+ for classical planning) is available.

We now discuss domain-independent methods for work distribution. Table \ref{grand-scheme} summarizes the previously proposed methods and their abbreviations.

\begin{table}[htb]
	\caption{Comparison of {\bf previous  automated domain-independent feature generation methods} for HDA*. 
CO: communication overhead, SO: search overhead, ``optimized'': the method explicitly tries to (approximately) optimizes the overhead. ``ad hoc'': the method seeks to mitigate the overhead but without an explicit objective function. ``not addressed'': the method does not address the overhead.}
	\label{grand-scheme}
	\centering

	\begin{tabular}{rl|cc} \hline
		abbreviation & method & CO & SO \\ \hline
		 FAZHDA* & \FAZHDA{}		    & ad hoc & ad hoc \\
		& (Sec. \ref{sec:fluencyafg})  \cite{jinnai2016automated} \\ \hline
		 GAZHDA* & \GAZHDA{} & ad hoc & ad hoc \\
		& (Sec. \ref{sec:greedyafg}) \cite{jinnai2016automated} \\ \hline
		OZHDA* & \OZHDA{}  & ad hoc & ad hoc \\ 
		& (Sec. \ref{sec:zhda})  \cite{jinnai2016automated} \\ \hline
		DAHDA* & \DAHDA{} & optimized  & not  \\
		& (Sec. \ref{sec:ahda}, Appendix \apDAHDA{})  \cite{jinnai2016automated} & & addressed \\ \hline
		AHDA*  & \AHDAZSSD{} & optimized  & not    \\
		& (Sec. \ref{sec:ahda}) \cite{burnslrz10} & & addressed\\ \hline
		ZHDA*  & \ZHDA{} & not    & optimized  \\ 
		& (Sec. \ref{sec:zhda})  \cite{kishimotofb09} & addressed \\ \hline
	\end{tabular}

\end{table}

For \ZHDA{}, automated domain-independent feature generation for classical planning problems represented in the SAS+ representation \cite{backstrom1995complexity} is straightforward \cite{kishimotofb13}.
For each possible assignment of value $k$ to variable $v_i$ in a SAS+ representation, e.g., $v_i=k$, there is a binary proposition $x_{i,k}$ (i.e., the corresponding STRIPS propositional representation).
Each such proposition $x_{i,k}$ is a feature to which a randomly generated bit string is assigned, and the Zobrist hash value of a state can be computed by xor'ing the propositions that describe the state, as in Equation \ref{eq:zobrist}.

For AHDA*, the abstract representation of the state-space can be generated by ignoring some of the features (SAS+ variables) and using the rest of the features to represent the abstraction. 
\citeauthor{burnslrz10} \citeyear{burnslrz10} used the greedy abstraction algorithm in \citeauthor{Zhou2006} \citeyear{Zhou2006} to select the subset of features, which we refer to as  SDD abstraction. It adds one atom group to the abstract graph at a time, choosing the atom group which minimizes the maximum out-degree of the abstract graph, until the graph size (number of abstract nodes) reaches the threshold given by a parameter. 
As we saw in Section \ref{sec:experiments-comb}, the hashing strategy for abstract state has little effect on the performance. We used the implementation of AHDA* with Zobrist hashing and SDD abstraction (\AHDAZSSD{}). 

For AZHDA* (\AZHDA{}), the feature projection function, which generates abstract features from raw features,
plays a critical role in determining the performance of AZHDA*, because AZHDA* relies on the feature projection in order to reduce communications overhead.
In this section, we discuss two methods to automatically generate the feature projection function for AZH: 
(1) Greedy abstract feature generation (GreedyAFG), which partitions each {\it domain transition graph} (DTG) into 2 abstract features, 
and (2) Fluency-based abstract feature generation (FluencyAFG), an extension of GreedyAFG which filters the DTGs to partition according to a fluency-based criterion.
GreedyAFG and FluencyAFG seek to generate efficient feature projection functions without an explicit model of what to optimize.
Further details on GreedyAFG and FluencyAFG can be found in our previous conference paper \cite{jinnai2016automated}.

\subsubsection{Greedy Abstract Feature Generation (GAZHDA*)}
\label{sec:greedyafg}

{\it Greedy abstract feature generation} (GreedyAFG) is a simple, domain-independent  abstract feature generation method, which partitions each feature into 2 abstract features \cite{jinnai2016structured}.
GreedyAFG  first identifies {\it atom groups} \cite{edelkamp2001planning} and its domain transition graph (DTG). 
Atom group is a set of mutually exclusive propositions from which exactly one will be true for each reachable state, e.g., the values of a 
SAS+ multi-valued variable \cite{backstrom1995complexity}.
GreedyAFG maps each atom group $X$ into 2 abstract features $S_1$ and $S_2$, based on $X$'s undirected DTG (nodes are values, edges are transitions), 
as follows: (1) 
assign the minimal degree node (node sharing least number of edges with other nodes) to $S_1$; (2) greedily add to $S_1$ the unassigned node which shares the most edges with nodes in $S_1$; (3) while $|S_1| < |X|/2$ repeat step 2; 
(4) assign all unassigned nodes to $S_2$ (Algorithm \ref{alg:greedy-abstract-feature-generation}).
Due to the loop criterion in step 3, this procedure guarantees a perfectly balanced bisection of the DTGs, i.e.,  $|S_2| \leq |S_1| \leq |S_2| + 1$,  
so this tends to lead to good load balance with respect to this DTG.

$A(x_i)$ in Equation \ref{eq:sz} corresponds to the mapping from $x_i$ to $S_1, S_2$, and $R_i$ is defined over $S_1$ and $S_2$.
We denote GAZHDA* as \GAZHDA{}, as it applies feature abstraction (FA) by cutting DTGs using GreedyAFG.

\begin{algorithm}
	\Input{$X$, an atom group}
	Assign the minimal degree node (node sharing least number of edges with other nodes) to $S_1$\;
	\While {$|S_1| < |X|/2$} {
		Greedily add to $S_1$ the unassigned node which shares the most edges with nodes in $S_1$\;	
	}
	Assign all unassigned nodes to $S_2$\;
	{\bf Return} ($S_1$, $S_2$)\;
	\caption{Greedy Abstract Feature Generation}
	\label{alg:greedy-abstract-feature-generation}
\end{algorithm}



\subsubsection{Fluency-Dependent Abstract Feature Generation (FAZHDA*)}
\label{sec:fluencyafg}

Since the hash value of the state changes if any abstract feature value changes, 
GreedyAFG fails to prevent high CO when any abstract feature changes its value very frequently, e.g., in the standard \pddl{blocks} domain, every operator in the domain changes the value of the SAS+ variable representing the state of the robot's hand  (\pddl{handempty} $\leftrightarrow$ \pddl{not-handempty}).
{\it Fluency-dependent abstract feature generation} (FluencyAFG) overcomes this limitation 
\cite{jinnai2016automated}.
The {\it fluency} of a variable $v$ is the number of ground actions which change the value of the $v$ divided by the total number of ground actions in the problem.
By ignoring variables with high fluency, FluencyAFG was shown to be quite successful in reducing CO and increasing speedup compared to GreedyAFG.

\begin{figure}[htb]
	\centering
	\subfloat[GreedyAFG]{\includegraphics[width=0.45\columnwidth]{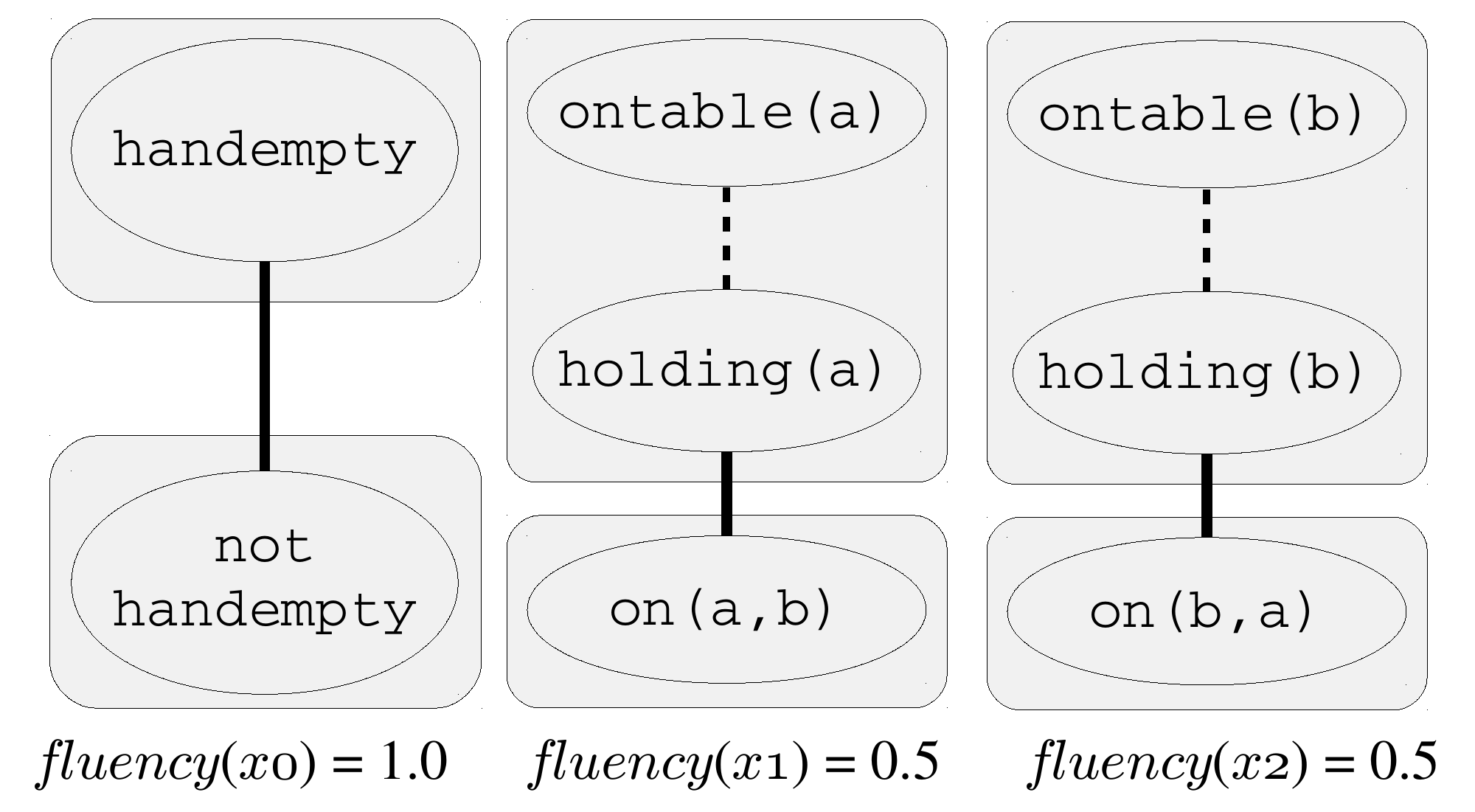}} \hspace{10pt}
	\subfloat[FluencyAFG]{\includegraphics[width=0.45\columnwidth]{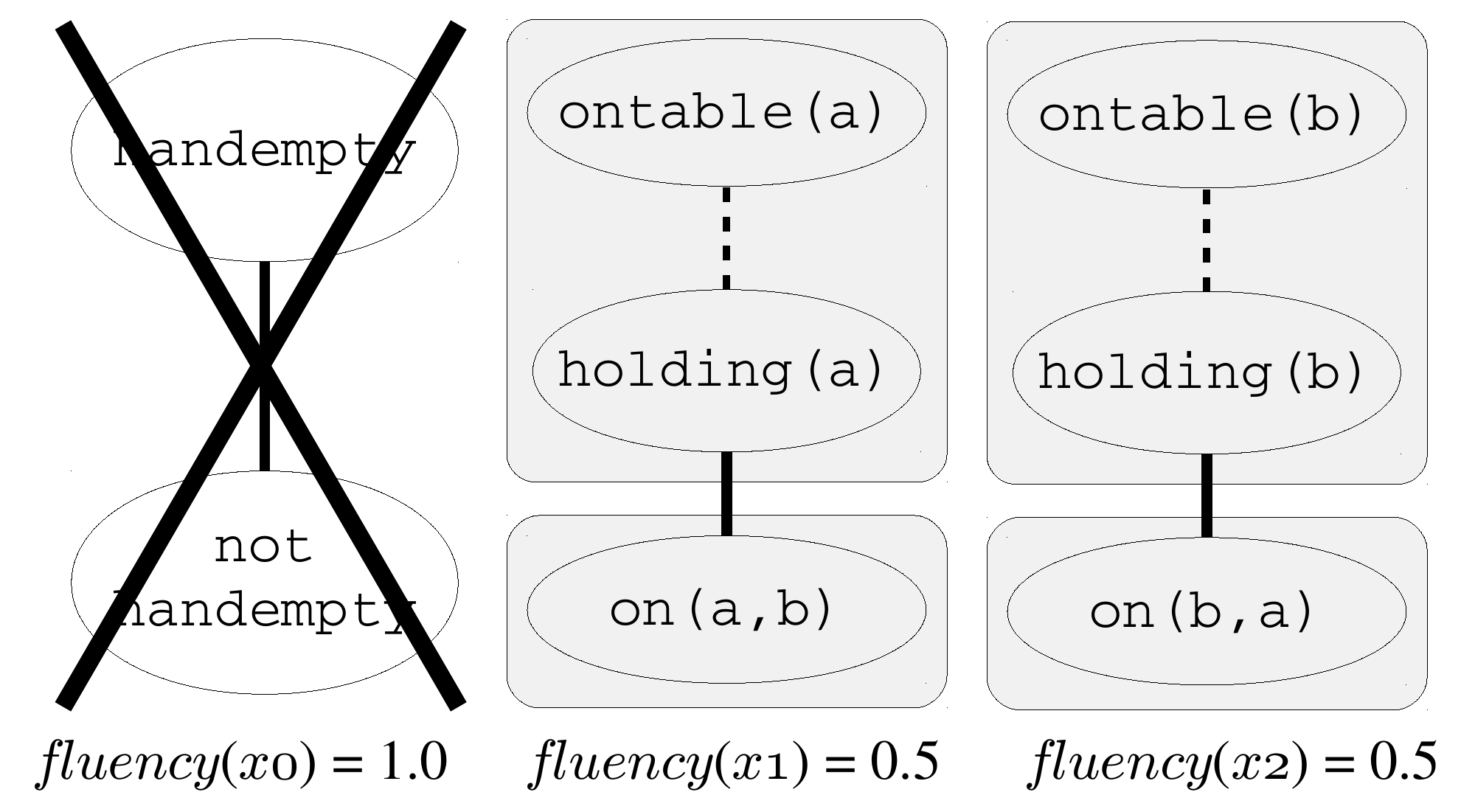}}

	\caption{Greedy abstract feature generation (\simple{}) and Fluency-dependent abstract feature generation (FluencyAFG) applied to blocksworld domain. The hash value for a state $s=(x_0,x_1,x_2)$ is given by $AZ(s) = R[A(x_0)]\;\xor\;R[A(x_1)]\;\xor\;R[A(x_2)]$. Grey squares are abstract features $A$ generated by \simple{}, so all propositions in the same square have same hash value (e.g. $R[A(\pddl{holding(a)})] = R[A(\pddl{ontable(a)})]$). $fluency(x_0) = 1$ since all actions in the blocks world domain change its value. In this case, any abstract features based on the other variables are rendered useless,  as all actions change $x_0$ and thus change the state's hash value. 
In this example, Fluency-dependent AFG will filter $x_0$ before calling \simple{} to compute abstract features based on the remaining variables (thus $AZ(s) = R[A(x_1)]\; \xor \; R[A(x_2)]$).} 
	\label{fig:feature-based}
\end{figure}

A problem with fluency is that 
in the AZHDA* framework, CO is associated with a change in value of an abstract feature, not the feature itself. However, FluencyAFG is based on the frequency with which features (not abstract features) change. This leads FluencyAFG to exclude variables from consideration unnecessarily, making it difficult to achieve good LB (in general, the more variables are excluded, the more difficult it becomes to reduce LB). 
Figure \ref{fig:feature-based} shows how fluency-based filtering is applied to the \pddl{blocks} domain.
The process of fluency-based filtering which ignores a subset of features can be considered an instance of abstraction. Therefore, we denote FAZHDA* as \FAZHDA{}, as it first applies fluency-based abstraction, followed by GreedyAFG. 

\section{A Graph Partitioning-Based Model for Work Distribution}
\label{sec:implicit-graph-partitioning}

Although GAZHDA* and FAZHDA*, the domain-independent abstract feature generation methods discussed in Section \ref{sec:domain-independent},
seek to reduce communications overhead compared to \ZHDA{}, they are not based on an 
explicit model which enables the prediction of the actual communications overhead achieved during the search.
Furthermore, the impact of these methods on search overhead is completely unspecified, and thus, it is not possible to
predict the parallel efficiency achieved during the search. Previous work relied on ad hoc, control parameter tuning in order to achieve good performance \cite{jinnai2016automated}.
In this section, we first show that a work distribution method can be modeled as a partition of the search space graph,  and that communication overhead and load balance can be understood as the number of cut edges and balance of the partition, respectively.
Using this model, we introduce a metric, {\it estimated efficiency}, and we experimentally show that the metric has a strong correlation to the actual efficiency. 
This leads to the GRAZHDA* feature generation method described in Section \ref{sec:graph-partitioning-based}.

\subsection{Work Distribution as Graph Partitioning}
\label{sec:graph}
Work distribution methods for hash-based parallel search distribute nodes by assigning a process to each node in the state space.
Our goal is to design a work distribution method which maximizes efficiency by reducing CO, SO, and load balance (LB). 
In particular, given a problem instance, we seek a principled method of quickly, automatically generating a  work distribution method (hash function) for HDA* for that particular problem instance.
We propose an approach which is based on optimizing {\it a priori} estimates of CO, SO, and LB.
In our approach, given a problem, we search a space of hash functions, using these estimates of CO, SO, LB as the basis for a (cheap) evaluation function for this search in the space of hash functions.
To enable this, we first develop a model for estimating algorithm performance based on  the notion of a workload graph.

To guarantee the optimality of a solution, a parallel search method needs to expand a goal node and all nodes with $f < f^*$ (relevant nodes $S$). 
The workload distribution of a parallel search can be modeled as a partitioning of an undirected, unit-cost {\it workload graph} $\wg$  which is isomorphic to the {\it relevant} search space graph, i.e., nodes in $\wg$ correspond to states in the search space with $f < f^*$ and goal nodes, and edges in the workload graph correspond to edges in the search space between nodes with $f < f^*$ and goal nodes. 
The distribution of nodes among $p$ processors corresponds to a $p$-way partition of $\wg$, where nodes in partition $S_i$ are assigned to process $p_i$.

The workload graph  $\wg$ only includes nodes with $f < f^*$, for the following reason.
We are ultimately trying to develop a method for quickly estimating SO, CO, and LB for a work distribution scheme $S$ without actually running $S$. 
In principle, if we knew exactly the actual portion of the graph which is explored by HDA* with a particular partitioning scheme, then this would allow us to accurately compute search efficiency. However, that requires running HDA* until a solution is found, so this is impractical, and we need an approximation of the actual explored nodes.
The set of nodes with $f < f^*$ is a reasonable approximation to the nodes which are explored by HDA*, because these are the set of nodes which must be expanded regardless of the hash function (partitioning method).
Depending on the hash function, some nodes with $f \geq f^*$ are expanded, but it is not possible to know how many such nodes will be expanded without actually running HDA* with that hash function.
Therefore, although the workload graph underestimates the size of the actual relevant search space, it is a reasonable approximation.
While underestimating the relevant search space is not ideal, the converse (considering states which are irrelevant to the actual HDA*) is problematic.
For example, if we consider how the entire search space (i.e., including all nodes with $f\geq f^*$) 
would be mapped to processors if the search algorithm continued to execute until the search space is exhausted, then  \PHDA{} (Section \ref{sec:15-puzzle}) successfully partitions the space evenly, i.e., ``perfect load balance''. 
However, as shown in Figure \ref{fig:15puzzle-lb-so}, \PHDA{} has the worst load balance among the schemes compared in the experiment. This is because the distribution of \PHDA{} is highly biased in the search space so that the relevant state space ($f \leq f^*$), which is a small fraction of the state space, is distributed unevenly. 
Considering only the nodes with $f < f^*$ allows us to capture this bias.
This example also illustrates how using a perfect hash function which balances the partitions for the entire search space does not  does not achieve good performance unless the partitions are also balanced with respect to portion of the the search space which is actually explored by the search algorithm. 

{\it Given a partitioning of $\wg$,  LB and CO can be estimated directly from the structure of the graph, without having to run HDA* and measure LB and CO experimentally, i.e., it is possible to predict and analyze the efficiency of a workload distribution method without actually executing HDA*.} 
Therefore, although it is necessary to run A* or HDA*
 {\it once} to generate a workload graph,\footnote{Hence, this is not yet a practical method for automatic hash function generation -- a further approximation of this model which does not require generating the workload graph, and yields a practical method is described in Section \ref{sec:graph-partitioning-based}.}
 we can subsequently compare the LB and CO of many partitioning methods without re-running HDA* for each partitioning method.
LB corresponds to load balance of the partitions and CO is the number of edges between partitions over the number of total edges, i.e., 
\begin{equation}
\label{eq:lb}
	CO = \frac{\sum_{i}^{p}\sum_{j > i}^{p} E(S_i,S_j)}{\sum_{i}^{p}\sum_{j \geq i}^{p} E(S_i,S_j)}, \; \; 
	LB = \frac{|S_{max}|}{mean |S_i|},
\end{equation}

where $|S_i|$ is the number of nodes in partition $S_i$, $E(S_i,S_j)$ is the number of edges between $S_i$ and $S_j$, $|S_{max}|$ is the maximum of $|S_i|$ over all processes, and $mean |S| = \frac{|S|}{p}$.



Next, consider the relationship between SO and LB.
It has been shown experimentally that an inefficient LB leads to high SO, but to our knowledge, there has been no previous analysis on {\it how} LB leads to SO in parallel best-first search.
Assume that the number of duplicate nodes is negligible\footnote{The number of duplicate nodes is closely related to LB and CO.
A duplicate occurs when a node is reached by a non-optimal path.
If the order of node expansion is exactly the same as A*, then the number of duplicates is 0.
If load balance is good and the order of node expansion is similar to A*, then we can expect the number of node duplicates to be small.
However, even if load balance is near optimal, communication latency may delay finding the optimal path and a suboptimal path may be discovered first. For example, Hyperplane Work Distribution reduces the CO of diagonal paths, resulting in a reduced number of duplicate nodes (Kobayashi et al. 2011). 
Therefore, reducing CO is also important to reduce duplicate nodes.
Figure \ref{fig:msa-eff} shows that \HWD{} and \AZHDA{} reduced the amount of duplicated nodes in MSA compared to \ZHDA{}.
Thus, by focusing on optimizing LB and CO, we can effectively reduce duplicate nodes.

},
and every process expands nodes at the same rate. 
Since HDA* needs to expand all nodes in $S$, each process expands $|S_{max}|$ nodes before HDA* terminates. 
As a consequence, process $p_i$ expands $|S_{max}| - |S_i|$ nodes not in the relevant set of nodes $S$.
By definition, such irrelevant nodes are search overhead, and therefore, we can express the overall search overhead as:
\begin{align}
\label{eq:so-lb}
	SO &= \sum_{i}^{p}(|S_{max}| - |S_i|) \nonumber \\
	&= p(LB-1).
\end{align}

\subsection{Parallel Efficiency and Graph Partitioning}
\label{sec:analysis}

In this section we develop a metric to estimate the walltime efficiency as a function of CO and SO. 
First, we define {\it time efficiency} $\teff := \frac{\mbox{speedup}}{\mbox{\# cores}}$, where $\mbox{speedup} = T_N/T_1$, $T_n$ is the runtime on $N$ cores and $T_1$ the runtime on 1 core. 
Our ultimate goal is to maximize $\teff$.

\noindent {\bf Communication Efficiency:}
Assume that the communication cost between every pair of  processors is identical. 
If $t_{com}$ is the time spent sending nodes from one core to another\footnote{In a multicore environment, the cost of ``sending'' a node from thread $p_1$ to $p_2$ is the time required to obtain access to the incoming queue for $p_2$ (via a successful {\tt try\_lock} instruction).}, and $t_{proc}$ is the time spent processing nodes (including node generation and evaluation).
Hence communication efficiency, the degradation of efficiency by communication cost, is $\ceff = \frac{1}{1+c CO}$, 
where $c=\frac{t_{com}}{t_{proc}}$.

\noindent {\bf Search Efficiency:}
Assuming all cores expand nodes at the same rate 
and that there are no idle cores, HDA* with $p$ processes expands $np$ nodes in the same wall-clock time A* requires to expand $n$ nodes.
Therefore, search efficiency, the degradation of efficiency by search overhead, is $\seff = \frac{1}{1+SO}$.



Using CO and LB (and SO from Equation \ref{eq:so-lb}), we can estimate the time efficiency $\teff$.
$\teff$ is proportional to the product of communication and search efficiency: $\teff \propto \ceff \cdot \seff$. There are overheads other than CO and SO such as hardware overhead (i.e. memory bus contention) that affect performance  \cite{burnslrz10,kishimotofb13}, but we assume that CO and SO are the dominant factors in determining efficiency.

We define {\it estimated efficiency} $\sceff$ as $\sceff := \ceff \cdot \seff$, and we use this metric to estimate the actual performance (efficiency) of a work distribution method. 
\begin{align}
\label{eq:sceff}
\sceff &= \ceff \cdot \seff \nonumber
	=  \frac{1}{(1+cCO)(1+SO)} \\
	&= \frac{1}{(1+cCO)(1+p(LB-1))}
\end{align}

\subsubsection{Experiment: $\sceff$ Model vs. Actual Efficiency}
\label{sec:sceff-validation}


To validate the usefulness of $\sceff$, we evaluated the correlation of $\sceff$ and actual efficiency on the following HDA* variants discussed in Section \ref{sec:previous-methods} on domain-independent planning.
{\small 
\begin{itemize}[leftmargin=* ] 
\itemsep0em 

\item
FAZHDA*: \FAZHDA{}, AZHDA* using fluency-based filtering (FluencyAFG). 

\item
GAZHDA*: \GAZHDA{}, AZHDA* using greedy abstract feature generation (GreedyAFG). 

\item
OZHDA*: \OZHDA{}, Operator-based Zobrist hashing (Sec. \ref{sec:zhda}). 

\item
DAHDA*: \DAHDA{}, AHDA* \cite{burnslrz10} with dynamic abstraction size threshold (Appendix \apDAHDA{}). 

\item
ZHDA*: \ZHDA{}, HDA* using Zobrist hashing \cite{kishimotofb13} (Sec. \ref{sec:zhda}).


\end{itemize}
}
We implemented these HDA* variants on top of the Fast Downward classical planner using the merge\&shrink heuristic \cite{helmert2014merge} (abstraction size =1000). 
We parallelized Fast Downward using MPICH3.
We selected a set of IPC benchmark instances that are difficult enough so that parallel performance differences could be observed. 
We ran experiments on a cluster of 6 machines, each with an 8-core \lucy, 
and 1000Mbps Ethernet interconnect. 
For FAZHDA*, we ignored 30\% of the variables with the highest fluency (we tested 10\%, 20\%, 30\%, 50\%, and 70\% and found that 30\% performed the best).
DAHDA* uses at most 30\% of the total number of features in the problem instance (we tested 10\%, 30\%, 50\%, and 70\% and found that 30\% performed the best).
We packed 100 states per MPI message in order to reduce the number of messages \cite{romein1999transposition}.

\begin{figure}[htb]
	\centering
	\includegraphics[width=0.4\linewidth]{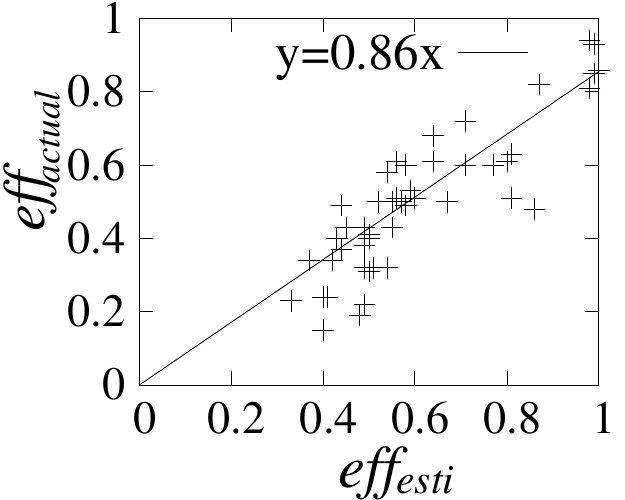}
	\caption{Comparison of $\sceff$ and the actual experimental efficiency when communication cost $c = 1.0$ and the number of processes $p=48$. The figure aggregates the data points of FAZHDA*, GAZHDA*, OZHDA*, DAHDA*, and ZHDA* shown in Figure \ref{48eff}.
$\teff = 0.86 \cdot \sceff$ with variance of residuals = 0.013 (least-squares regression).}
	\label{fig:sceff-eff}
	\captionlistentry[experiment]{sparsity-sceff-teff, Figure \ref{fig:sceff-eff}, lucy, 48p, M\&S-IJCAI, mpi, newmaterial}
\end{figure}

Table \ref{48} shows the speedups (time for 1 process / time for 48 processes).
We included the time for initializing work distribution methods (for all runs, the initializations completed in $\leq 1$ second), but
excluded the time for initializing the abstraction table for the merge\&shrink 
heuristic.
From the measured runtimes, we can compute actual efficiency $\teff$.
%
Then, we calculated the performance estimate $\sceff$ as follows.
We generated the workload graph $\wg$ for each instance (i.e., enumerated all nodes with $f \leq  f^*$ and edges between these nodes), and calculated LB, CO, SO, and $\sceff$ using Eqs \ref{eq:lb}-\ref{eq:sceff}.
Figure \ref{fig:sceff-eff}, which compares estimated efficiency $\sceff$ vs. the actual measured efficiency $\teff$, indicates a strong correlation between $\sceff$ and $\teff$.  
Using least-square regression to estimate the coefficient $a$ in  $\teff = a \cdot \sceff$, we obtained $a = 0.86$ with variance of residuals 0.013.
Note that  $a<1.0$ because there are other sources of overhead which not accounted for in $\sceff$, (e.g. memory bus contention) which affect performance \cite{burnslrz10,kishimotofb13}. 

\begin{observation} 
The $\sceff$ metric for a partitioning scheme, which  can be computed from the workload distribution graph (without running HDA* using that partitioning scheme), 
is strongly correlated with the actual measured efficiency $\teff$ of HDA*. 
\end{observation}

\section{Graph Partitioning-Based Abstract Feature Generation (GRAZHDA*)}
\label{sec:graph-partitioning-based}

A standard approach to workload balancing in  parallel scientific computing is graph partitioning, where the workload is represented as a graph, and a partitioning of the graph according to some objective (usually the cut-edge ratio metric) represents the allocation of the workload among the processors \cite{hendrickson2000graph,bulucc2013recent}.

In Section \ref{sec:implicit-graph-partitioning}, we showed that work distributions for parallel search on an implicit graph can be modeled as partitions of a workload graph which is isomorphic to the search space, and that this workload graph can be used to estimate the CO and LB of a work distribution.
If we were given a workload graph, then by defining a graph cut objective such that partitioning the nodes in the search space (with $f \leq f^*$) corresponds to maximizing the efficiency, so we would have a method of generating an optimal workload distribution.
%
Unfortunately, this is impractical as the workload graph is an explicit representation of the relevant state space graph, i.e., this a solution to the search problem itself!

However, a practical alternative is to apply graph partitioning to a graph which serves an approximate, proxy for the actual state space graph.
We propose {\it GRaph partitioning-based Abstract Zobrist HDA*} ({\bf GRAZHDA*}), which approximates the optimal graph partitioning-based strategy by partitioning {\it domain transition graphs} (DTG).
Given a  classical planning problem represented in SAS+, 
the domain transition graph (DTG) of a SAS+ variable $X$, $\mathcal{D}_{X}(E,V)$, is a directed graph where vertices $V$ corresponds to the possible values of a variable $X$, edges $E$ represent transitions among the values of $X$, and 
$(v, v') \in E$ iff there is an operator (action) $o$ with $v \in del(o)$ and $v' \in add(o)$ \cite{jonsson1998state}.

\begin{adjustbox}{width=\textwidth,keepaspectratio}
\lstset{language=pddl,basicstyle=\ttfamily}
\begin{lstlisting}[caption=Sliding-tile puzzle PDDL]
(define (domain strips-sliding-tile)
  (:requirements :strips)
  (:predicates
   (tile ?x) (position ?x)
   (at ?t ?x ?y) (blank ?x ?y)
   (inc ?p ?pp) (dec ?p ?pp))
  (:action move-up
    :parameters (?omf ?px ?py ?by)
    :precondition (and
		   (tile ?omf) (position ?px) (position ?py) (position ?by)
		   (dec ?by ?py) (blank ?px ?by) (at ?omf ?px ?py))
    :effect (and (not (blank ?px ?by)) (not (at ?omf ?px ?py))
		 (blank ?px ?py) (at ?omf ?px ?by)))
  (:action move-left
	.
	.
\end{lstlisting}
\end{adjustbox}

\begin{figure}[htb]
	\centering
	\subfloat{{\includegraphics[width=0.9\linewidth]{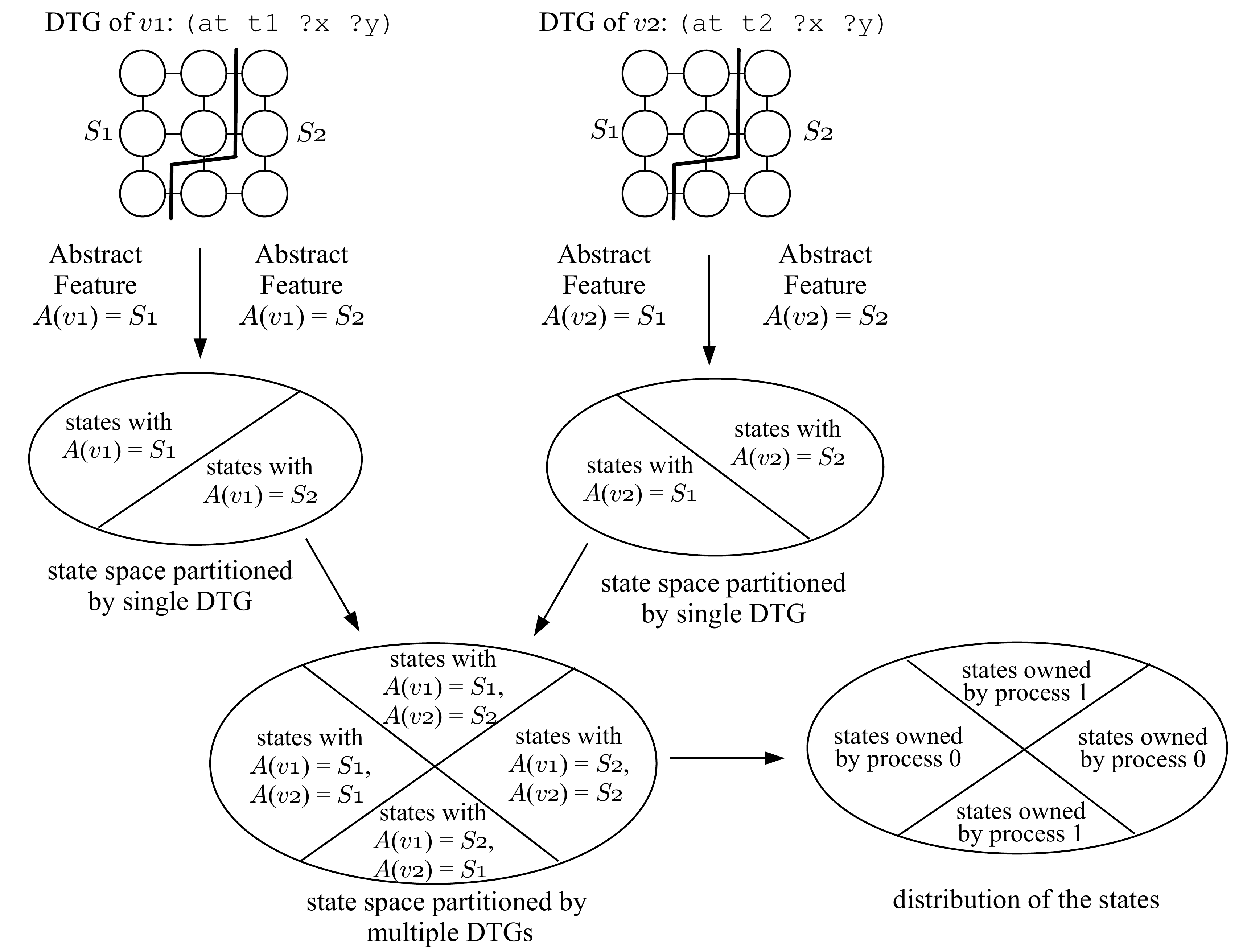}}}
	\caption{
GRAZHDA* applied to 8 puzzle domain. The SAS+ variable $v_1$ and $v_2$ correspond to the position of tile 1 and 2. The domain transition graphs (DTGs) of $v_1$ and $v_2$ are shown in the top of the figure (e.g. $v_1 = $ \{\texttt{(at t1 x1 y1), (at t1 x1 y2), (at t1 x1 y3),...}\}).
GRAZHDA* partitions each DTG with given objective function to generate abstract feature $S_1$ and $S_2$, and $A(v_1) = S_1, S_2$. 
Thus, the hash value of abstract feature $R[A(v_1)]$ corresponds to which partition $v_1$ belongs to.
As DTGs are compressed representation of the state space graph, partitioning a DTG corresponds to partitioning a state space graph.
By xor'ing $R[A(v_1)], R[A(v_2)],...$, the hash value $AZ(s)$ represents for each variable $v_i$ which partition it belongs to.}
	\label{fig:sazhda-diagram}
\end{figure}


The DTGs for a problem provide a highly compressed representation which reflects the structure of the search space, and is easily extracted automatically from the formal domain description (e.g., PDDL/SAS+). 
We expect DTGs to be good proxies for the search space because DTGs tend to be orthogonal to each other -- otherwise the propositions of the DTG is redundant (this is not always true as PDDL may contain dual representations, e.g. \pddl{sokoban}). 


GRAZHDA* partitions each DTG into two abstract features according to an objective function. 
That is, each DTG is partitioned into two subsets $S_1$ and $S_2$. 
Projection $A(x)$ is defined on the value of the DTG, and returns 1 or 0 depending on whether $S_1$ or $S_2$  it is included in. 
Abstract Zobrist hashing is then applied using these abstract features (random table $R$ in Equation \ref{eq:sz} is defined on $S_1$ and $S_2$). 
In GRAZHDA*, AZH uses each partition of the DTG as an abstract feature, assigning a hash value to each abstract feature (Figure \ref{fig:sazhda-diagram}).
Since the AZH value of a state is the XOR of the hash values of the abstract features (Equation \ref{eq:sz}), two nodes in the state space are in different partitions if and only if they are partitioned in {\it any} of the DTGs. 
Therefore, GRAZHDA* generates $2^n$ partitions from $n$ DTGs, which are then projected to the $p$ processors (by taking the hash value modulo $p$, {\it processor(s)} = {\it hashvalue(s)}  $\bmod \; p$).\footnote{In HDA* the owner of a state is computed as {\it processor(s)} = {\it hashvalue(s)} $\bmod \; p$, so it is possible that states with different hash values are assigned to the same thread. Also, while extremely unlikely, it is theoretically possible that $s$ and $s'$ may have the same hash value even if they have different abstract features due to the randomized nature of Zobrist hashing (in all our HDA* variants, we detect such collisions by always comparing the values stored in the hash table whenever hash keys point to a nonempty hash table entry).} 
We denote GRAZHDA* as \GRAZHDA{}, where DTG stands for DTG-partitioning.

\subsection{Previous Methods and Their Relationship to GRAZHDA*} 
\label{sec:previous-methods}

\begin{figure}[htb]
	\centering
	\subfloat{{\includegraphics[width=0.99\linewidth]{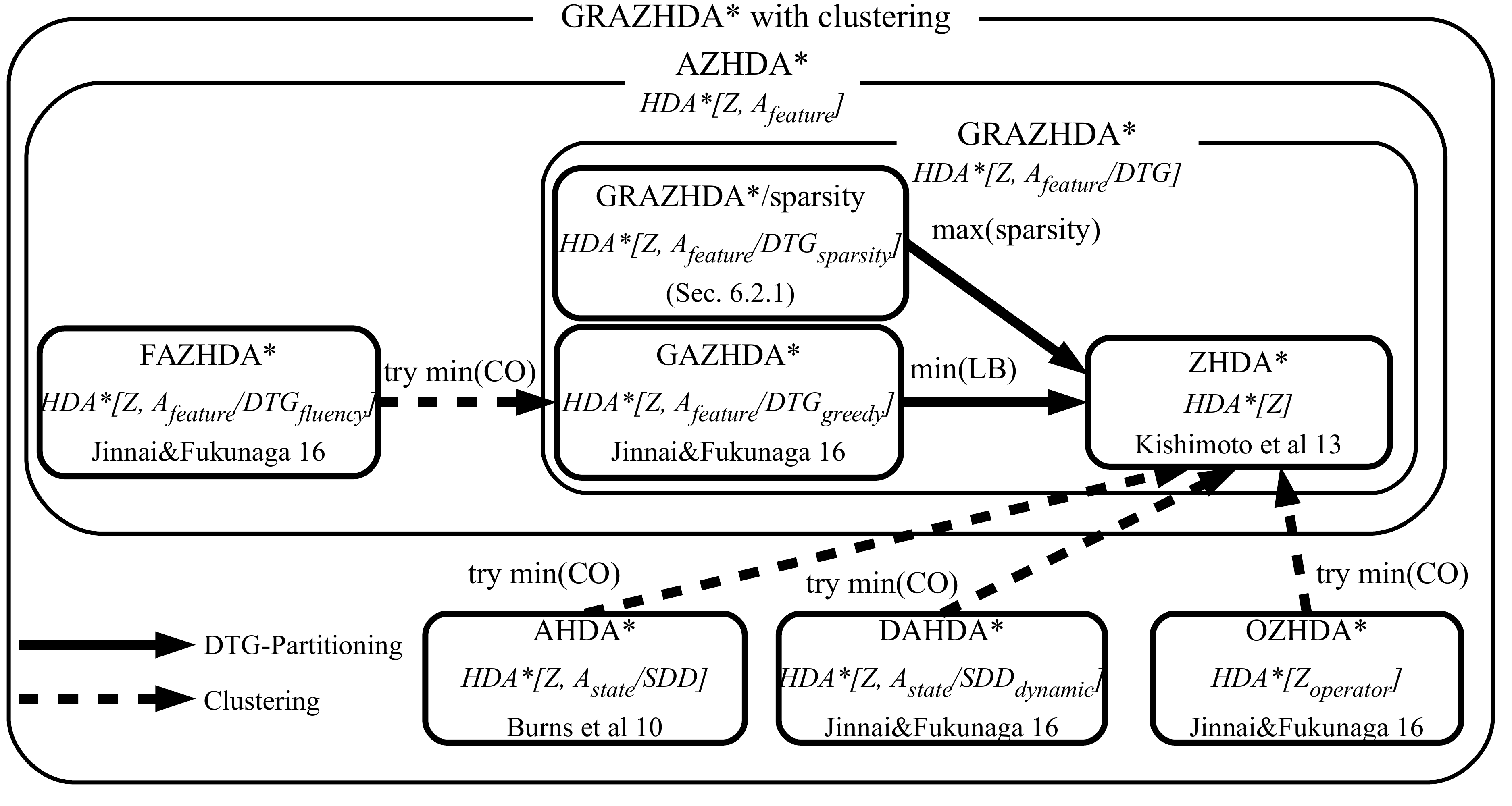}}}
	\caption{Work distribution methods described as an instances of GRAZHDA* with clustering. Previous methods can be seen as GRAZHDA* + clustering with suboptimal objective function. The arrows represent the relationship of methods. For example, FAZHDA* applies fluency-based filtering to ignore some variables, and then applies GreedyAFG to partition DTGs. This can be described as applying clustering, partitioning, and then Zobrist hashing. As such, all previous methods discussed in this paper can be characterized as instances of GRAZHDA* (with clustering).} 
	\label{fig:work-distribution-methods}
\end{figure}


In this section we show that previously proposed methods for the HDA* framework can be interpreted as instances of GRAZHDA*.
First, we define a DTG-partitioning as follows: given $s = (v_0, v_1,...,v_n)$, a DTG-partitioning maps a state $s$ to an abstract state $s' = (A_0[v_0], A_1[v_1],...,A_n[v_n])$, 
where $A_i[v_i]$ is defined by a graph partitioning on each DTG while optimizing a given objective function. DTG-partitioning corresponds to 
$\mathit{A_{feature}/DTG}$
for an abstraction strategy.
Then, in order to model non-DTG based methods, we refer to all other 
 methods which map a state space to an abstract state space with or without objectives a \emph{clustering}. For example, by ignoring subset of the variables, we get an abstract state $s' = (v_0,...,v_m)$ where $m<n$. Clustering corresponds to any abstraction strategy other than DTG-partitioning.
Using this terminology, the relationship between GRAZHDA* and previous methods is summarized in Figure \ref{fig:work-distribution-methods}.

First,  \ZHDA{}, the original Zobrist-hashing based HDA* \cite{kishimotofb09,kishimotofb13}, corresponds to an extreme case where every node in DTG is assigned to a different partition (for all $A_i$, $A_i[v_i] \neq A_i[v_i']$ if $v_i \neq v_i'$).

GAZHDA* (GreedyAFG) \cite{jinnai2016structured}, described in  Section \ref{sec:greedyafg} is in fact applying DTG-partitioning whose objective function is to minimize LB as the primary objective, with a secondary objective of (greedily) minimizing CO, as it tries to assign the most connected node but does not optimize. Thus, GAZHDA* an instance of GRAZHDA*. 

AHDA* \cite{burnslrz10} (Section \ref{sec:ahda}),  
FAZHDA* \cite{jinnai2016automated} (Section \ref{sec:fluencyafg}), OZHDA* \cite{jinnai2016automated} (Section \ref{sec:zhda}), and DAHDA* \cite{jinnai2016automated} (Section\ref{sec:ahda}), are instances of GRAZHDA* with {\it clustering},  which map the state space graph to an abstract state space graph, and then apply DTG-partitioning to the abstract state space graph so that the nodes mapped to the same abstract state 
are guaranteed to be assigned to the same partition, so that no communication overhead is incurred when generating a node that is in the same abstract state as its parent. 

AHDA* generates an abstract state space by ignoring some of the features (DTGs) in the state representation and then it applies 
hashing to the abstract state space. 
Ignoring part of the state representation can be interpreted as a {\it clustering} of nodes so that all of the nodes in a cluster are allocated to the same processor.
The problem with AHDA* is the criteria used to determine which features to ignore (conversely, which features to take into account). It minimizes the highest degree of the abstract nodes, as the abstraction method used by AHDA* was originally proposed for duplicate detection of external search \cite{Zhou2006}. However, this doe not correspond to a natural objective function which  optimizes parallel work distribution objective such as edge cut or load balancing.
Therefore, although the projection of AHDA* results in significantly reduced CO, it does not explicitly try to optimize it; CO is reduced as a fortunate  side-effect of generating efficient 
abstract state space for external search.
DAHDA* \cite{jinnai2016automated} improves upon AHDA* by dynamically tuning the number of DTGs which are ignored
(see Appendix \apDAHDA{}), but the state projection mechanism is the same as AHDA*.

FAZHDA* is a variant of GAZHDA*, which,  instead of using all the variables as GAZHDA* does, FAZHDA* ignores some of the variables in the state based on their {\it fluency}, which is defined as the number of ground actions which change the value of the variable divided by the total number of ground actions in the problem. As we pointed out above for AHDA*, ignoring variables can be interpreted as a clustering.
Although fluency-based filtering is intended to reduce CO, ignoring high fluency variables is only a heuristic which succeeds in reducing CO only some of the time.
This is because fluency is defined as the frequency of the change of the feature (value), but the change of {\it abstract} feature is what really incurs CO --
even if the fluency of a variable is 1.0, the value may change within an abstract feature,
in which case eliminating the DTG does not improve CO. 
Wherease fluency-based filtering only takes into account of the fluency of the variable, the GRAZHDA* framework considers each transition in the DTG to determine how to treat the variable. 
 

OZHDA* clusters nodes connected with selected operators and applies Zobrist hashing, so that the selected operator does not cost communication.
The clustering of OZHDA* is bottom-up, in the sense that state space nodes connected by selected operators are directly clustered, instead of using SAS+ variables or DTGs.
The problem with OZHDA* is that the clustering is ad hoc and unbalanced --  some of the nodes are clustered but the others are not, and the choice of which nodes to cluster or not is not explicitly optimized.
The clustered nodes are then partitioned by assigning each node to a separate partition, as with ZHDA* (see above), but this is dangerous, since OZHDA* ends up treating clustered nodes and original nodes equally, without considering that the clustered nodes should have larger edge cut costs than original single nodes.
Thus, although the clustering done by OZHDA*  is intended to reduce CO, it comes at the price of load balance -- the edge costs for the (implicit) workload graph are not aggregated when the clusters are formed, so load balance is being sacrificed without an explicit objective function controlling the tradeoff.

Thus, we have shown that  all previously proposed methods for work distribution in the HDA* framework  can be 
viewed as instances of GRAZHDA* using {\em ad hoc} criteria for clustering and optimization. 

\subsection{Effective Objective Functions for GRAZHDA*}
\label{sec:objective-function}


In the previous section, we showed that previous variants of HDA* can
be seen as instances of GRAZHDA* which partitioned the
workload graph based on \emph{ad hoc} criteria.
However, since the GRAZHDA* framework formulates workload distribution as a graph partitioning problem, 
a natural idea is to design an objective function for the partitioning which directly leads to a desired tradeoff between search and communication overheads, resulting in good overall efficiency. Fortunately, a metric which can be used as the basis for such an objective is available: $\sceff$.

In Section \ref{sec:sceff-validation}, we showed that $\sceff$, based on the workload  is an effective predictor for the actual efficiency of a work distribution strategy.
In this section, we propose approximations to $\sceff$ which can be used as objective functions for the DTG partitioning in GRAZHDA*.

In principle, in order to maximize the performance of GRAZHDA*, it is desirable to have  a function which approximates $\sceff$ as closely as possible.
However, since GRAZHDA* partitions the domain transition graph as opposed to the actual workload graph (which is isomorphic to the search space graph), and the DTG is only an approximation to the actual workload graph, a perfect approximation of $\sceff$ is not feasible.
Fortunately, in practice, it turns out that using a straightforward approximation of $\sceff$ as an objective function for GRAZHDA* result in good performance when compared to previous work distribution methods.

\subsubsection{Sparsest Cut Objective Function (GRAZHDA*/sparsity)}
\label{sec:sparsity}
One straightforward objective function which is clearly related to $\sceff$ is a 
{\it sparsest cut} objective, which maximizes {\it sparsity}, defined as 

\begin{equation}
\label{eq:sparsity}
	sparsity := \frac{\prod_{i}^{p}|S_i|}{\sum_{i}^{p}\sum_{j > i}^{p} E(S_i,S_j)},
\end{equation}

where $p$ is the number of partitions (= number of processors), $|S_i|$ is the number of nodes in partition $S_i$ divided by the total number of nodes,  
and $E(S_i, S_j)$ is the sum of edge weights between partition $S_i$ and $S_j$. 
Consider the relationship between the sparsity of a state space graph for a search problem and the $\sceff$ metric defined in the previous section.
By equations \ref{eq:sceff} and \ref{eq:lb}, 
sparsity simultaneously considers both LB and CO, as the numerator $\prod_{i}^{p}|S_i|$ corresponds to LB and the denominator $\sum_{i}^{p}\sum_{j > i}^{p} E(S_i,S_j)$ corresponds to CO.

Sparsity is used as a metric for parallel workloads in computer networks \cite{leighton1999multicommodity,jyothi2014measuring}, but to our knowledge this is the first proposal  to use sparsity in the context of parallel search of an implicit graph. 

\begin{figure}[htb]
	\centering
	\includegraphics[width=0.50\columnwidth]{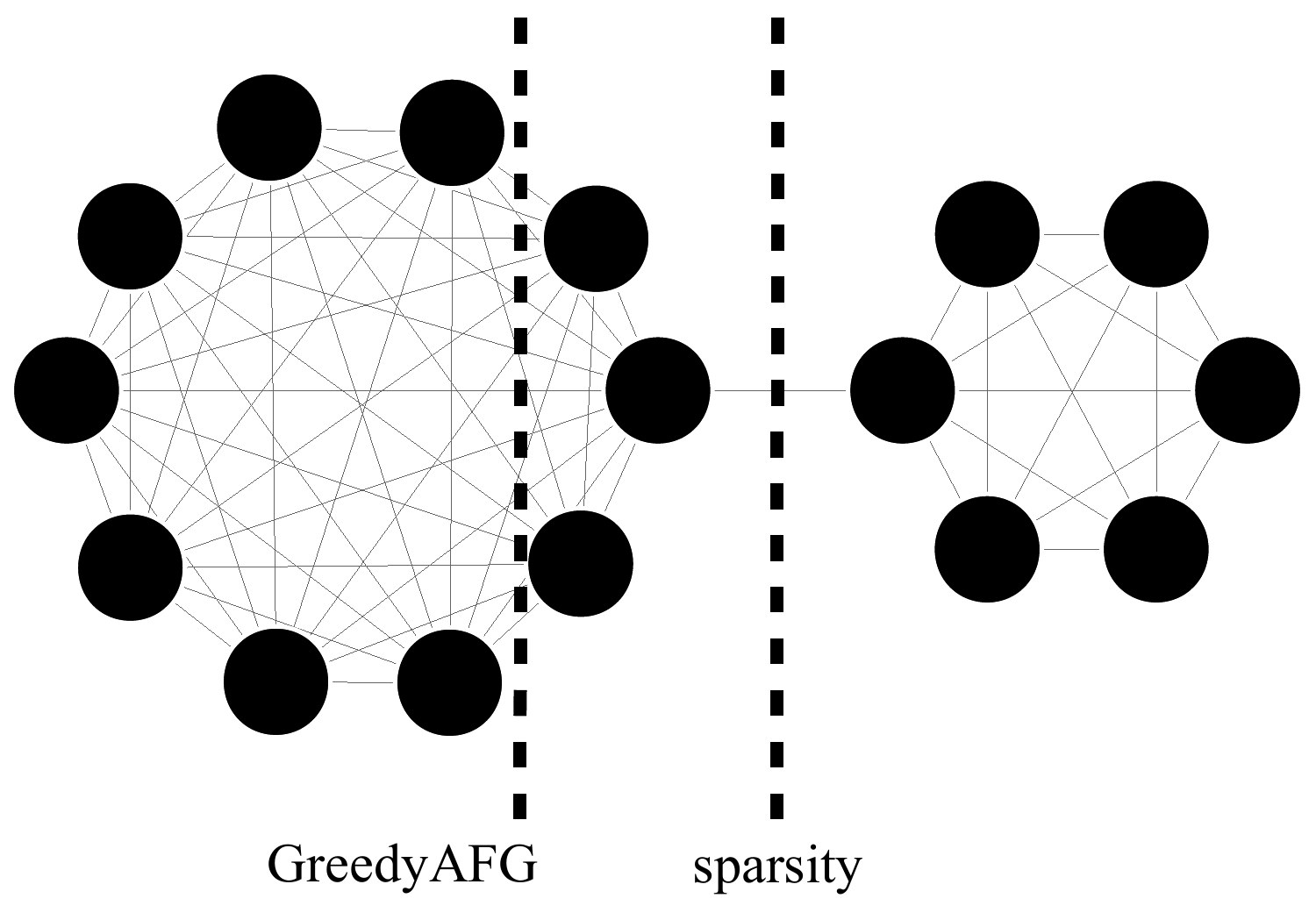}
	\caption{GRAZHDA*/sparsity and  Greedy abstract feature generation (\simple{}) applied to DTG on logistics domain of 2 cities with 10/6 locations. 
Each node in the domain transition graph above corresponds to a location of the package \pddl{(at obj12 ?loc)}.
GreedyAFG potentially cuts many edges because it requires the best load balance possible for the cut (bisection), while GRAZHDA*/sparsity takes into account of the number of edge cut as an objective function.}
	\label{fig:sazhda-vs-gazhda}
\end{figure}

Figure \ref{fig:sazhda-vs-gazhda} shows the sparsest cut of a DTG (for the variable representing package location) in the standard {\tt {\small logistics}} domain.
Each edge in a DTG corresponds to a transition of its value.
Edge costs $w_e$ represent the ratio of operators which corresponds to its transition over the total number of operators in the DTG.
For example in logistics, each edge corresponds to 2 operators, one in each direction (\pddl{(drive-truck ?truck pos0 pos1)} and \pddl{(drive-truck ?truck pos1 pos0)}, or \pddl{(fly-airplane ?plane pos0 pos1)} and \pddl{(fly-airplane ?plane pos1 pos0)} ). The total number of operator in the graph is 120, thus $w_e$ for each edge is $2/120 = 1/60$. We use this to calculate sparsity (Equation \ref{eq:sparsity}).
Maximizing sparsity results in cutting only 1 edge (Figure \ref{fig:sazhda-vs-gazhda}): it cuts the graph with $|S_1| \cdot |S_2| = 10/16 \cdot 6/16$, and edge cuts $E(S_1, S_2) = 1 \cdot w_e$, thus $sparsity = \frac{|S_1| \cdot |S_2|}{E(S_1, S_2)} = 26.72$, whereas the partition by GreedyAFG results in cutting 21 edges ($sparsity = 0.71$).
The problem with GreedyAFG is that it imposes a hard constraint requiring the partition to be perfectly balanced. While this optimizes load balance, locality (i.e., the number of cut edges) is sacrificed.
GRAZHDA*/sparsity takes into account both load balance and CO without the hard constraint of bisection, resulting in a partitioning which preserves more locality.

\subsubsection{Experiment: Validating the Relationship between Sparsity and $\sceff$} 

\begin{figure}[htb]
	\centering
	\subfloat[Comparison of $\sceff$ for various work distribution methods]{\label{fig:sceffs} \includegraphics[width=0.80\linewidth]{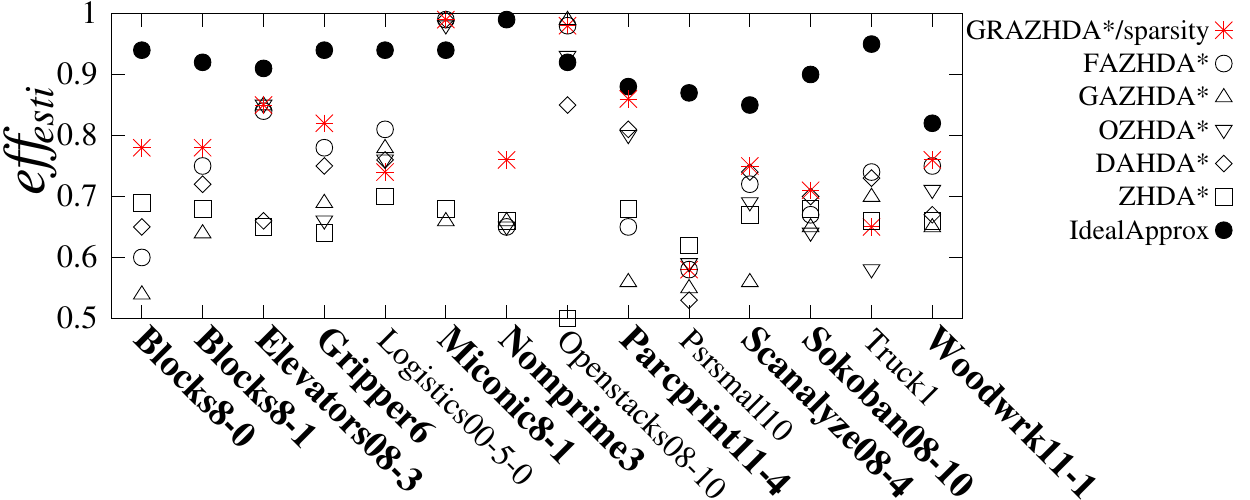}}
	
	\centering
	\subfloat[{\it sparsity} vs. $\sceff$]{\label{fig:spr-eff} \includegraphics[width=0.4\linewidth]{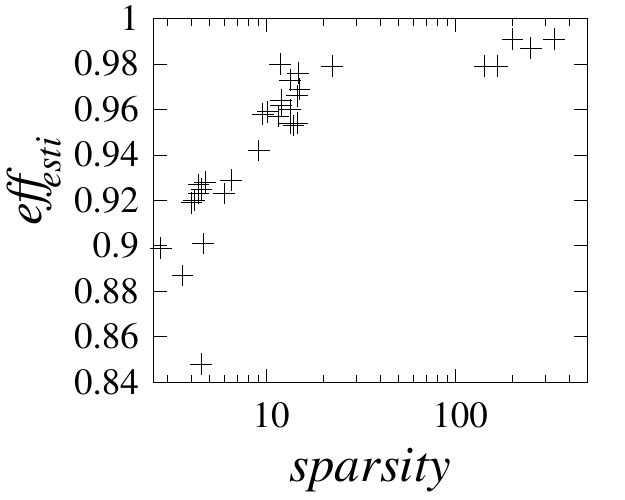}}
	\caption{Figure \ref{fig:sceffs} compares $\sceff$ when communication cost $c=1.0$, the number of processes $p=48$. {\bf Bold} indicates that GRAZHDA*/sparsity has the best $\sceff$ (except for IdealApprox). 
Figure \ref{fig:spr-eff} compares sparsity vs. $\sceff$. For each instance, we generated 3 different partitions using METIS with load balancing constraints which force METIS to balance randomly selected nodes, to see how degraded sparsity affects $\sceff$. There was no partition with $\sceff < 0.84$.}
	\captionlistentry[experiment]{sparsity-sceff-teff, Figure \ref{fig:sceffs} and \ref{fig:spr-eff}, lucy, 48p, M\&S-IJCAI, mpi, newmaterial}
\end{figure}

To validate the correlation between sparsity and estimated efficiency $\sceff$,  we used the METIS  (approximate) graph partitioning package \cite{karypis1998fast} 
to partition modified versions of the search spaces of the instances used in Fig. \ref{fig:sceffs}.
We partitioned each instance 3 times, where each run had a different set of random, artificial constraints added to the instance (we chose 50\% of the nodes randomly and forced METIS to distribute them equally among the partitions -- these constraints degrade the achievable sparsity).
Figure \ref{fig:spr-eff} compares sparsity vs. $\sceff$ on partitions generated by METIS with random constraints.
There is a clear correlation between sparsity and $\sceff$. 
Thus, partitioning a graph to maximize $sparsity$ should  maximize the $\sceff$ objective, which should in turn maximize actual walltime efficiency.

\subsubsection{Partitioning the DTGs}

Given an objective function such as sparsity, 
GRAZHDA* partitions each DTG into two abstract features, as described above in Section \ref{sec:graph-partitioning-based}.
Since each domain transition graph  typically only has  fewer than 10 nodes, we compute the optimal partition for both objective functions with a straightforward depth-first branch-and-bound procedure. 
It is possible that branch-and-bound becomes impractical 
in case a domain 
has very large DTGs, or we may develop a more complicated objective function for partitioning the DTGs.
In such cases, we can use heuristic partitioning methods such as the FM algorithm \cite{fiduccia1982linear}. However, to date, branch-and-bound has been sufficient -- in all of the standard IPC benchmark domains we evaluated, the abstract feature generation procedure (which includes partitioning all of the DTGs) take less than 4 seconds on every instance we tested (most instances take $<1$ second). 



\subsection{Evaluation of Automated, Domain-Independent Work Distribution Methods}
\label{sec:experiments-planning}


\begin{table}[hbtp]
	\caption{Comparison of $\teff$ and $\sceff$ on a commodity cluster with 6 nodes, 48 processes. 
$\sceff$ ($\teff$) with {\bf bold} font indicates the method has the best $\sceff$ ($\teff$).
A {\bf bold} instance name indicates that the best $\sceff$ method has the best $\teff$.
}
	\label{48eff}
	\centering
        \resizebox{0.87\textwidth}{!}{
	\begin{tabular}{l|rr|cc|cc} \hline
		Instance & \multicolumn{2}{c|}{A*}  & \multicolumn{2}{c|}{GRAZHDA*/}  & \multicolumn{2}{c}{FAZHDA*} \\
		 & \multicolumn{2}{c|}{}  & \multicolumn{2}{c|}{sparsity}  & \multicolumn{2}{c}{}  \\
		 & \multicolumn{2}{c|}{}  & \multicolumn{2}{c|}{\GRAZHDAST{}}  & \multicolumn{2}{c}{\FAZHDAT{}} \\
		          & time & expd & $\teff$ & $\sceff$  & $\teff$ & $\sceff$  \\ \hline
{\bf Blocks10-0        } &   129.29  &   11065451  & {\bf 0.57 } & {\bf 0.57 } & 0.54  &  0.43  \\
{\bf Blocks11-1        } &   813.86  &   52736900  & {\bf 0.71 } & {\bf 0.53 } &  0.71  &  0.50  \\
Elevators08-5      &   165.22  &    7620122  &  0.34  &  0.51  &   0.26  &  0.49  \\
{\bf Elevators08-6     } &   453.21  &   18632725  &  0.45  &  0.50  & 0.38  &  0.36  \\
{\bf Gripper8          } &   517.41  &   50068801  &  0.56  &  0.60  & {\bf 0.57 } & {\bf 0.63 } \\
{\bf Logistics00-10-1  } &   559.45  &   38720710  & {\bf 0.94 } & {\bf 0.70 } &  0.91  &  0.61  \\
Miconic11-0        &   232.07  &   12704945  &  0.87  &  0.95  &  0.88  &  0.91  \\
{\bf Miconic11-2}  &   262.01  &   14188388  &  {\bf 0.94 } & {\bf 0.97 } & 0.93  &  0.92  \\
{\bf NoMprime5 }   &   309.14  &    4160871  & {\bf 0.50 } & {\bf 0.58 } & 0.48  &  0.53  \\
NoMystery10        &   179.52  &    1372207  & {\bf 0.72 } &  0.61  &  0.48  & {\bf 0.75 } \\
Openstacks08-19    &   282.45  &   15116713  &  0.51  &  0.59  & 0.42  &  0.58  \\
Openstacks08-21    &   554.63  &   19901601  &  0.53  &  0.65  & 0.52  &  0.62  \\
{\bf Parcprinter11-11} &   307.19  &    6587422  & {\bf 0.42 } &  {\bf 0.54}  &   0.27  &  0.49  \\
Parking11-5    &   237.05  &    2940453  & {\bf 0.62 } &  0.55  &   0.62  &  0.54  \\
Pegsol11-18        &   801.37  &  106473019  &  0.44  & {\bf 0.72 } &   0.44  &  0.71  \\
{\bf PipesNoTk10  } &   157.31  &    2991859  & {\bf 0.33 } & {\bf 0.52 } & 0.33  &  0.49  \\
PipesTk12   &   321.55  &   15990349  &  0.70 & {\bf 0.66 } & {\bf 0.83}  &  0.65  \\
PipesTk17   &   356.14  &   18046744  &  0.92  & {\bf 0.65 } & {\bf 0.94 } &  0.63  \\
Rovers6            &  1042.69  &   36787877  & {\bf  0.86 } & {\bf 0.79 } & 0.84  &  0.72  \\
{\bf Scanalyzer08-6   } &   195.49  &   10202667  & {\bf 0.69 } & {\bf 0.92 } & 0.63  &  0.86  \\
{\bf Scanalyzer11-6    } &   152.92  &    6404098  & {\bf 0.91 } & {\bf 0.78 } & 0.57  &  0.63  \\ \hline
{\bf Average           } &   382.38  &   21557805  & {\bf 0.64 } & {\bf 0.62 } & 0.60  &  0.61  \\
	\end{tabular}
	}
        \resizebox{0.92\textwidth}{!}{
	\begin{tabular}{l|cc|cc|cc|cc} \hline
		Instance & \multicolumn{2}{c|}{GAZHDA*}  & \multicolumn{2}{c|}{OZHDA*} & \multicolumn{2}{c|}{DAHDA*}  & \multicolumn{2}{c}{ZHDA*}  \\
		 & \multicolumn{2}{c|}{\GAZHDAT{}}  & \multicolumn{2}{c|}{\OZHDAT{}} & \multicolumn{2}{c|}{\DAHDAT{}}  & \multicolumn{2}{c}{\ZHDAT{}}  \\
		          & $\teff$ & $\sceff$  & $\teff$ & $\sceff$  & $\teff$ & $\sceff$  & $\teff$ & $\sceff$  \\ \hline
{\bf Blocks10-0        } &  0.45  &  0.44  &  0.32  &  0.37  &  0.52  &  0.47  &  0.31  &  0.48 \\
{\bf Blocks11-1        } &  0.61  &  0.48  &  0.61  &  0.47  &  0.52  &  0.43  &  0.58  &  0.48 \\
Elevators08-5      &  {\bf 0.61}  &  0.58  &  0.46  & {\bf 0.64 } &  0.57  &  0.51  &  0.57  &  0.47 \\
{\bf Elevators08-6     } & {\bf 0.72 } & {\bf 0.76 } &  0.68  &  0.56  &  0.32  &  0.39  &  0.38  &  0.49 \\
{\bf Gripper8          } &  0.46  &  0.50  &  0.52  &  0.44  &  0.45  &  0.45  &  0.45  &  0.47 \\
{\bf Logistics00-10-1  } &  0.24  &  0.42  &  0.24  &  0.43  &  0.36  &  0.53  &  0.34  &  0.48 \\
Miconic11-0        &  0.27  &  0.53  &  0.79  & {\bf 0.96 } & {\bf 0.96 } &  0.91  &  0.15  &  0.48 \\
{\bf Miconic11-2}  &  0.18  &  0.37  &  0.77  &  0.90  &  0.70  &  0.81  &  0.31  &  0.48 \\
{\bf NoMprime5 }   &  0.39  &  0.48  &  0.35  &  0.51  &  0.38  &  0.49  &  0.35  &  0.47 \\
NoMystery10        &  0.40  &  0.66  &  0.45  &  0.50  &  0.59  &  0.60  &  0.45  &  0.49 \\
Openstacks08-19    &  0.46  &  0.58  &  0.36  &  0.55  &  0.51  & {\bf 0.66 } & {\bf 0.54 } &  0.47 \\
Openstacks08-21    &  0.53  &  0.65  & {\bf 0.82 } &  0.49  &  0.56  & {\bf 0.68 } &  0.81  &  0.51 \\
{\bf Parcprinter11-11} &  0.35  &  0.40  &  0.33  &  0.34  &  0.15  &  0.15  &  0.40  &  0.48 \\
Parking11-5    &  0.59  &  0.49  &  0.56  &  0.46  &  0.60  & {\bf 0.59 } &  0.56  &  0.47 \\
Pegsol11-18        &  0.34  &  0.53  & {\bf 0.55 } &  0.71  &  0.46  &  0.70  &  0.35  &  0.47 \\
{\bf PipesNoTk10  } &  0.32  &  0.50  &  0.32  &  0.48  &  0.32  &  0.48  &  0.07  &  0.48 \\
PipesTk12   &  0.41  &  0.48  &  0.45  &  0.49  &  0.52  &  0.57  &  0.41  &  0.48 \\
PipesTk17   &  0.56  &  0.50  &  0.60  &  0.52  &  0.65  &  0.60  &  0.55  &  0.49 \\
Rovers6            &  0.70  &  0.61  &  0.85  &  0.71  &  0.53  &  0.73  &  0.63  &  0.53 \\
{\bf Scanalyzer08-6   } &  0.42  &  0.54  &  0.49  &  0.58  &  0.44  &  0.51  &  0.34  &  0.48 \\
{\bf Scanalyzer11-6    } &  0.34  &  0.41  &  0.81  &  0.68  &  0.41  &  0.44  &  0.42  &  0.48 \\ \hline
{\bf Average           } &  0.45  &  0.51  &  0.54  &  0.53  &  0.50  &  0.47  &  0.43  &  0.49 \\
	\end{tabular}
	}
\end{table}

\begin{table}[htbp]
	\caption{Comparison of speedup,  communication overhead (CO), and search overhead (SO) on a commodity cluster with 6 nodes, 48 processes using merge\&shrink heuristic (average over 10 runs). The results with standard deviation are shown in Appendix \ref{tab:48-results-with-sd}.}
	\label{48}
	\centering
        \resizebox{0.82\textwidth}{!}{
	\begin{tabular}{l|rr|rrr|rrr} \hline
		Instance & \multicolumn{2}{c|}{A*}  & \multicolumn{3}{c|}{GRAZHDA*/}  & \multicolumn{3}{c}{FAZHDA*} \\
		 & \multicolumn{2}{c|}{}  & \multicolumn{3}{c|}{sparsity}  & \multicolumn{3}{c}{}\\
		 & \multicolumn{2}{c|}{}  & \multicolumn{3}{c|}{\GRAZHDAST{}}  & \multicolumn{3}{c}{\FAZHDAT{}}\\
		          & expd & time & speedup & CO & SO  & speedup & CO & SO   \\ \hline
Blocks10-0 & 129.29 & 11065451 & {\bf 27.17} & 0.28 & 0.38 & 26.02 & 0.70 & 0.35 \\
Blocks11-1 & 813.86 & 52736900 & {\bf 34.25} & 0.66 & 0.15 & 34.25 & 0.66 & 0.15 \\
Elevators08-5 & 165.22 & 7620122 & 16.43 & 0.47 & 0.33 & 12.34 & 0.32 & 0.51 \\
Elevators08-6 & 453.21 & 18632725 & 21.47 & 0.49 & 0.37 & 18.05 & 0.52 & 0.81 \\
Gripper8 & 517.41 & 50068801 & 26.67 & 0.50 & 0.15 & {\bf 27.45} & 0.43 & 0.10 \\
Logistics00-10-1 & 559.45 & 38720710 & {\bf 45.16} & 0.43 & 0.01 & 43.85 & 0.57 & 0.02 \\
Miconic11-0 & 232.07 & 12704945 & 41.97 & 0.01 & 0.07 & 42.43 & 0.01 & 0.06 \\
Miconic11-2 & 262.01 & 14188388 & {\bf 45.26} & 0.01 & 0.05 & 44.87 & 0.01 & 0.05 \\
NoMprime5 & 309.14 & 4160871 & {\bf 23.95} & 0.80 & -0.04 & 22.87 & 0.79 & -0.05 \\
NoMystery10 & 179.52 & 1372207 & {\bf 34.80} & 0.51 & 0.12 & 22.99 & 0.24 & -0.44 \\
Openstacks08-19 & 282.45 & 15116713 & 24.67 & 0.27 & 0.34 & 20.00 & 0.24 & 0.37 \\
Openstacks08-21 & 554.63 & 19901601 & 25.23 & 0.17 & 0.35 & 24.97 & 0.15 & 0.35 \\
Parcprinter11-11 & 307.19 & 6587422 & {\bf 20.26} & 0.26 & 0.55 & 13.08 & 0.26 & 0.61 \\
Parking11-5 & 237.05 & 2940453 & {\bf 29.75} & 0.40 & 0.34 & 29.67 & 0.63 & 0.11 \\
Pegsol11-18 & 801.37 & 106473019 & 21.03 & 0.39 & 0.02 & 20.97 & 0.39 & 0.00 \\
PipesNoTk10 & 157.31 & 2991859 & {\bf 15.73} & 0.98 & 0.01 & 15.64 & 0.98 & 0.01 \\
PipesTk12 & 321.55 & 15990349 & 33.78 & 0.46 & 0.05 & {\bf 39.65} & 0.46 & 0.03 \\
PipesTk17 & 356.14 & 18046744 & 43.92 & 0.54 & 0.01 & {\bf 45.03} & 0.54 & 0.01 \\
Rovers6 & 1042.69 & 36787877 & {\bf 41.17} & 0.15 & 0.14 & 40.48 & 0.15 & 0.17 \\
Scanalyzer08-6 & 195.49 & 10202667 & {\bf 32.92} & 0.12 & 0.01 & 30.31 & 0.12 & 0.01 \\
Scanalyzer11-6 & 152.92 & 6404098 & {\bf 43.83} & 0.16 & 0.13 & 27.31 & 0.18 & 0.34 \\ \hline
Average & 382.38 & 21557805 & {\bf 30.92} & 0.38 & 0.17 & 28.68 & 0.40 & 0.17 \\ \hline
Total walltime & 8029.97 & 452713922 & \multicolumn{3}{r|}{{\bf 277.91}} & \multicolumn{3}{r}{301.38}  \\ 
	\end{tabular}
	}

        \resizebox{0.95\textwidth}{!}{
	\begin{tabular}{l|rrr|rrr|rrr|rrr} \hline
		Instance & \multicolumn{3}{c|}{GAZHDA*}  & \multicolumn{3}{c|}{OZHDA*} & \multicolumn{3}{c|}{DAHDA*}  & \multicolumn{3}{c}{ZHDA*}  \\
		 & \multicolumn{3}{c|}{\GAZHDAT{}}  & \multicolumn{3}{c|}{\OZHDAT{}} & \multicolumn{3}{c|}{\DAHDAT{}}  & \multicolumn{3}{c}{\ZHDAT{}}  \\
		          & speedup & CO & SO & speedup & CO & SO & speedup & CO & SO   & speedup & CO & SO  \\ \hline
Blocks10-0 & 21.81 & 0.99 & 0.12 & 15.47 & 0.98 & 0.34 & 25.11 & 0.88 & 0.08 & 14.93 & 0.98 & 0.30 \\ 
Blocks11-1 & 29.20 & 0.99 & 0.03 & 29.20 & 0.99 & 0.03 & 24.88 & 0.91 & 0.21 & 27.98 & 0.98 & 0.07 \\ 
Elevators08-5 & {\bf 29.35} & 0.65 & -0.00 & 21.86 & 0.09 & 0.44 & 27.59 & 0.83 & -0.03 & 27.54 & 0.98 & -0.03 \\ 
Elevators08-6 & {\bf 34.52} & 0.24 & -0.09 & 32.70 & 0.41 & 0.22 & 15.28 & 0.88 & 0.31 & 18.19 & 0.96 & 0.06 \\ 
Gripper8 & 21.86 & 0.81 & 0.06 & 24.77 & 0.98 & 0.14 & 21.80 & 0.98 & 0.08 & 21.66 & 0.98 & 0.08 \\ 
Logistics00-10-1 & 11.68 & 0.85 & 0.25 & 11.68 & 0.85 & 0.25 & 17.52 & 0.84 & 0.00 & 16.09 & 0.99 & 0.00 \\ 
Miconic11-0 & 13.15 & 0.53 & 0.24 & 37.86 & 0.02 & 0.02 & {\bf 46.05} & 0.01 & 0.08 & 7.40 & 0.96 & 0.13 \\ 
Miconic11-2 & 8.53 & 0.53 & 0.74 & 36.86 & 0.02 & 0.07 & 33.81 & 0.01 & 0.18 & 14.67 & 0.96 & 0.05 \\ 
NoMprime5 & 18.55 & 0.95 & -0.06 & 16.66 & 0.94 & 0.00 & 18.46 & 0.90 & -0.05 & 16.63 & 0.98 & -0.02 \\ 
NoMystery10 & 18.98 & 0.42 & -0.07 & 21.61 & 0.74 & 0.11 & 28.41 & 0.60 & -0.07 & 21.68 & 0.99 & -0.07 \\ 
Openstacks08-19 & 22.14 & 0.38 & 0.21 & 17.11 & 0.34 & 0.32 & 24.54 & 0.24 & 0.18 & {\bf 25.99} & 0.99 & -0.05 \\ 
Openstacks08-21 & 25.67 & 0.15 & 0.31 & {\bf 39.34} & 0.92 & 0.05 & 26.72 & 0.13 & 0.28 & 39.06 & 0.92 & -0.00 \\ 
Parcprinter11-11 & 16.85 & 0.74 & 0.41 & 15.98 & 0.82 & 0.56 & 7.00 & 0.19 & 4.38 & 19.15 & 0.97 & 0.08 \\ 
Parking11-5 & 28.43 & 0.98 & 0.02 & 26.76 & 0.97 & 0.07 & 28.84 & 0.52 & 0.07 & 27.09 & 0.98 & 0.04 \\ 
Pegsol11-18 & 16.22 & 0.77 & 0.05 & {\bf 26.17} & 0.34 & -0.03 & 22.16 & 0.34 & -0.01 & 16.97 & 0.98 & 0.03 \\ 
PipesNoTk10 & 15.58 & 0.98 & 0.01 & 15.22 & 0.98 & 0.02 & 15.58 & 0.98 & 0.01 & 3.22 & 0.98 & -0.44 \\ 
PipesTk12 & 19.84 & 0.99 & 0.01 & 21.40 & 0.88 & 0.04 & 25.12 & 0.67 & 0.00 & 19.78 & 0.98 & 0.00 \\ 
PipesTk17 & 26.64 & 0.98 & 0.00 & 28.82 & 0.88 & 0.00 & 31.16 & 0.60 & 0.01 & 26.27 & 0.98 & 0.00 \\ 
Rovers6 & 33.49 & 0.56 & 0.01 & 41.00 & 0.31 & 0.03 & 25.48 & 0.05 & 0.26 & 30.01 & 0.76 & 0.00 \\ 
Scanalyzer08-6 & 20.28 & 0.77 & 0.01 & 23.70 & 0.66 & 0.01 & 21.23 & 0.94 & 0.00 & 16.54 & 0.98 & 0.01 \\ 
Scanalyzer11-6 & 16.36 & 0.65 & 0.49 & 38.82 & 0.30 & 0.09 & 19.51 & 0.50 & 0.46 & 20.36 & 0.98 & 0.05 \\ \hline 
Average & 21.39 & 0.71 & 0.13 & 25.86 & 0.64 & 0.13 & 24.11 & 0.57 & 0.31 & 20.53 & 0.96 & 0.01 \\  \hline
Total walltime & \multicolumn{3}{r|}{398.75} & \multicolumn{3}{r|}{331.18} & \multicolumn{3}{r|}{377.86} & \multicolumn{3}{r}{433.23}  \\ 
	\end{tabular}
	}
	\captionlistentry[experiment]{EvaluationCommodity, Table \ref{48}, lucy, 48p, M\&S-IJCAI, mpi, newmaterial+updatedmaterial (ICAPSnewheuristic)}

\end{table}

In addition to the methods in Section \ref{sec:sceff-validation}, we evaluated the performance of GRAZHDA*/sparsity.
We used the CGL-B (CausalGraph-Goal-Level\&Bisimulation) merge\&shrink heuristic \cite{helmert2014merge}, which is more recent and more efficient than LFPA merge\&shrink \cite{helmert2007flexible} used in the previous conference paper which evaluated GAZHDA* and FAZHDA* \cite{jinnai2016automated}. For example on the IPC  Block10-1 instance, CGL-B expands 11,065,451 nodes while LFPA 51,781,104 expands nodes. We set the abstraction size for merge\&shrink to 1000. 
The choice of heuristic affects the behavior of parallel search if different node expansion rates are obtained depending on the heuristic, because node expansion rate affects the relative cost of communication. 
As CGL-B and LFPA have roughly the same node expansion rate, we did not observe a significant difference on the effect of work distribution methods.
Therefore, we show the result using CGL-B because it runs faster on sequential A*.
We discuss the effect of node expansion rate in Section \ref{sec:lmcut}. 
We did not apply fluency-based filtering (Section \ref{sec:fluencyafg}) and used all DTGs in GRAZHDA*/sparsity because it did not improve the performance. 



Figure \ref{fig:sceffs} shows $\sceff$ for the various work distribution methods, including GRAZHDA*
(see Section \ref{sec:sceff-validation} for experimental setup and list of methods included in comparison).
To evaluate how these methods compare to an ideal (but impractical) model which actually applies graph partitioning to the entire search space (instead of partitioning DTG as done by GRAZHDA*), we also evaluated 
{\it IdealApprox}, a model which partitions the entire state space graph  using the METIS  (approximate) graph partitioner \cite{karypis1998fast}.
IdealApprox first enumerates a graph containing all nodes with $f \leq  f^*$ and edges between these nodes and ran METIS with the sparsity objective (Equation \ref{eq:sparsity}) to generate the partition for the work distribution. 
Generating the input graph for METIS takes an enormous amount of time (much longer than the search itself), so 
 IdealApprox is clearly an impractical model, but it provides a useful approximation for an ideal work distribution which can be used to evaluate practical methods.

Not surprisingly, IdealApprox has the highest $\sceff$, but among all of the practical methods, 
 GRAZHDA*/sparsity has the highest $\sceff$ overall.
As we saw in Section \ref{sec:sceff-validation}  
that $\sceff$ is a good estimate of actual efficiency, the result suggest that GRAZHDA*/sparsity outperforms other methods. In fact, as shown in Tables \ref{48eff} and \ref{48}, GRAZHDA*/sparsity achieved a good balance between CO and SO and had the highest actual speedup overall, significantly outperforming all other previous methods. 
Note that as IdealApprox is only an approximation of the sparsest-cut, other methods can sometimes achieve better $\sceff$. 

\subsubsection{The Effect of the Number of Cores on Speedup}
\begin{figure}[htb]
	\centering
	\includegraphics[width=0.5\linewidth]{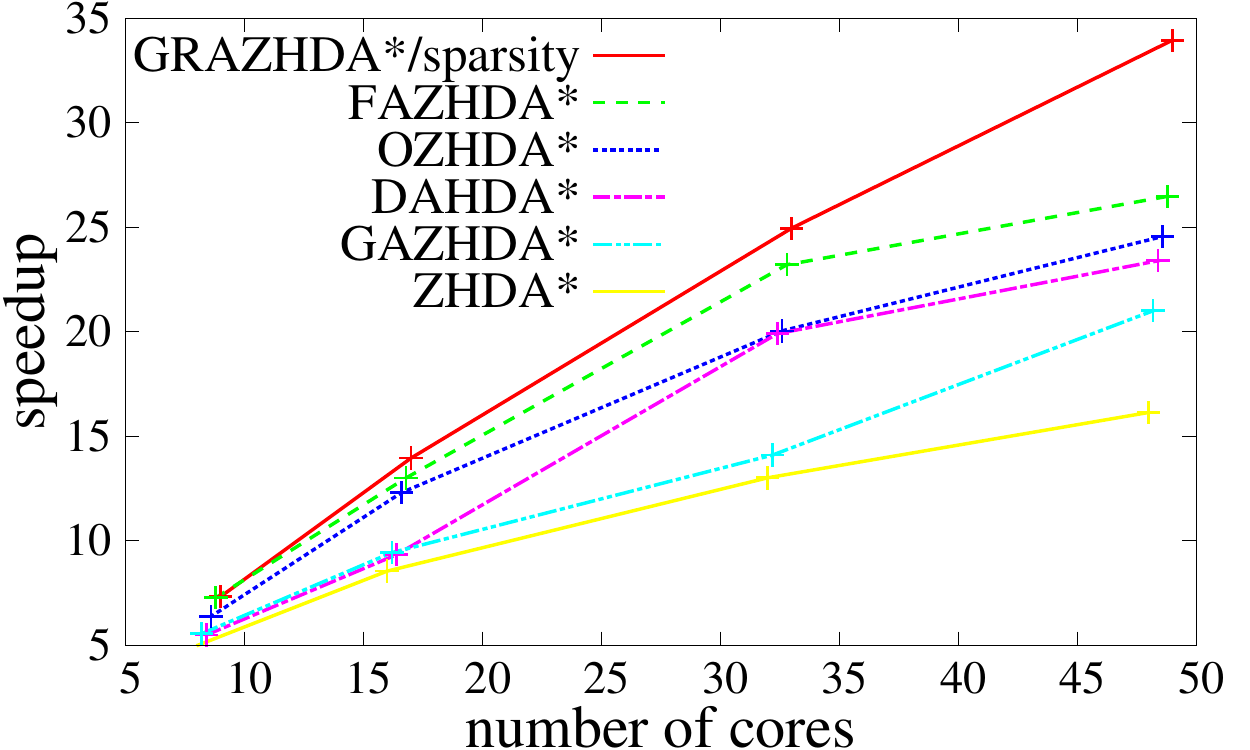} \\
	\hfil
	\caption{ Speedup of HDA* variants (average over all instances in Table \ref{48}). Results are for 1 node (8 cores), 2 nodes (16 cores), 4 nodes (32 cores) and 6 nodes (48 cores).}
	\label{fig:cores-speedup}
	\captionlistentry[experiment]{CoresSpeedup, Figure \ref{fig:cores-speedup}, lucy, 48p, M\&S-IJCAI, mpi, updatedmaterial (ICAPSnewheuristic)}

\end{figure}

Figure \ref{fig:cores-speedup} shows the speedup of the algorithms as the number of cores increased from 8 to 48.
GRAZHDA*/sparsity outperformed consistently outperformed the other methods.
The performance gap between the better methods (GRAZHDA*/sparsity, FAZHDA*, OZHDA*, DAHDA*) and the baseline ZHDA* increases with the number of the cores.
This is because as the number of cores increases, communications overheads increases with the number of cores, degrading the performance of ZHDA*, while the better work distribution methods successfully mitigate communications overhead.

\subsubsection{Cloud Environment Results}
\label{sec:cloud-results}

\begin{table}[htb]
	\caption{Comparison of walltime, communication/search overhead (CO, SO) on a cloud cluster (EC2) with 128 virtual cores (32 m1.xlarge EC2 instances) using the merge\&shrink heuristic. We run sequential A* on a different machine with 128 GB memory because some of the instances cannot be solved by A* on a single m1.xlarge instance due to memory limits. Therefore we report walltime instead of speedup.}
	\label{128}
	\centering
        \resizebox{0.8\textwidth}{!}{
	\begin{tabular}{l|r|rrr|rrr} \hline
		Instance & \multicolumn{1}{c|}{A*}  & \multicolumn{3}{c|}{GRAZHDA*/sparsity}    & \multicolumn{3}{c}{FAZHDA*} \\
		 & \multicolumn{1}{c|}{}  & \multicolumn{3}{c|}{\GRAZHDAST{}}    & \multicolumn{3}{c}{\FAZHDAT{}} \\
		          & expd & time & CO & SO  & time & CO & SO   \\ \hline
Airport18 & 48782782 & 102.34 & 0.59 & 0.49 & {\bf 95.48} & 0.59 & 0.29 \\
Blocks11-0 & 28664755 & {\bf 12.40} & 0.42 & 0.37 & 22.86 & 0.68 & 0.53 \\
Blocks11-1 & 45713730 & {\bf 17.21} & 0.42 & 0.25 & 32.60 & 0.66 & 0.82 \\
Elevators08-7 & 74610558 & {\bf 51.90} & 0.54 & 0.25 & 121.90 & 0.55 & 0.26 \\
Gripper9 & 243268770 & {\bf 78.90} & 0.42 & 0.01 & 82.90 & 0.43 & 0.06 \\
Openstacks08-21 & 19901601 & 6.30 & 0.23 & 0.06 & 5.76 & 0.19 & -0.05 \\
Openstacks11-18 & 115632865 & {\bf 33.10} & 0.24 & -0.14 & 33.25 & 0.23 & -0.12 \\
Pegsol08-29 & 287232276 & 58.85 & 0.44 & 0.16 & 81.75 & 0.42 & 0.55 \\
PipesNoTk16 & 60116156 & 120.64 & 0.94 & 0.84 & {\bf 106.28} & 0.94 & 0.72 \\
Trucks6 & 19109329 & {\bf 8.01} & 0.17 & 0.46 & 51.51 & 0.19 & 0.34 \\ \hline
Average & 99361115 & {\bf 43.03} & 0.42 & 0.25 & 59.87 & 0.48 & 0.39 \\ \hline
Total walltime & 894250040 & \multicolumn{3}{r|}{{\bf 387.31}} & \multicolumn{3}{r}{538.81} \\ 
	\end{tabular}
	}

        \resizebox{1.0\textwidth}{!}{
	\begin{tabular}{l|rrr|rrr|rrr|rrr} \hline
		Instance & \multicolumn{3}{c|}{GAZHDA*} & \multicolumn{3}{c|}{OZHDA*} & \multicolumn{3}{c|}{DAHDA*}  & \multicolumn{3}{c}{ZHDA*}  \\
		 & \multicolumn{3}{c|}{\GAZHDAT{}} & \multicolumn{3}{c|}{\OZHDAT{}} & \multicolumn{3}{c|}{\DAHDAT{}}  & \multicolumn{3}{c}{\ZHDAT{}}  \\
		          & time & CO & SO & time & CO & SO & time & CO & SO   & time & CO & SO  \\ \hline
Airport18 & 128.22 & 0.98 & 0.02 & 123.09 & 0.90 & 0.56 & 143.27 & 0.92 & 0.36 & 106.80 & 0.99 & 0.02 \\ 
Blocks11-0 & 21.75 & 0.98 & 0.65 & 21.70 & 0.99 & 0.70 & 20.29 & 0.95 & 0.88 & 29.19 & 0.99 & 0.35 \\ 
Blocks11-1 & 25.84 & 0.98 & 0.56 & 24.84 & 0.86 & 0.78 & 29.52 & 0.94 & 0.83 & 36.04 & 1.00 & 0.52 \\ 
Elevators08-7 & 61.16 & 0.70 & 0.05 & 86.65 & 0.07 & 0.22 & 52.09 & 0.96 & 0.18 & 59.88 & 1.00 & 0.04 \\ 
Gripper9 & 85.98 & 1.00 & 0.16 & 90.98 & 0.98 & 0.20 & 95.72 & 1.00 & 0.15 & 105.78 & 1.00 & 0.17 \\ 
Openstacks08-21 & {\bf 5.67} & 0.71 & -0.35 & 40.06 & 0.96 & 0.00 & 6.94 & 0.69 & -0.17 & 14.65 & 1.00 & -0.09 \\ 
Openstacks11-18 & 71.34 & 0.77 & -0.09 & 79.34 & 0.81 & -0.00 & 84.67 & 0.76 & 0.01 & 49.97 & 1.00 & -0.53 \\ 
Pegsol08-29 & 98.53 & 0.98 & 0.06 & {\bf 54.13} & 0.34 & 0.13 & 108.17 & 1.00 & 0.11 & 120.27 & 0.98 & 0.16 \\ 
PipesNoTk16 & 108.28 & 0.95 & 0.78 & 120.21 & 0.99 & 0.73 & 125.37 & 1.00 & 0.72 & 149.96 & 1.00 & 0.73 \\ 
Trucks6 & 30.22 & 0.94 & 0.41 & 32.22 & 0.96 & 0.57 & 17.19 & 0.53 & 0.43 & 28.22 & 1.00 & 0.34 \\ \hline 
Average & 56.53 & 0.89 & 0.29 & 61.13 & 0.77 & 0.41 & 60.00 & 0.87 & 0.36 & 66.00 & 1.00 & 0.29 \\ \hline
Total walltime & \multicolumn{3}{r|}{508.77} & \multicolumn{3}{r|}{550.13} & \multicolumn{3}{r|}{539.96} & \multicolumn{3}{r}{593.96} \\ 
	\end{tabular}
	}
	\captionlistentry[experiment]{EvaluationCloud, Table \ref{128}, EC2, 128p, M\&S-IJCAI, mpi, newmaterial}
\end{table}

In addition to the 48 core cluster, we evaluated GRAZHDA*/sparsity on an Amazon EC2 cloud cluster with 128 virtual cores (vCPUs) and 480GB aggregated RAM (a cluster of 32 m1.xlarge EC2 instances, each with 4 vCPUs, 3.75 GB RAM/core. 
This 
is a less favorable environment for parallel search compared to a ``bare-metal'' cluster because physical processors are shared with other users and network performance is inconsistent \cite{iosup2011performance}. 
We intentionally chose this configuration 
in order to evaluate work distribution methods in an environment which is significantly different from our other experiments.
%
Table \ref{128} shows that as with the smaller-scale cluster results, GRAZHDA*/sparsity outperformed other methods in this large-scale cloud environment.


\subsubsection{24-Puzzle Experiments}

\begin{figure}[htb]
\captionsetup{justification=centering}
	\centering
	\subfloat[15-puzzle \ZHDA{}]{\hspace{8pt}{\includegraphics[width=0.15\linewidth]{illustration/15puzzle/zbr.pdf}}\hspace{8pt}} \hspace{12pt}
	\subfloat[15-puzzle GRAZHDA*/sparsity]{\label{fig:15sazhda} \hspace{12pt}{\includegraphics[width=0.15\linewidth]{illustration/15puzzle/az8.pdf}} \hspace{12pt}} \hspace{12pt}
	\subfloat[24-puzzle \ZHDA{}]{\hspace{8pt}{\includegraphics[width=0.15\linewidth]{illustration/24puzzle/24zbr.pdf}}\hspace{8pt}} \hspace{12pt}
	\subfloat[24-puzzle GRAZHDA*/sparsity]{\label{fig:24sazhda} \hspace{12pt} {\includegraphics[width=0.15\linewidth]{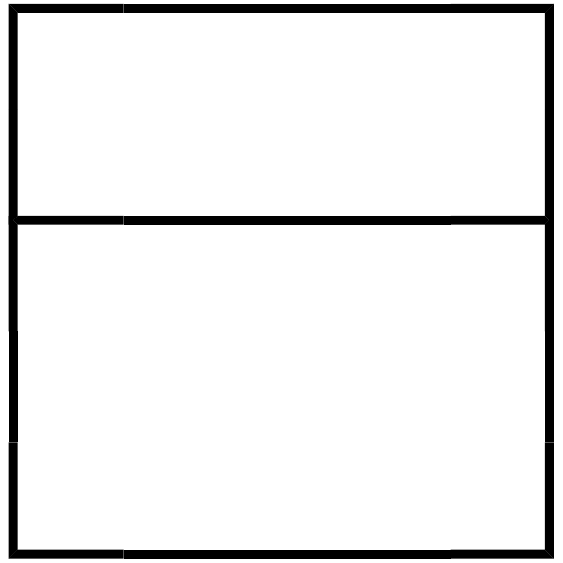}} \hspace{12pt} }%

\captionsetup{justification=justified}
	\caption{Abstract features generated by GRAZHDA*/sparsity (\GRAZHDAS{}) for 15-puzzle and 24-puzzle. Abstract features generated on 15-puzzle exactly corresponds to the hand-crafted hash function of Figure \ref{fig:15azh}.}
	\label{fig:sparsestafg-tile}
\end{figure}

We evaluated GRAZHDA*/sparsity on the 24-puzzle using the same configuration as Section \ref{sec:24puzzle}.
Abstract feature generated by GRAZHDA*/sparsity is shown in Figure \ref{fig:24sazhda}.
We compared GR\-AZHDA*/sparsity (automated abstract feature generation) vs.  AZHDA* with the hand-crafted work distribution (\AZHDA{}) (Figure \ref{fig:24azh}) and \ZHDA{}. 
With 8 cores, the speedups were 7.84 (GRAZHDA*/sparsity), 7.85 (\AZHDA{}), and 5.95 (\ZHDA{}).
Thus, the completely automated GRAZHDA*/sparsity is  competitive with a carefully hand-designed work distribution method.
For the 15-puzzle, the partition generated by GRAZHDA*/sparsity exactly corresponds to the hand-crafted hash function of Figure \ref{fig:15azh}, so the performance is identical.

\subsubsection{Evaluation of Parallel Search Overheads and Performance in Low Communications-Cost Environments}
\label{sec:lmcut}

\begin{table}[htbp]
	\caption{Comparison of speedups, communication/search overheads (CO, SO) using expensive heuristic (LM-cut).}
	\centering
	\subfloat[Single multicore machine (8 cores)] {
	\label{lmcut-8}
        \resizebox{0.94\textwidth}{!}{
	\begin{tabular}{l|rr|rrr|rrr|rrr} \hline
		Instance & \multicolumn{2}{c|}{A*}  & \multicolumn{3}{c|}{GRAZHDA*/sparsity}    & \multicolumn{3}{c|}{DAHDA*} & \multicolumn{3}{c}{ZHDA*} \\
		 & \multicolumn{2}{c|}{}  & \multicolumn{3}{c|}{\GRAZHDAST{}}    & \multicolumn{3}{c|}{\DAHDAT{}} & \multicolumn{3}{c}{\ZHDAT{}} \\
		         & time & expd & speedup & CO & SO  & speedup & CO & SO   & speedup & CO & SO \\ \hline
Blocks14-1 & 351.03 & 191948 & {\bf 5.53} & 0.33 & 0.19 & 2.08 & 0.30 & 1.82 & 5.40 & 0.90 & 0.05 \\ 
Elevators08-7 & 1182.92 & 1465914 & 3.47 & 0.48 & 1.70 & 3.30 & 0.73 & 0.04 & {\bf 3.75} & 0.88 & 0.00 \\ 
Elevators08-8 & 742.65 & 344304 & {\bf 7.19} & 0.41 & 0.06 & 4.80 & 0.72 & 0.03 & 5.65 & 0.82 & 0.00 \\ 
Floortile11-4 & 1783.44 & 2876492 & 3.54 & 0.50 & 0.28 & {\bf 4.03} & 0.41 & 0.01 & 3.17 & 0.96 & 0.00 \\ 
Gripper7 & 903.96 & 10082501 & 1.41 & 0.68 & 0.27 & {\bf 2.60} & 0.56 & 0.00 & 2.27 & 0.94 & 0.00 \\ 
Openstacks08-15 & 707.31 & 11309809 & {\bf 4.95} & 0.32 & -0.07 & 4.26 & 0.27 & 0.03 & 3.91 & 0.88 & -0.04 \\ 
Openstacks11-12 & 309.49 & 4250213 & {\bf 4.59} & 0.38 & -0.00 & 4.40 & 0.29 & -0.00 & 3.94 & 0.92 & -0.01 \\ 
Openstacks11-15 & 1187.58 & 13457961 & 4.04 & 0.36 & 0.10 & {\bf 4.09} & 0.28 & 0.01 & 3.59 & 0.89 & 0.00 \\ 
PipesNoTk10 & 997.62 & 662717 & 2.65 & 0.86 & 0.00 & 2.10 & 0.96 & 0.01 & {\bf 3.02} & 0.89 & 0.00 \\ 
PipesNoTk12 & 201.07 & 200502 & 4.36 & 0.84 & 0.00 & 4.65 & 0.47 & 0.09 & {\bf 4.69} & 0.90 & 0.00 \\ 
PipesNoTk15 & 323.59 & 212678 & 4.61 & 0.85 & 0.00 & 3.83 & 0.57 & 0.22 & {\bf 4.91} & 0.89 & 0.01 \\ 
PipesTk11 & 572.00 & 382587 & {\bf 6.45} & 0.37 & 0.01 & 3.57 & 0.64 & 0.00 & 3.69 & 0.86 & 0.00 \\ 
Scanalyzer11-6 & 1149.31 & 699932 & {\bf 5.89} & 0.13 & -0.01 & 3.14 & 0.44 & -0.00 & 2.75 & 0.88 & -0.00 \\ 
Storage15 & 330.79 & 155979 & 4.70 & 0.70 & 0.04 & 4.67 & 0.68 & 0.01 & {\bf 4.95} & 0.85 & 0.00 \\ 
Trucks9 & 199.02 & 65531 & {\bf 7.38} & 0.05 & -0.04 & 3.72 & 0.06 & 0.42 & 3.40 & 0.87 & -0.01 \\ 
Trucks10 & 800.02 & 384585 & {\bf 6.85} & 0.04 & 0.01 & 4.42 & 0.04 & 0.15 & 2.03 & 0.91 & 0.03 \\ 
Visitall11-7half & 181.05 & 519064 & {\bf 6.59} & 0.14 & 0.24 & 5.62 & 0.16 & 0.15 & 6.09 & 0.87 & 0.00 \\ 
Woodwrk11-6 & 283.73 & 172077 & {\bf 7.10} & 0.39 & -0.00 & 5.97 & 0.27 & -0.00 & 3.25 & 0.94 & -0.00 \\ \hline
Average & 678.14 & 2635266 & {\bf 5.07} & 0.43 & 0.15 & 3.96 & 0.44 & 0.17 & 3.91 & 0.89 & 0.00 \\ 
\hline 
Total walltime & 12206.58 & 47434794 & \multicolumn{3}{r|}{{\bf 3215.00}} & \multicolumn{3}{r|}{3513.00} & \multicolumn{3}{r}{3711.51}  \\
	\end{tabular}
	}
	\captionlistentry[experiment]{LMCUT:multicore, Table \ref{lmcut-8}, lucy, 8p, LMCUT, mpi, newmaterial}
	}

	\subfloat[Commodity Cluster with 6 nodes (48 cores)] {
	\label{lmcut-48}
	\centering
        \resizebox{0.94\textwidth}{!}{
	\begin{tabular}{l|rr|rrr|rrr|rrr} \hline
		Instance & \multicolumn{2}{c|}{A*}  & \multicolumn{3}{c|}{GRAZHDA*/sparsity}    & \multicolumn{3}{c|}{DAHDA*} & \multicolumn{3}{c}{ZHDA*} \\
		 & \multicolumn{2}{c|}{}  & \multicolumn{3}{c|}{\GRAZHDAST{}}    & \multicolumn{3}{c|}{\DAHDAT{}} & \multicolumn{3}{c}{\ZHDAT{}} \\
		         & time & expd & speedup & CO & SO  & speedup & CO & SO   & speedup & CO & SO \\ \hline

Blocks14-1 & 351.03 & 191948 & {\bf 22.86} & 0.34 & 0.50 & 20.22 & 0.32 & 0.65 & 16.79 & 0.98 & 0.18 \\ 
Elevators08-7 & 1182.92 & 1465914 & 18.17 & 0.53 & 0.36 & {\bf 22.25} & 0.81 & 0.07 & 20.91 & 0.97 & 0.02 \\ 
Elevators08-8 & 742.65 & 344304 & 25.58 & 0.45 & 0.51 & 30.78 & 0.84 & 0.10 & {\bf 31.43} & 0.95 & 0.05 \\ 
Floortile11-4 & 1783.44 & 2876492 & 18.25 & 0.99 & 0.09 & {\bf 24.65} & 0.46 & 0.10 & 21.56 & 0.99 & 0.02 \\ 
Gripper7 & 903.96 & 10082501 & 12.59 & 0.66 & 0.02 & {\bf 16.17} & 0.61 & 0.07 & 12.59 & 0.99 & 0.01 \\ 
Openstacks11-11 & 721.30 & 11309809 & {\bf 43.09} & 0.36 & -0.43 & 10.19 & 0.25 & 1.22 & 20.75 & 0.99 & -0.02 \\ 
Openstacks11-15 & 1187.58 & 13457961 & 15.50 & 0.28 & 0.28 & 17.39 & 0.23 & 0.25 & {\bf 19.53} & 0.99 & 0.01 \\ 
Parcprinter11-12 & 195.51 & 218595 & {\bf 46.90} & 0.04 & 0.02 & 44.14 & 0.05 & 0.01 & 18.02 & 0.99 & 0.24 \\ 
PipesNoTk10 & 997.62 & 662717 & {\bf 15.57} & 0.98 & 0.01 & 14.80 & 0.99 & 0.01 & 15.38 & 0.98 & 0.01 \\ 
PipesNoTk12 & 201.07 & 200502 & 26.05 & 0.89 & 0.28 & {\bf 32.86} & 0.52 & 0.34 & 22.03 & 0.98 & 0.30 \\ 
PipesNoTk15 & 323.59 & 212678 & {\bf 25.18} & 0.94 & 0.18 & 19.54 & 0.62 & 0.72 & 21.11 & 0.98 & 0.39 \\ 
PipesTk8 & 1141.00 & 145828 & 17.96 & 0.98 & 0.04 & 17.38 & 0.98 & 0.06 & {\bf 18.99} & 0.98 & 0.03 \\ 
PipesTk11 & 572.00 & 382587 & {\bf 30.35} & 0.41 & 0.16 & 23.62 & 0.65 & 0.06 & 19.31 & 0.98 & 0.04 \\ 
Scanalyzer11-06 & 1149.31 & 699932 & {\bf 42.21} & 0.13 & 0.04 & 20.18 & 0.49 & 0.01 & 15.45 & 0.98 & 0.00 \\ 
Storage15 & 330.79 & 155979 & 22.50 & 0.88 & 0.22 & {\bf 30.35} & 0.74 & 0.09 & 24.15 & 0.96 & 0.19 \\ 
Trucks9 & 199.02 & 65531 & {\bf 24.82} & 0.05 & 0.78 & 18.92 & 0.06 & 1.42 & 12.24 & 0.98 & 0.96 \\ 
Trucks10 & 800.02 & 384585 & 17.61 & 0.05 & 0.60 & {\bf 41.74} & 0.04 & 0.25 & 12.53 & 1.00 & 0.04 \\ 
Visitall11-07half & 181.05 & 519064 & 12.97 & 0.16 & 2.59 & 12.88 & 0.17 & 2.60 & {\bf 22.14} & 0.98 & 0.58 \\ 
Woodwrk08-7 & 819.62 & 33871 & {\bf 36.12} & 0.74 & 0.07 & 31.91 & 0.74 & 0.37 & 26.71 & 1.00 & 0.07 \\ 
Woodwrk11-6 & 283.73 & 172077 & {\bf 42.67} & 0.42 & 0.07 & 21.38 & 0.29 & 0.03 & 16.81 & 0.99 & 0.05 \\
\hline 
Average & 756.91 & 2527318 & {\bf 26.06} & 0.55 & 0.17 & 23.98 & 0.52 & 0.24 & 19.26 & 0.98 & 0.09 \\ 
\hline 
Total walltime & 12867.52 & 42964409 & \multicolumn{3}{r|}{{\bf 637.57}} & \multicolumn{3}{r|}{646.66} & \multicolumn{3}{r}{709.89}  \\ 

	\end{tabular}
	}
	\captionlistentry[experiment]{LMCUT:cluster, Table \ref{lmcut-48}, lucy, 48p, LMCUT, mpi, newmaterial}
	}
\end{table}

In previous experiments,  we compared work distribution functions  using domain-specific solvers with very fast node generation rates (Section \ref{sec:experiments-comb}), 
as well as domain-independent planning using a fast heuristic function (Section \ref{sec:experiments-planning}).
Next, we evaluate search overheads and performance when node generation rates are low due to expensive node evaluations.
In such domains, the impact of communications overheads is minimal because overheads for queue insertion, buffering, etc. are negligible compared to the computation costs associated with node generation and evaluation.
As a consequence, search overhead is the dominant factor which determines search performance.

In particular, we evaluate different parallel work distribution strategies when applied to domain-independent planning
using the landmark-cut (LM-cut) heuristic, a state-of-the-art heuristic which is a relatively expensive heuristic.
While there is no known dominance relationship among planners using cheap heuristics such as merge\&shrink heuristics (which require only a table lookup during search)
and expensive heuristics  such as LM-cut, recent work in forward-search based planning has focused on heuristics which tend to be slow, such as heuristics that require the solution of a linear program at every search node \cite{PommereningRHB2014,ImaiF15}, so parallel strategies that focus on minimizing search overheads is of practical importance.
Previous evaluations of parallel work distribution strategies in domain-independent planning used relatively fast heuristics. 
\citeauthor{kishimotofb13} \citeyear{kishimotofb13}, as well as \citeauthor {jinnai2016structured} \citeyear{jinnai2016structured,jinnai2016automated} used merge\&shrink abstraction based heuristics. \citeauthor{zhou2007parallel} \citeyear{zhou2007parallel} and \citeauthor{burnslrz10} \citeyear{burnslrz10} used the max-pair heuristic \cite{HaslumG00}. 
Thus, to our knowledge, this is the first evaluation of parallel forward search for domain-independent planning using an expensive heuristic.


To evaluate the effect of SO and CO with the LM-cut heuristic, we compared the performance of ZHDA*, DAHDA*, and GRAZHDA*/sparsity as representatives of methods which optimize SO, CO, and both SO and CO, respectively.
The instances used for this experiment are different from the experiments using  merge\&shrink (Table \ref{48}), because some of the instances used for the merge\&shrink experiments were too easy to solve with LM-cut and not suitable for evaluating parallel algorithms.
The average node expansion rate by sequential A* on the selected instances was 3886.02 node/sec.
Compared to the expansion rate with the merge\&shrink heuristic used in Section \ref{sec:experiments-planning} (56378.03 node/sec), the expansion rate is 14.5 times slower.
Therefore, the relative cost of communication is expected to be significantly smaller when using the  LM-cut heuristic, compared to using the merge\&shrink heuristic.

Table \ref{lmcut-8} shows the results on a single multicore machine with 8 cores.
Overall, GRAZHDA*/sp\-arsity outperformed ZHDA* and DAHDA*.
Interestingly, although GRAZHDA*/sparsity has higher SO, it was still faster than ZHDA* because of lower CO.
Even in this low communication cost environment, CO continues to be a significant overhead for HDA*.

Table \ref{lmcut-48} shows the results on a commodity cluster with 48 cores.
As in the multicore environment, GRAZHDA*/sparsity outperformed ZHDA* and DAHDA*.
However, the relative speedup of ZHDA* to GRAZHDA*/sparsity is higher with LM-cut (0.75) than with merge\&shrink (0.66) (note that we used different instance set, so it may due to other factors).
Some of the instances (\pddl{trucks9}, \pddl{visitall11-07-half}) are too easy for a distributed environment, and therefore on these instances, high SO is incurred due to the burst effect (Section \ref{sec:burst}). Therefore, some of the instances have high SO even in ZHDA* where good LB is achieved.

\section{Conclusions}
\label{sec:conclusions}


We investigated node distribution methods for HDA*, and showed that previous methods suffered from high communication overhead (\ZHDA{}), high search overhead (\AHDA{}), or both (\PHDA{}), which limited their efficiency.
We proposed  Abstract Zobrist hashing, a new distribution method which combines the strengths of both Zobrist hashing and abstraction, and AZHDA* (\AZHDA{}), a new variant of HDA* which is based on AZH.
Our experimental results showed that AZHDA* achieves a successful trade-off between communication and search overheads, resulting in better performance than previous work distribution methods with hand-crafted abstract features.

We then extended the investigation to automated, domain-independent approaches for  generate work distribution.
We formulated work distribution as graph partitioning, and proposed and validated  $\sceff$, a model of search and communication overheads for HDA* which can be used to predict the actual walltime efficiency.  
We proposed and evaluated GRAZHDA*, a new top-down approach to work distribution for parallel best-first search in the HDA* framework which approximate the optimal graph partitioning by partitioning domain transition graphs according to an objective function such as sparsity.


We experimentally showed that GRAZHDA*/sparsity significantly improves  both estimated efficiency ($\sceff$) as well as the actual performance (walltime efficiency) compared to previous work distribution methods.
Our results demonstrate the viability of  approximating the partitioning of the entire search space by  applying graph partitioning to an abstraction  of the state space (i.e., the DTG). 
While our results indicate that sparsity works well as a partitioning objective for GRAHZDA*, it is possible that a different objective function might yield better results, since DTG-partitioning is only an approximation to $\wg$ partitioning.
We have experimented with another objective,  MIN(CO+LB), which minimizes $(CO+LB)$, and found that the performance is  comparable to sparsity. Investigation of other objective functions is a direction for future work.

Despite significant improvements compared to previous work distribution approaches, there is room for improvement.
The gap between the $\sceff$ metric for GRAZHDA* and an  ideal model (IdealApprox) in Figure \ref{fig:sceffs} represents the gap between 
actually partitioning the state space graph (as IdealApprox does) vs. the approximation obtained by the GRAZHDA* DTG partitioning. Closing this gap in $\sceff$  should lead to corresponding improvements in actual walltime efficiency, and poses challenges for future work.
One possible approach to closing this gap is to partition a merged DTG which represents multiple SAS+ variables instead of partitioning a DTG of a single SAS+ variable. 
As merged DTGs have a richer representation of the state space graph, partitioning them using an objective function may result in a better approximation of the ideal partitioning. 
This approach is similar to merge-and-shrink heuristic \cite{helmert2014merge} which merging multiple DTGs into abstract state space to better estimate the state-space graph.


In this paper, we assumed identical distance between each two cores.
However, communication costs vary among pairs of processors in distributed search, especially in cloud cluster environments. Furthermore, as the number of cores scales to thousands or tens of thousands or more, some consideration of core locality is likely to be necessary.
Incorporating the technique to distribute nodes considering the locality of processors such as LOHA\&QE \cite{mahapatra1997scalable} may further improve the performance.
Finally,  GPU-based massively parallel search has recently been shown to be successful \cite{zhou2015massively,HorieF17}. Investigation of tradeoffs between communication and search overhead in a heterogeneous algorithm which seeks to effectively utilize all normal cores as well as GPU cores using a framework based on abstract feature-based hashing  is a direction for future work.



\begin{appendices}
\appendix
\section{Dynamic AHDA* (DAHDA*), an Improvement to AHDA* for Distributed Memory Systems}
\label{appendix:dahda*}

This section presents an improvement to AHDA* \cite{burnslrz10}. 
In our experiments, we used AHDA* as one of the baselines for evaluating our new AZHDA* strategies.
The baseline implementation of AHDA* (\AHDAZSSD{}) is based on the greedy abstraction algorithm described in \cite{Zhou2006}, and
selects a subset of DTGs (atom groups). 
The greedy abstraction algorithm adds one DTG to the abstract graph ($G$) at a time, choosing the DTG which minimizes the maximum out-degree of the abstract graph, until the graph size (\# of nodes) reaches the threshold given by a parameter $N_{max}$.
PSDD requires a $N_{max}$ to be derived from the size of the available RAM.
We found that AHDA* with a static $N_{max}$ threshold as in PSDD performed poorly for a benchmark set with varying difficulty because a fixed size abstract graph results in very poor load balance.
While poor load balance can lead to low efficiency and poor performance, 
a bad choice for $N_{max}$ can be catastrophic when the system has a relatively small amount of RAM per core, as poor load balance causes concentrated memory usage in the overloaded processors, resulting in early memory exhaustion (i.e., AHDA* crashes because a thread/process which is allocated a large number of states exhausts its local heap).


The AHDA* results in Table \ref{dahda} are for a 48-core cluster, 2GB/core, and uses $N_{max}=10^2$, $10^3$, $10^4$, $10^5$, $10^6$ nodes based on Fast-Downward \cite{Helmert2006} using merge\&shrink heuristic \cite{helmert2014merge}. 
Smaller $N_{max}$ results in lower CO, but when $N_{max}$ is too small for the problem, load imbalance results in a concentration of the nodes and memory exhaustion.
Although the total amount of RAM in current systems is growing, the amount of RAM per core has remained relatively small because the number of cores has also been increasing (and is expected to continue increasing).
Thus, this is a significant issue  with the straightforward implementation of AHDA* which uses a static $N_{max}$.
To avoid this problem, $N_{max}$ must be set dynamically according to the size of the state space for each instance.
Thus, we implemented Dynamic AHDA* (DAHDA* = \DAHDA), which dynamically set the size of the abstract graph according to the number of DTGs (the state space size is exponential in the number of DTGs).
We set the threshold of the total number of features in the DTGs to be 30\% of the total number of features in the problem instance
(we tested 10\%, 30\%, 50\%, and 70\% and found that 30\% performed best). 
Note that the threshold is relative to the number of features, not the state space size as in AHDA*, which is exponential in the number of features.
Therefore, DAHDA* tries to take into account of certain amount of features, whereas AHDA* sometimes use only a fraction of features.


\begin{table}[htb]
	\caption{
Performance of AHDA* with fixed threshold (on 48 cores). Note that $|G| > |G'|$ does not indicate that all atom groups used in $G$ are used in $G'$. DAHDA* limits the size of abstract graph according to the number of features in abstract graph, whereas AHDA* set maximum to $N_{max}$. Due to this difference, DAHDA* tends to end up with a different set of atom groups than AHDA*.
}
	\label{dahda}
	\centering
        \resizebox{0.9\textwidth}{!}{
	\begin{tabular}{l|rr|rrrr|rrr} \hline
		Instance & \multicolumn{2}{c|}{A*}  & \multicolumn{4}{c|}{DAHDA*}   & \multicolumn{3}{c}{AHDA*}\\ 
		 & \multicolumn{2}{c|}{}  & \multicolumn{4}{c|}{\DAHDAT{}}   & \multicolumn{3}{c}{\AHDAT{}}\\ 
		         & \multicolumn{2}{c|}{}  & \multicolumn{4}{c|}{}      & \multicolumn{3}{c}{$N_{max} = 100$}    \\
		         & time & expd              & speedup & CO & SO & $|G|$  & speedup & CO & SO            \\ \hline
Blocks10-0 & 129.29 & 11065451 & {\bf 25.11} & 0.88 & 0.08 & 14641 & 5.61 & 0.48 & 4.72  \\
Blocks11-1 & 621.74 & 52736900 & {\bf 24.88} & 0.91 & 0.21 & 20736 & \multicolumn{3}{c}{memory exhaustion} \\
Elevators08-5 & 165.22 & 7620122 & {\bf 27.59} & 0.83 & -0.03 & 1500 & 6.54 & 0.61 & 1.84  \\
Elevators08-6 & 453.21 & 18632725 & 15.28 & 0.88 & 0.31 & 73125 & \multicolumn{3}{c}{memory exhaustion} \\
Gripper8 & 517.41 & 50068801 & 21.80 & 0.98 & 0.08 & 39366 & 16.84 & 0.39 & 0.58  \\
Logistics00-10-1 & 559.45 & 38720710 & {\bf 17.52} & 0.84 & 0.00 & 140608 & \multicolumn{3}{c}{memory exhaustion} \\
Miconic11-0 & 232.07 & 12704945 & {\bf 46.05} & 0.01 & 0.08 & 2048 & 7.61 & 0.00 & 5.39  \\
Nomprime5 & 309.14 & 4160871 & 18.46 & 0.90 & -0.05 & 4194304 & \multicolumn{3}{c}{memory exhaustion} \\
Openstacks08-21 & 554.63 & 19901601 & {\bf 26.72} & 0.13 & 0.28 & 8388608 & \multicolumn{3}{c}{memory exhaustion} \\
PipesNoTk10 & 157.31 & 2991859 & 15.58 & 0.98 & 0.01 & 32768 & 5.89 & 0.17 & 1.15  \\
Scanalyzer08-6 & 195.49 & 10202667 & 21.23 & 0.94 & 0.00 & 16384 & {\bf 40.15} & 0.02 & -0.07  \\
	\end{tabular}
	}

        \resizebox{1.0\textwidth}{!}{
	\begin{tabular}{l|rrr|rrr|rrr|rrr} \hline
			 & \multicolumn{12}{c}{AHDA*} \\
			 & \multicolumn{12}{c}{\AHDAT{}} \\
		         & \multicolumn{3}{c|}{$N_{max} = 1000$} & \multicolumn{3}{c|}{$N_{max} = 10000$} & \multicolumn{3}{c|}{$N_{max} = 100000$}  & \multicolumn{3}{c}{$N_{max} = 1000000$}  \\
		         & speedup & CO & SO        & speedup & CO & SO            & speedup & CO & SO            & speedup & CO & SO       \\ \hline
Blocks10-0 & \multicolumn{3}{c|}{memory exhaustion} & 18.24 & 0.39 & 0.29 & 16.72 & 0.47 & 0.25 & 14.77 & 0.54 & 0.24 \\ 
Blocks11-1 & \multicolumn{3}{c|}{memory exhaustion} & \multicolumn{3}{c|}{memory exhaustion} & 21.38 & 0.65 & 0.12 & 15.38 & 0.68 & 0.10 \\ 
Elevators08-5 & 18.02 & 0.79 & 0.65 & 19.30 & 0.87 & 0.44 & 18.20 & 0.90 & 0.37 & 18.84 & 0.92 & 0.35 \\ 
Elevators08-6 & 17.86 & 0.67 & 0.50 & 18.38 & 0.86 & 0.23 & 16.99 & 0.91 & 0.09 & {\bf 22.66 } & 0.89 & -0.02 \\ 
Gripper8 & \multicolumn{3}{c|}{memory exhaustion} & {\bf 30.17} & 0.53 & 0.31 & 25.31 & 0.65 & 0.21 & 24.65 & 0.70 & 0.16 \\
Logistics00-10-1 & \multicolumn{3}{c|}{memory exhaustion} & \multicolumn{3}{c|}{memory exhaustion} & \multicolumn{3}{c|}{memory exhaustion} & \multicolumn{3}{c}{memory exhaustion} \\
Miconic11-0 & 6.75 & 0.00 & 5.60 & 26.90 & 0.01 & 0.19 & 26.22 & 0.01 & 0.25 & 25.77 & 0.02 & 0.40 \\ 
Nomprime5 & 18.28 & 0.31 & 0.07 & 16.47 & 0.43 & 0.10 & {\bf 19.92} & 0.58 & 0.01 & 17.07 & 0.60 & 0.00 \\ 
Openstacks08-21 & \multicolumn{3}{c|}{memory exhaustion} & \multicolumn{3}{c|}{memory exhaustion} & \multicolumn{3}{c|}{memory exhaustion} & \multicolumn{3}{c}{memory exhaustion} \\
PipesNoTk10 & 18.38 & 0.30 & 0.10 & {\bf 21.74} & 0.43 & 0.04 & 18.50 & 0.56 & 0.03 & 14.36 & 0.64 & 0.03 \\ 
Scanalyzer08-6 & 38.11 & 0.03 & -0.03 & 39.26 & 0.26 & -0.07 & 30.17 & 0.47 & -0.07 & 26.46 & 0.64 & -0.07 \\
	\end{tabular}
	}

	\captionlistentry[experiment]{DAHDA*vsAHDA*, Table \ref{dahda}, lucy, 48p, M\&S-ICAPS, mpi, newmaterial (ICAPSsupplement)} 
\end{table}
\captionsetup{labelformat=empty,labelsep=none}


\clearpage
\newpage

\section{Experimental Results with Standard Deviations}
\begin{table}[h]
	\caption{Table 11: Comparison of speedups, communication/search overhead (CO, SO) and their standard deviations on a commodity cluster with 6 nodes, 48 processes using merge\&shrink heuristic (Expanded version of Table \ref{48}).}

	\label{tab:48-results-with-sd}
	\centering
        \resizebox{1.0\textwidth}{!}{
	\begin{tabular}{l|rr|rrr|rrr} \hline
		Instance & \multicolumn{2}{c|}{A*}  & \multicolumn{3}{c|}{GRAZHDA*/sparsity}    & \multicolumn{3}{c}{FAZHDA*} \\
		 & \multicolumn{2}{c|}{}  & \multicolumn{3}{c|}{\GRAZHDAST{}}    & \multicolumn{3}{c}{\FAZHDAT{}} \\
	& time & expd & speedup & CO & SO  & speedup & CO & SO   \\ \hline
Blocks10-0   &      129.29  &  11065451  &  {\bf 27.17}  (4.11) &  0.28  (0.02) &  0.38  (0.41) &  26.02  (0.74) &  0.70  (0.00) &  0.35  (0.04) \\
Blocks11-1   &      813.86  &  52736900  &  {\bf 34.25}  (3.54) &  0.66  (0.00) &  0.15  (0.13) &  34.25  (0.64) &  0.66  (0.00) &  0.15  (0.03) \\
Elevators08-5  &      165.22  &   7620122  &  16.43  (2.81) &  0.47  (0.01) &  0.33  (0.06) &  12.34  (0.24) &  0.32  (0.00) &  0.51  (0.01) \\
Elevators08-6  &      453.21  &  18632725  &  21.47  (0.90) &  0.49  (0.00) &  0.37  (0.04) &  18.05  (0.61) &  0.52  (0.00) &  0.81  (0.09) \\
Gripper8     &      517.41  &  50068801  &  26.67  (0.75) &  0.50  (0.00) &  0.15  (0.08) &  {\bf 27.45}  (0.73) &  0.43  (0.00) &  0.10  (0.12) \\
Logistics00-10-1  &      559.45  &  38720710  &  {\bf 45.16}  (3.28) &  0.43  (0.00) &  0.01  (0.03) &  43.85  (3.05) &  0.57  (0.00) &  0.02  (0.00) \\
Miconic11-0  &      232.07  &  12704945  &  41.97  (0.54) &  0.01  (0.00) &  0.07  (0.01) &  42.43  (0.57) &  0.01  (0.00) &  0.06  (0.01) \\
Miconic11-2  &      262.01  &  14188388  &  {\bf 45.26}  (0.60) &  0.01  (0.00) &  0.05  (0.00) &  44.87  (1.18) &  0.01  (0.00) &  0.05  (0.01) \\
NoMprime5    &      309.14  &   4160871  &  {\bf 23.95}  (0.85) &  0.80  (0.00) &  -0.04  (0.02) &  22.87  (2.98) &  0.79  (0.00) &  -0.05  (0.03) \\
Nomystery10  &      179.52  &   1372207  &  {\bf 34.80}  (0.87) &  0.51  (0.00) &  0.12  (0.03) &  22.99  (4.55) &  0.24  (0.00) &  -0.44  (0.10) \\
Openstacks08-19  &      282.45  &  15116713  &  24.67  (1.25) &  0.27  (0.01) &  0.34  (0.05) &  20.00  (0.86) &  0.24  (0.00) &  0.37  (0.04) \\
Openstacks08-21  &      554.63  &  19901601  &  25.23  (0.51) &  0.17  (0.00) &  0.35  (0.03) &  24.97  (0.42) &  0.15  (0.00) &  0.35  (0.02) \\
Parcprinter11-11  &      307.19  &   6587422  &  {\bf 20.26}  (0.93) &  0.26  (0.00) &  0.55  (0.29) &  13.08  (4.09) &  0.26  (0.03) &  0.61  (0.67) \\
Parking11    &      237.05  &   2940453  &  {\bf 29.75}  (0.48) &  0.40  (0.00) &  0.34  (0.01) &  29.67  (4.12) &  0.63  (0.00) &  0.11  (0.10) \\
Pegsol11-18  &      801.37  &  106473019  &  21.03  (0.65) &  0.39  (0.00) &  0.02  (0.01) &  20.97  (0.21) &  0.39  (0.00) &  0.00  (0.01) \\
PipesNoTk10  &      157.31  &   2991859  &  {\bf 15.73}  (0.38) &  0.98  (0.00) &  0.01  (0.00) &  15.64  (0.35) &  0.98  (0.00) &  0.01  (0.00) \\
PipesTk12    &      321.55  &  15990349  &  33.78  (4.22) &  0.46  (0.00) &  0.05  (0.01) &  {\bf 39.65}  (2.65) &  0.46  (0.00) &  0.03  (0.01) \\
PipesTk17    &      356.14  &  18046744  &  43.92  (2.69) &  0.54  (0.00) &  0.01  (0.00) &  {\bf 45.03}  (3.81) &  0.54  (0.00) &  0.01  (0.00) \\
Rovers6      &     1042.69  &  36787877  &  {\bf 41.17}  (2.51) &  0.15  (0.00) &  0.14  (0.09) &  40.48  (1.40) &  0.15  (0.00) &  0.17  (0.04) \\
Scanalyzer08-6  &      195.49  &  10202667  &  {\bf 32.92}  (0.74) &  0.12  (0.00) &  0.01  (0.00) &  30.31  (0.56) &  0.12  (0.00) &  0.01  (0.00) \\
Scanalyzer11-6  &      152.92  &   6404098  &  {\bf 43.83}  (0.54) &  0.16  (0.00) &  0.13  (0.00) &  27.31  (1.68) &  0.18  (0.00) &  0.34  (0.05) \\ \hline
Average      &      382.38  &  21557805  &  {\bf 30.92}  (1.58) &  0.38  (0.00) &  0.17  (0.06) &  28.68  (1.69) &  0.40  (0.00) &  0.17  (0.07) \\ \hline
Total walltime & 8029.97 & 452713922 & \multicolumn{3}{r|}{{\bf 277.91} (14.20)} & \multicolumn{3}{r}{301.38 (17.65)} \\
	\end{tabular}
	}
\end{table}

\begin{table}[t]
	\centering
	\caption{Table 11. (Cont.)}
        \resizebox{0.9\textwidth}{!}{
	\begin{tabular}{l|rrr|rrr} \hline
		Instance & \multicolumn{3}{c|}{GAZHDA*}  & \multicolumn{3}{c}{OZHDA*} \\
		 & \multicolumn{3}{c|}{\GAZHDAT{}}  & \multicolumn{3}{c}{\OZHDAT{}} \\
	& speedup & CO & SO & speedup & CO & SO  \\ \hline
Blocks10-0   &  21.81  (3.26) &  0.99  (0.00) &  0.12  (0.30) &  15.47  (4.37) &  0.98  (0.00) &  0.34  (0.34)  \\
Blocks11-1   &  29.20  (3.22) &  0.99  (0.00) &  0.03  (0.16) &  29.20  (4.99) &  0.99  (0.00) &  0.03  (0.21)  \\
Elevators08-5  &  {\bf 29.35}  (2.77) &  0.65  (0.04) &  -0.00  (0.36) &  21.86  (0.47) &  0.09  (0.00) &  0.44  (0.03)  \\
Elevators08-6  &  {\bf 34.52}  (4.09) &  0.24  (0.00) &  -0.09  (0.00) &  32.70  (2.96) &  0.41  (0.00) &  0.22  (0.03)  \\
Gripper8     &  21.86  (0.58) &  0.81  (0.00) &  0.06  (0.02) &  24.77  (3.56) &  0.98  (0.04) &  0.14  (0.00)  \\
Logistics00-10-1  &  11.68  (0.95) &  0.85  (0.00) &  0.25  (0.00) &  11.68  (2.14) &  0.85  (0.00) &  0.25  (0.05)  \\
Miconic11-0  &  13.15  (3.27) &  0.53  (0.00) &  0.24  (0.16) &  37.86  (0.81) &  0.02  (0.00) &  0.02  (0.02)  \\
Miconic11-2  &  8.53  (0.97) &  0.53  (0.00) &  0.74  (0.16) &  36.86  (0.65) &  0.02  (0.00) &  0.07  (0.01)  \\
NoMprime5    &  18.55  (0.69) &  0.95  (0.00) &  -0.06  (0.01) &  16.66  (0.44) &  0.94  (0.00) &  0.00  (0.02)  \\
Nomystery10  &  18.98  (4.04) &  0.42  (0.00) &  -0.07  (0.06) &  21.61  (1.44) &  0.74  (0.00) &  0.11  (0.04)  \\
Openstacks08-19  &  22.14  (1.19) &  0.38  (0.01) &  0.21  (0.05) &  17.11  (1.28) &  0.34  (0.00) &  0.32  (0.13)  \\
Openstacks08-21  &  25.67  (0.82) &  0.15  (0.00) &  0.31  (0.04) &  {\bf 39.34}  (0.52) &  0.92  (0.00) &  0.05  (0.11)  \\
Parcprinter11-11  &  16.85  (2.71) &  0.74  (0.00) &  0.41  (0.49) &  15.98  (1.44) &  0.82  (0.00) &  0.56  (0.03)  \\
Parking11    &  28.43  (1.01) &  0.98  (0.00) &  0.02  (0.03) &  26.76  (3.07) &  0.97  (0.00) &  0.07  (0.14)  \\
Pegsol11-18  &  16.22  (0.27) &  0.77  (0.00) &  0.05  (0.01) &  {\bf 26.17}  (0.26) &  0.34  (0.00) &  -0.03  (0.00)  \\
PipesNoTk10  &  15.58  (0.36) &  0.98  (0.00) &  0.01  (0.00) &  15.22  (0.35) &  0.98  (0.00) &  0.02  (0.00)  \\
PipesTk12    &  19.84  (3.18) &  0.99  (0.01) &  0.01  (0.00) &  21.40  (0.94) &  0.88  (0.00) &  0.04  (0.02)  \\
PipesTk17    &  26.64  (0.20) &  0.98  (0.00) &  0.00  (0.00) &  28.82  (0.13) &  0.88  (0.00) &  0.00  (0.00)  \\
Rovers6      &  33.49  (1.01) &  0.56  (0.00) &  0.01  (0.02) &  41.00  (2.13) &  0.31  (0.00) &  0.03  (0.02)  \\
Scanalyzer08-6  &  20.28  (2.22) &  0.77  (0.00) &  0.01  (0.00) &  23.70  (1.53) &  0.66  (0.00) &  0.01  (0.00)  \\
Scanalyzer11-6  &  16.36  (3.89) &  0.65  (0.00) &  0.49  (0.16) &  38.82  (1.64) &  0.30  (0.00) &  0.09  (0.01)  \\ \hline
Average      &  21.39  (1.94) &  0.71  (0.00) &  0.13  (0.10) &  25.86  (1.67) &  0.64  (0.00) &  0.13  (0.06)  \\ \hline
Total walltime & \multicolumn{3}{r|}{398.75 (36.16)} & \multicolumn{3}{r}{331.18 (21.39)} 

	\end{tabular}
	}

	\centering
        \resizebox{0.9\textwidth}{!}{
	\begin{tabular}{l|rrr|rrr} \hline
		Instance & \multicolumn{3}{c|}{DAHDA*}  & \multicolumn{3}{c}{ZHDA*}  \\
		 & \multicolumn{3}{c|}{\DAHDAT{}}  & \multicolumn{3}{c}{\ZHDAT{}}  \\
	& speedup & CO & SO   & speedup & CO & SO  \\ \hline
Blocks10-0   &  25.11  (4.89) &  0.88  (0.00) &  0.08  (0.05) &  14.93  (4.05) &  0.98  (0.00) &  0.30   (0.25) \\ 
Blocks11-1   &  24.88  (2.00) &  0.91  (0.00) &  0.21  (0.01) &  27.98  (2.28) &  0.98  (0.00) &  0.07   (0.09) \\ 
Elevators08-5  &  27.59  (4.07) &  0.83  (0.01) &  -0.03  (0.05) &  27.54  (2.72) &  0.98  (0.01) &  -0.03   (0.03) \\ 
Elevators08-6  &  15.28  (1.77) &  0.88  (0.00) &  0.31  (0.06) &  18.19  (3.15) &  0.96  (0.00) &  0.06   (0.14) \\ 
Gripper8     &  21.80  (2.92) &  0.98  (0.04) &  0.08  (0.05) &  21.66  (3.42) &  0.98  (0.01) &  0.08   (0.03) \\ 
Logistics00-10-1  &  17.52  (0.80) &  0.84  (0.00) &  0.00  (0.00) &  16.09  (0.56) &  0.99  (0.00) &  0.00   (0.02) \\ 
Miconic11-0  &  {\bf 46.05}  (0.87) &  0.01  (0.00) &  0.08  (0.01) &  7.40  (2.74) &  0.96  (0.00) &  0.13   (0.04) \\ 
Miconic11-2  &  33.81  (1.35) &  0.01  (0.00) &  0.18  (0.00) &  14.67  (2.65) &  0.96  (0.00) &  0.05   (0.06) \\ 
NoMprime5    &  18.46  (0.59) &  0.90  (0.00) &  -0.05  (0.01) &  16.63  (0.57) &  0.98  (0.00) &  -0.02   (0.01) \\ 
Nomystery10  &  28.41  (2.29) &  0.60  (0.00) &  -0.07  (0.10) &  21.68  (3.30) &  0.99  (0.00) &  -0.07   (0.22) \\ 
Openstacks08-19  &  24.54  (1.05) &  0.24  (0.00) &  0.18  (0.03) &  {\bf 25.99}  (3.40) &  0.99  (0.00) &  -0.05   (0.19) \\ 
Openstacks08-21  &  26.72  (1.06) &  0.13  (0.00) &  0.28  (0.05) &  39.06  (2.71) &  0.92  (0.00) &  -0.00   (0.12) \\ 
Parcprinter11-11  &  7.00  (2.91) &  0.19  (0.01) &  4.38  (1.54) &  19.15  (2.95) &  0.97  (0.00) &  0.08   (0.16) \\ 
Parking11    &  28.84  (0.82) &  0.52  (0.00) &  0.07  (0.02) &  27.09  (3.55) &  0.98  (0.00) &  0.04   (0.16) \\ 
Pegsol11-18  &  22.16  (0.83) &  0.34  (0.00) &  -0.01  (0.02) &  16.97  (1.05) &  0.98  (0.00) &  0.03   (0.03) \\ 
PipesNoTk10  &  15.58  (0.46) &  0.98  (0.00) &  0.01  (0.00) &  11.22  (0.38) &  0.98  (0.00) &  0.03   (0.00) \\ 
PipesTk12    &  25.12  (0.31) &  0.67  (0.00) &  0.00  (0.00) &  19.78  (0.36) &  0.98  (0.00) &  0.00   (0.00) \\ 
PipesTk17    &  31.16  (0.58) &  0.60  (0.00) &  0.01  (0.00) &  26.27  (4.15) &  0.98  (0.01) &  0.00   (0.00) \\ 
Rovers6      &  25.48  (2.86) &  0.05  (0.00) &  0.26  (0.07) &  30.01  (2.50) &  0.76  (0.00) &  0.00   (0.07) \\ 
Scanalyzer08-6  &  21.23  (2.62) &  0.94  (0.00) &  0.00  (0.00) &  16.54  (0.43) &  0.98  (0.00) &  0.01   (0.00) \\ 
Scanalyzer11-6  &  19.51  (3.55) &  0.50  (0.00) &  0.46  (0.14) &  20.36  (0.66) &  0.98  (0.00) &  0.05   (0.01) \\ \hline
Average      &  24.11  (1.84) &  0.57  (0.00) &  0.31  (0.11) &  20.53  (2.27) &  0.96  (0.00) &  0.01   (0.08) \\ \hline
Total walltime & \multicolumn{3}{r|}{377.86 (28.85)} & \multicolumn{3}{r}{433.23 (47.90)}  \\
	\end{tabular}
	}
\end{table}

\end{appendices}

\clearpage

\bibliographystyle{theapa}
\bibliography{ref-jf16}

\begin{thebibliography}{}

\bibitem[\protect\BCAY{Asai\ \BBA\ Fukunaga}{Asai\ \BBA\
  Fukunaga}{2016}]{asai2016tiebreaking}
Asai, M.\BBACOMMA\  \BBA\ Fukunaga, A. \BBOP2016\BBCP.
\newblock \BBOQ Tiebreaking strategies for {A}* search: How to explore the
  final frontier\BBCQ\
\newblock In {\Bem Proceedings of the AAAI Conference on Artificial %
  Intelligence (AAAI)}.

\bibitem[\protect\BCAY{B{\"a}ckstr{\"o}m\ \BBA\ Nebel}{B{\"a}ckstr{\"o}m\ \BBA\
  Nebel}{1995}]{backstrom1995complexity}
B{\"a}ckstr{\"o}m, C.\BBACOMMA\  \BBA\ Nebel, B. \BBOP1995\BBCP.
\newblock \BBOQ Complexity results for {SAS}+ planning\BBCQ\
\newblock {\Bem Computational Intelligence}, {\Bem 11\/}(4), 625--655.

\bibitem[\protect\BCAY{Buluc, Meyerhenke, Safro, Sanders,\ \BBA\ Schulz}{Buluc
  et~al.}{2015}]{bulucc2013recent}
Buluc, A., Meyerhenke, H., Safro, I., Sanders, P., \BBA\ Schulz, C.
  \BBOP2015\BBCP.
\newblock \BBOQ Recent advances in graph partitioning\BBCQ\
\newblock {\Bem arXiv preprint arXiv:1311.3144}.

\bibitem[\protect\BCAY{Burns, Lemons, Ruml,\ \BBA\ Zhou}{Burns
  et~al.}{2010}]{burnslrz10}
Burns, E., Lemons, S., Ruml, W., \BBA\ Zhou, R. \BBOP2010\BBCP.
\newblock \BBOQ Best-first heuristic search for multicore machines\BBCQ\
\newblock {\Bem Journal of Artificial Intelligence Research (JAIR)}, {\Bem 39},
  689--743.

\bibitem[\protect\BCAY{Burns, Hatem, Leighton,\ \BBA\ Ruml}{Burns
  et~al.}{2012}]{burns2012implementing}
Burns, E.~A., Hatem, M., Leighton, M.~J., \BBA\ Ruml, W. \BBOP2012\BBCP.
\newblock \BBOQ Implementing fast heuristic search code\BBCQ\
\newblock In {\Bem Proceedings of the Annual Symposium on Combinatorial
  Search}, \BPGS\ 25--32.

\bibitem[\protect\BCAY{Edelkamp}{Edelkamp}{2001}]{edelkamp2001planning}
Edelkamp, S. \BBOP2001\BBCP.
\newblock \BBOQ Planning with pattern databases\BBCQ\
\newblock In {\Bem European Conference on Planning (ECP)}, \BPGS\ 13--24.

\bibitem[\protect\BCAY{Evans}{Evans}{2006}]{evans2006scalable}
Evans, J. \BBOP2006\BBCP.
\newblock \BBOQ A scalable concurrent {malloc (3)} implementation for
  {FreeBSD}\BBCQ\
\newblock In {\Bem Proc. BSDCan Conference}.

\bibitem[\protect\BCAY{Evett, Hendler, Mahanti,\ \BBA\ Nau}{Evett
  et~al.}{1995}]{evett1995massively}
Evett, M., Hendler, J., Mahanti, A., \BBA\ Nau, D. \BBOP1995\BBCP.
\newblock \BBOQ {PRA*}: {M}assively parallel heuristic search\BBCQ\
\newblock {\Bem Journal of Parallel and Distributed Computing}, {\Bem 25\/}(2),
  133--143.

\bibitem[\protect\BCAY{Fiduccia\ \BBA\ Mattheyses}{Fiduccia\ \BBA\
  Mattheyses}{1982}]{fiduccia1982linear}
Fiduccia, C.~M.\BBACOMMA\  \BBA\ Mattheyses, R.~M. \BBOP1982\BBCP.
\newblock \BBOQ A linear-time heuristic for improving network partitions\BBCQ\
\newblock In {\Bem Conference on Design Automation}, \BPGS\ 175--181.

\bibitem[\protect\BCAY{Fukunaga, Botea, Jinnai,\ \BBA\ Kishimoto}{Fukunaga
  et~al.}{2017}]{fukunaga2017survey}
Fukunaga, A., Botea, A., Jinnai, Y., \BBA\ Kishimoto, A. \BBOP2017\BBCP.
\newblock \BBOQ A survey of parallel {A}*\BBCQ\
\newblock {\Bem arXiv preprint arXiv:1708.05296}.

\bibitem[\protect\BCAY{Hart, Nilsson,\ \BBA\ Raphael}{Hart
  et~al.}{1968}]{HartNR68}
Hart, P.~E., Nilsson, N.~J., \BBA\ Raphael, B. \BBOP1968\BBCP.
\newblock \BBOQ A formal basis for the heuristic determination of minimum cost
  paths\BBCQ\
\newblock {\Bem {IEEE} Transactions on Systems Science and Cybernetics}, {\Bem
  4\/}(2), 100--107.

\bibitem[\protect\BCAY{Haslum\ \BBA\ Geffner}{Haslum\ \BBA\
  Geffner}{2000}]{HaslumG00}
Haslum, P.\BBACOMMA\  \BBA\ Geffner, H. \BBOP2000\BBCP.
\newblock \BBOQ Admissible heuristics for optimal planning\BBCQ\
\newblock In {\Bem Proceedings of the International Conference on Automated
  Planning and Scheduling (ICAPS)}, \BPGS\ 140--149.

\bibitem[\protect\BCAY{Helmert}{Helmert}{2006}]{Helmert2006}
Helmert, M. \BBOP2006\BBCP.
\newblock \BBOQ The {Fast Downward} planning system\BBCQ\
\newblock {\Bem Journal of Artificial Intelligence Research}, {\Bem 26},
  191--246.

\bibitem[\protect\BCAY{Helmert, Haslum,\ \BBA\ Hoffmann}{Helmert
  et~al.}{2007}]{helmert2007flexible}
Helmert, M., Haslum, P., \BBA\ Hoffmann, J. \BBOP2007\BBCP.
\newblock \BBOQ Flexible abstraction heuristics for optimal sequential
  planning\BBCQ\
\newblock In {\Bem Proceedings of the International Conference on Automated
  Planning and Scheduling (ICAPS)}, \BPGS\ 176--183.

\bibitem[\protect\BCAY{Helmert, Haslum, Hoffmann,\ \BBA\ Nissim}{Helmert
  et~al.}{2014}]{helmert2014merge}
Helmert, M., Haslum, P., Hoffmann, J., \BBA\ Nissim, R. \BBOP2014\BBCP.
\newblock \BBOQ Merge-and-shrink abstraction: A method for generating lower
  bounds in factored state spaces\BBCQ\
\newblock {\Bem Journal of the ACM (JACM)}, {\Bem 61\/}(3), 16.

\bibitem[\protect\BCAY{Hendrickson\ \BBA\ Kolda}{Hendrickson\ \BBA\
  Kolda}{2000}]{hendrickson2000graph}
Hendrickson, B.\BBACOMMA\  \BBA\ Kolda, T.~G. \BBOP2000\BBCP.
\newblock \BBOQ Graph partitioning models for parallel computing\BBCQ\
\newblock {\Bem Parallel computing}, {\Bem 26\/}(12), 1519--1534.

\bibitem[\protect\BCAY{Holzmann}{Holzmann}{2008}]{holzmann2008stack}
Holzmann, G.~J. \BBOP2008\BBCP.
\newblock \BBOQ A stack-slicing algorithm for multi-core model checking\BBCQ\
\newblock {\Bem Electronic Notes in Theoretical Computer Science}, {\Bem
  198\/}(1), 3--16.

\bibitem[\protect\BCAY{Holzmann\ \BBA\ Bo\^{s}na\^{c}ki}{Holzmann\ \BBA\
  Bo\^{s}na\^{c}ki}{2007}]{holzmannb07}
Holzmann, G.~J.\BBACOMMA\  \BBA\ Bo\^{s}na\^{c}ki, D. \BBOP2007\BBCP.
\newblock \BBOQ The design of a multicore extension of the {SPIN} model
  checker\BBCQ\
\newblock {\Bem IEEE Transactions on Software Engineering}, {\Bem 33\/}(10),
  659--674.

\bibitem[\protect\BCAY{Horie\ \BBA\ Fukunaga}{Horie\ \BBA\
  Fukunaga}{2017}]{HorieF17}
Horie, S.\BBACOMMA\  \BBA\ Fukunaga, A.~S. \BBOP2017\BBCP.
\newblock \BBOQ Block-parallel {IDA}* for {GPUs}\BBCQ\
\newblock In {\Bem Proceedings of the Tenth International Symposium on
  Combinatorial Search, Edited by Alex Fukunaga and Akihiro Kishimoto, 16-17
  June 2017, Pittsburgh, Pennsylvania, {USA.}}, \BPGS\ 134--138.

\bibitem[\protect\BCAY{Imai\ \BBA\ Fukunaga}{Imai\ \BBA\
  Fukunaga}{2015}]{ImaiF15}
Imai, T.\BBACOMMA\  \BBA\ Fukunaga, A. \BBOP2015\BBCP.
\newblock \BBOQ On a practical, integer-linear programming model for
  delete-free tasks and its use as a heuristic for cost-optimal planning\BBCQ\
\newblock {\Bem Journal of Artificial Intelligence Research}, {\Bem 54},
  631--677.

\bibitem[\protect\BCAY{Iosup, Ostermann, Yigitbasi, Prodan, Fahringer,\ \BBA\
  Epema}{Iosup et~al.}{2011}]{iosup2011performance}
Iosup, A., Ostermann, S., Yigitbasi, M.~N., Prodan, R., Fahringer, T., \BBA\
  Epema, D.~H. \BBOP2011\BBCP.
\newblock \BBOQ Performance analysis of cloud computing services for many-tasks
  scientific computing\BBCQ\
\newblock {\Bem IEEE Transactions on Parallel and Distributed Systems}, {\Bem
  22\/}(6), 931--945.

\bibitem[\protect\BCAY{Irani\ \BBA\ Shih}{Irani\ \BBA\ Shih}{1986}]{IraniS86}
Irani, K.\BBACOMMA\  \BBA\ Shih, Y. \BBOP1986\BBCP.
\newblock \BBOQ Parallel {A*} and {AO*} algorithms: An optimality criterion and
  performance evaluation\BBCQ\
\newblock In {\Bem International Conference on Parallel Processing}, \BPGS\
  274--277.

\bibitem[\protect\BCAY{Jabbar\ \BBA\ Edelkamp}{Jabbar\ \BBA\
  Edelkamp}{2006}]{JabbarE06}
Jabbar, S.\BBACOMMA\  \BBA\ Edelkamp, S. \BBOP2006\BBCP.
\newblock \BBOQ Parallel external directed model checking with linear
  {I/O}\BBCQ\
\newblock In {\Bem Verification, Model Checking, and Abstract Interpretation,
  7th International Conference, {VMCAI} 2006, Charleston, SC, USA, January
  8-10, 2006, Proceedings}, \BPGS\ 237--251.

\bibitem[\protect\BCAY{Jinnai\ \BBA\ Fukunaga}{Jinnai\ \BBA\
  Fukunaga}{2016a}]{jinnai2016structured}
Jinnai, Y.\BBACOMMA\  \BBA\ Fukunaga, A. \BBOP2016a\BBCP.
\newblock \BBOQ Abstract {Z}obrist hashing: An efficient work distribution
  method for parallel best-first search\BBCQ\
\newblock In {\Bem Proceedings of the AAAI Conference on Artificial %
  Intelligence (AAAI)}, \BPGS\ 717--723.

\bibitem[\protect\BCAY{Jinnai\ \BBA\ Fukunaga}{Jinnai\ \BBA\
  Fukunaga}{2016b}]{jinnai2016automated}
Jinnai, Y.\BBACOMMA\  \BBA\ Fukunaga, A. \BBOP2016b\BBCP.
\newblock \BBOQ Automated creation of efficient work distribution functions for
  parallel best-first search\BBCQ\
\newblock In {\Bem Proceedings of the International Conference on Automated
  Planning and Scheduling (ICAPS)}.

\bibitem[\protect\BCAY{Jonsson\ \BBA\ B{\"a}ckstr{\"o}m}{Jonsson\ \BBA\
  B{\"a}ckstr{\"o}m}{1998}]{jonsson1998state}
Jonsson, P.\BBACOMMA\  \BBA\ B{\"a}ckstr{\"o}m, C. \BBOP1998\BBCP.
\newblock \BBOQ State-variable planning under structural restrictions:
  Algorithms and complexity\BBCQ\
\newblock {\Bem Artificial Intelligence}, {\Bem 100\/}(1), 125--176.

\bibitem[\protect\BCAY{Jyothi, Singla, Godfrey,\ \BBA\ Kolla}{Jyothi
  et~al.}{2014}]{jyothi2014measuring}
Jyothi, S.~A., Singla, A., Godfrey, P., \BBA\ Kolla, A. \BBOP2014\BBCP.
\newblock \BBOQ Measuring and understanding throughput of network
  topologies\BBCQ\
\newblock {\Bem arXiv preprint arXiv:1402.2531}.

\bibitem[\protect\BCAY{Karypis\ \BBA\ Kumar}{Karypis\ \BBA\
  Kumar}{1998}]{karypis1998fast}
Karypis, G.\BBACOMMA\  \BBA\ Kumar, V. \BBOP1998\BBCP.
\newblock \BBOQ A fast and high quality multilevel scheme for partitioning
  irregular graphs\BBCQ\
\newblock {\Bem SIAM Journal on scientific Computing}, {\Bem 20\/}(1),
  359--392.

\bibitem[\protect\BCAY{Kishimoto, Fukunaga,\ \BBA\ Botea}{Kishimoto
  et~al.}{2013}]{kishimotofb13}
Kishimoto, A., Fukunaga, A., \BBA\ Botea, A. \BBOP2013\BBCP.
\newblock \BBOQ Evaluation of a simple, scalable, parallel best-first search
  strategy\BBCQ\
\newblock {\Bem Artificial Intelligence}, {\Bem 195}, 222--248.

\bibitem[\protect\BCAY{Kishimoto, Fukunaga,\ \BBA\ Botea}{Kishimoto
  et~al.}{2009}]{kishimotofb09}
Kishimoto, A., Fukunaga, A.~S., \BBA\ Botea, A. \BBOP2009\BBCP.
\newblock \BBOQ Scalable, parallel best-first search for optimal sequential
  planning\BBCQ\
\newblock In {\Bem Proceedings of the International Conference on Automated
  Planning and Scheduling (ICAPS)}, \BPGS\ 201--208.

\bibitem[\protect\BCAY{Kobayashi, Kishimoto,\ \BBA\ Watanabe}{Kobayashi
  et~al.}{2011}]{Kobayashi2011evaluations}
Kobayashi, Y., Kishimoto, A., \BBA\ Watanabe, O. \BBOP2011\BBCP.
\newblock \BBOQ Evaluations of {Hash Distributed A*} in optimal sequence
  alignment\BBCQ\
\newblock In {\Bem Proceedings of the International Joint Conference on %
  Artificial Intelligence (IJCAI)}, \BPGS\ 584--590.

\bibitem[\protect\BCAY{Korf}{Korf}{1985}]{korf:85a}
Korf, R. \BBOP1985\BBCP.
\newblock \BBOQ Depth-first iterative deepening: An optimal admissible tree
  search\BBCQ\
\newblock {\Bem Artificial Intelligence}, {\Bem 97}, 97--109.

\bibitem[\protect\BCAY{Korf\ \BBA\ Felner}{Korf\ \BBA\ Felner}{2002}]{Korf2002}
Korf, R.~E.\BBACOMMA\  \BBA\ Felner, A. \BBOP2002\BBCP.
\newblock \BBOQ Disjoint pattern database heuristics\BBCQ\
\newblock {\Bem Artificial Intelligence}, {\Bem 134\/}(1), 9--22.

\bibitem[\protect\BCAY{Korf\ \BBA\ Schultze}{Korf\ \BBA\
  Schultze}{2005}]{korf2005large}
Korf, R.~E.\BBACOMMA\  \BBA\ Schultze, P. \BBOP2005\BBCP.
\newblock \BBOQ Large-scale parallel breadth-first search\BBCQ\
\newblock In {\Bem Proceedings of the AAAI Conference on Artificial %
  Intelligence (AAAI)}, \BPGS\ 1380--1385.

\bibitem[\protect\BCAY{Korf, Zhang, Thayer,\ \BBA\ Hohwald}{Korf
  et~al.}{2005}]{korf2005frontier}
Korf, R.~E., Zhang, W., Thayer, I., \BBA\ Hohwald, H. \BBOP2005\BBCP.
\newblock \BBOQ Frontier search\BBCQ\
\newblock {\Bem Journal of the ACM (JACM)}, {\Bem 52\/}(5), 715--748.

\bibitem[\protect\BCAY{Kumar, Ramesh,\ \BBA\ Rao}{Kumar
  et~al.}{1988}]{Kumar1988parallel}
Kumar, V., Ramesh, K., \BBA\ Rao, V.~N. \BBOP1988\BBCP.
\newblock \BBOQ Parallel best-first search of state-space graphs: A summary of
  results.\BBCQ\
\newblock In {\Bem Proceedings of the AAAI Conference on Artificial %
  Intelligence (AAAI)}, \lowercase{\BVOL}~88, \BPGS\ 122--127.

\bibitem[\protect\BCAY{Leighton\ \BBA\ Rao}{Leighton\ \BBA\
  Rao}{1999}]{leighton1999multicommodity}
Leighton, T.\BBACOMMA\  \BBA\ Rao, S. \BBOP1999\BBCP.
\newblock \BBOQ Multicommodity max-flow min-cut theorems and their use in
  designing approximation algorithms\BBCQ\
\newblock {\Bem Journal of the ACM (JACM)}, {\Bem 46\/}(6), 787--832.

\bibitem[\protect\BCAY{Mahapatra\ \BBA\ Dutt}{Mahapatra\ \BBA\
  Dutt}{1997}]{mahapatra1997scalable}
Mahapatra, N.~R.\BBACOMMA\  \BBA\ Dutt, S. \BBOP1997\BBCP.
\newblock \BBOQ Scalable global and local hashing strategies for duplicate
  pruning in parallel {A}* graph search\BBCQ\
\newblock {\Bem {IEEE} Transactions on Parallel and Distributed Systems}, {\Bem
  8\/}(7), 738--756.

\bibitem[\protect\BCAY{Niewiadomski, Amaral,\ \BBA\ Holte}{Niewiadomski
  et~al.}{2006}]{niewiadomski06}
Niewiadomski, R., Amaral, J.~N., \BBA\ Holte, R.~C. \BBOP2006\BBCP.
\newblock \BBOQ Sequential and parallel algorithms for frontier {A*} with
  delayed duplicate detection\BBCQ\
\newblock In {\Bem Proceedings of the 21st National Conference on Artificial
  Intelligence (AAAI)}, \BPGS\ 1039--1044.

\bibitem[\protect\BCAY{Pearl}{Pearl}{1984}]{Pearl84}
Pearl, J. \BBOP1984\BBCP.
\newblock {\Bem Heuristics - Intelligent Search Strategies for Computer Problem
  Solving}.
\newblock Addison--Wesley.

\bibitem[\protect\BCAY{Pearson}{Pearson}{1990}]{pearson1990}
Pearson, W.~R. \BBOP1990\BBCP.
\newblock \BBOQ Rapid and sensitive sequence comparison with {FASTP} and
  {FASTA}\BBCQ\
\newblock {\Bem Methods in enzymology}, {\Bem 183}, 63--98.
\newblock Matrix score is available at
  http://prowl.rockefeller.edu/aainfo/pam250.htm.

\bibitem[\protect\BCAY{Phillips, Likhachev,\ \BBA\ Koenig}{Phillips
  et~al.}{2014}]{PhillipsLK14}
Phillips, M., Likhachev, M., \BBA\ Koenig, S. \BBOP2014\BBCP.
\newblock \BBOQ {PA*SE}: Parallel {A*} for slow expansions\BBCQ\
\newblock In {\Bem Proceedings of the International Conference on Automated
  Planning and Scheduling (ICAPS)}.

\bibitem[\protect\BCAY{Pommerening, R\"oger, Helmert,\ \BBA\ Bonet}{Pommerening
  et~al.}{2014}]{PommereningRHB2014}
Pommerening, F., R\"oger, G., Helmert, M., \BBA\ Bonet, B. \BBOP2014\BBCP.
\newblock \BBOQ {LP}-based heuristics for cost-optimal planning\BBCQ\
\newblock In {\Bem Proceedings of the International Conference on Automated
  Planning and Scheduling (ICAPS)}.

\bibitem[\protect\BCAY{Romein, Plaat, Bal,\ \BBA\ Schaeffer}{Romein
  et~al.}{1999}]{romein1999transposition}
Romein, J.~W., Plaat, A., Bal, H.~E., \BBA\ Schaeffer, J. \BBOP1999\BBCP.
\newblock \BBOQ Transposition table driven work scheduling in distributed
  search\BBCQ\
\newblock In {\Bem Proceedings of the AAAI Conference on Artificial %
  Intelligence (AAAI)}, \BPGS\ 725--731.

\bibitem[\protect\BCAY{Thompson, Koehl, Ripp,\ \BBA\ Poch}{Thompson
  et~al.}{2005}]{Thompson2005}
Thompson, J.~D., Koehl, P., Ripp, R., \BBA\ Poch, O. \BBOP2005\BBCP.
\newblock \BBOQ {BAliBASE} 3.0: {L}atest developments of the multiple sequence
  alignment benchmark\BBCQ\
\newblock {\Bem Proteins: Structure, Function and Genetics ({PROTEINS})}, {\Bem
  61\/}(1), 127--136.

\bibitem[\protect\BCAY{Zhou\ \BBA\ Hansen}{Zhou\ \BBA\
  Hansen}{2004}]{zhou2004structured}
Zhou, R.\BBACOMMA\  \BBA\ Hansen, E.~A. \BBOP2004\BBCP.
\newblock \BBOQ Structured duplicate detection in external-memory graph
  search\BBCQ\
\newblock In {\Bem Proceedings of the AAAI Conference on Artificial %
  Intelligence (AAAI)}, \BPGS\ 683--689.

\bibitem[\protect\BCAY{Zhou\ \BBA\ Hansen}{Zhou\ \BBA\
  Hansen}{2006a}]{zhou2006breadth}
Zhou, R.\BBACOMMA\  \BBA\ Hansen, E.~A. \BBOP2006a\BBCP.
\newblock \BBOQ Breadth-first heuristic search\BBCQ\
\newblock {\Bem Artificial Intelligence}, {\Bem 170\/}(4), 385--408.

\bibitem[\protect\BCAY{Zhou\ \BBA\ Hansen}{Zhou\ \BBA\
  Hansen}{2006b}]{Zhou2006}
Zhou, R.\BBACOMMA\  \BBA\ Hansen, E.~A. \BBOP2006b\BBCP.
\newblock \BBOQ Domain-independent structured duplicate detection\BBCQ\
\newblock In {\Bem Proceedings of the AAAI Conference on Artificial %
  Intelligence (AAAI)}, \BPGS\ 1082--1087.

\bibitem[\protect\BCAY{Zhou\ \BBA\ Hansen}{Zhou\ \BBA\
  Hansen}{2007}]{zhou2007parallel}
Zhou, R.\BBACOMMA\  \BBA\ Hansen, E.~A. \BBOP2007\BBCP.
\newblock \BBOQ Parallel structured duplicate detection\BBCQ\
\newblock In {\Bem Proceedings of the AAAI Conference on Artificial %
  Intelligence (AAAI)}, \BPGS\ 1217--1223.

\bibitem[\protect\BCAY{Zhou\ \BBA\ Zeng}{Zhou\ \BBA\
  Zeng}{2015}]{zhou2015massively}
Zhou, Y.\BBACOMMA\  \BBA\ Zeng, J. \BBOP2015\BBCP.
\newblock \BBOQ Massively parallel {A}* search on a {GPU}\BBCQ\
\newblock In {\Bem Proceedings of the AAAI Conference on Artificial %
  Intelligence (AAAI)}, \BPGS\ 1248--1255.

\bibitem[\protect\BCAY{Zobrist}{Zobrist}{1970}]{Zobrist1970}
Zobrist, A.~L. \BBOP1970\BBCP.
\newblock \BBOQ A new hashing method with application for game playing\BBCQ\
\newblock {\Bem {reprinted in} {I}nternational Computer Chess Association
  Journal ({ICCA})}, {\Bem 13\/}(2), 69--73.

\end{thebibliography}

\end{document}